# An Algebraic Graphical Model for Decision with Uncertainties, Feasibilities, and Utilities

**Cédric Pralet** CEDRIC.PRALET@ONERA.FR
*ONERA Toulouse, France*
*2 av. Edouard Belin, 31400 Toulouse*

**Gérard Verfaillie** GERARD.VERFAILLIE@ONERA.FR
*ONERA Toulouse, France*
*2 av. Edouard Belin, 31400 Toulouse*

**Thomas Schiex** THOMAS.SCHIEX@TOULOUSE.INRA.FR
*INRA Toulouse, France*
*Chemin de Borde Rouge, 31320 Castanet-Tolosan*

## Abstract

Numerous formalisms and dedicated algorithms have been designed in the last decades to model and solve decision making problems. Some formalisms, such as constraint networks, can express "simple" decision problems, while others are designed to take into account uncertainties, unfeasible decisions, and utilities. Even in a single formalism, several variants are often proposed to model different types of uncertainty (probability, possibility...) or utility (additive or not). In this article, we introduce an algebraic graphical model that encompasses a large number of such formalisms: (1) we first adapt previous structures from Friedman, Chu and Halpern for representing uncertainty, utility, and expected utility in order to deal with generic forms of sequential decision making; (2) on these structures, we then introduce composite graphical models that express information via variables linked by "local" functions, thanks to conditional independence; (3) on these graphical models, we finally define a simple class of queries which can represent various scenarios in terms of observabilities and controllabilities. A natural decision-tree semantics for such queries is completed by an equivalent operational semantics, which induces generic algorithms. The proposed framework, called the *Plausibility-Feasibility-Utility* (PFU) framework, not only provides a better understanding of the links between existing formalisms, but it also covers yet unpublished frameworks (such as possibilistic influence diagrams) and unifies formalisms such as quantified boolean formulas and influence diagrams. Our backtrack and variable elimination generic algorithms are a first step towards unified algorithms.

## 1. Introduction

In the last decades, numerous formalisms have been developed to express and solve decision making problems. In such problems, an agent must make decisions consisting of either choosing actions and ways to fulfill them (as in action planning, task scheduling, or resource allocation), or choosing explanations of observed phenomena (as in diagnosis or situation assessment). These choices may depend on various parameters:

1. uncertainty measures, which we call *plausibilities*, may describe beliefs about the state of the environment;

2. preconditions may have to be satisfied for a decision to be *feasible*;

3. the possible states of the environment and decisions do not generally have the same value from the decision makers point of view. *Utilities* can be expressed to model costs, gains, risks, satisfaction degrees, hard requirements, and more generally, preferences;

4. when time is involved, decision processes may be *sequential* and the environment may be *partially observable*. This means that there may be several decision steps, and that the values of some variables may be observed between two steps, as in chess where each player plays in turn and can observe the move of the opponent before playing again;

5. there may be adversarial or collaborative decision makers, each of them controlling a set of decisions. Hence, a multi-agent aspect can yield partial controllabilities.

Given the plausibilities defined over the states of the environment, the feasibility constraints on the decisions, the utilities defined over the decisions and the states of the environment, and given the possible multiple decision steps, the objective is to provide the decision maker with optimal decision rules for the decision variables he controls, depending on the environment and on other agents. To be concise, the class of such problems is denoted as the class of *sequential decision problems with plausibilities, feasibilities, and utilities*.

Various formalisms have been designed to cope with problems of this class, sometimes in a degenerated form (covering only a subset of the features of the general problem):

- formalisms developed in the boolean satisfiability framework: the satisfiability problem (SAT), quantified boolean formulas, stochastic SAT (Littman, Majercik, & Pitassi, 2001), and extended stochastic SAT (Littman et al., 2001);

- formalisms developed in the very close constraint satisfaction framework: constraint satisfaction problems (CSPs, Mackworth, 1977), valued/semiring CSPs (Bistarelli, Montanari, Rossi, Schiex, Verfaillie, & Fargier, 1999) (covering classical, fuzzy, additive, lexicographic, probabilistic CSPs), mixed CSPs and probabilistic mixed CSPs (Fargier, Lang, & Schiex, 1996), quantified CSPs (Bordeaux & Monfroy, 2002), and stochastic CSPs (Walsh, 2002);

- formalisms developed to represent uncertainty and extended to represent decision problems under uncertainty: Bayesian networks (Pearl, 1988), Markov random fields (Chellappa & Jain, 1993) (also known as Gibbs networks), chain graphs (Frydenberg, 1990), hybrid or mixed networks (Dechter & Larkin, 2001; Dechter & Mateescu, 2004), influence diagrams (Howard & Matheson, 1984), unconstrained (Jensen & Vomlelova, 2002), asymmetric (Smith, Holtzman, & Matheson, 1993; Nielsen & Jensen, 2003), or sequential (Jensen, Nielsen, & Shenoy, 2004) influence diagrams, valuation networks (Shenoy, 1992), and asymmetric (Shenoy, 2000) or sequential (Demirer & Shenoy, 2001) valuation networks;

- formalisms developed in the classical planning framework, such as STRIPS planning (Fikes & Nilsson, 1971; Ghallab, Nau, & Traverso, 2004), conformant planning (Goldman & Boddy, 1996), and probabilistic planning (Kushmerick, Hanks, & Weld, 1995);

- formalisms such as Markov decision processes (MDPs), probabilistic, possibilistic, or using Spohn's epistemic beliefs (Spohn, 1990; Wilson, 1995; Giang & Shenoy, 2000), factored or not, possibly partially observable (Puterman, 1994; Monahan, 1982; Sabbadin, 1999; Boutilier, Dean, & Hanks, 1999; Boutilier, Dearden, & Goldszmidt, 2000).

Many of these formalisms present interesting similarities:

- they include variables modeling the state of the environment (*environment variables*) or the decisions (*decision variables*);
- they use sets of functions which, depending on the formalism considered, can model plausibilities, feasibilities, or utilities;
- they use operators either to combine local information (such as $\times$ to aggregate probabilities under independence hypothesis, $+$ to aggregate gains and costs), or to project a global information (such as $+$ to compute a marginal probability, min or max to compute an optimal decision).

Even if the meaning of variables, functions, and combination or projection operators may be specific to each formalism, they can all be seen as *graphical models* in the sense that they exploit, implicitly or explicitly, a hypergraph of local functions between variables. This article shows that it is possible to build a generic algebraic framework subsuming many of these formalisms by reducing decision making problems to a sequence of so-called "variable eliminations" on an aggregation of local functions.

Such a generic framework will be able to provide:

- *a better understanding of existing formalisms*: a generic framework has an obvious theoretical and pedagogical interest, since it can bring to light similarities and differences between the formalisms covered and help people of different communities to communicate on a common basis;
- *increased expressive power*: a generic framework may be able to capture problems that cannot be directly modeled in any existing formalism. This increased expressiveness should be reachable by capturing the essential algebraic properties of existing frameworks;
- *generic algorithms*: ultimately, besides a generic framework, it should be possible to define generic algorithms capable of solving problems defined in this framework. This objective fits into a growing effort to identify common algorithmic approaches that were developed for solving different AI problems. It may also facilitate cross-fertilization by allowing each subsumed framework to reuse algorithmic ideas defined in another one.

#### 1.0.1 Article Outline

After the introduction of some notations and notions, the article starts by showing, with a catalog of existing formalisms for decision making, that a generic algebraic framework can be informally identified. This generic framework, called the Plausibility-Feasibility-Utility (PFU) framework, is then formally introduced in three steps: (1) algebraic structures capturing plausibilities, feasibilities, and utilities are introduced (Section 4), (2) these algebraic structures are exploited to build a generic form of graphical model (Section 5), and (3) problems over such graphical models are captured by the notion of queries (Section 6). The framework is analyzed in Section 7 and generic algorithms are defined in Section 8. A table recapitulating the main notations used is available in Appendix A and the proofs of all propositions and theorems appear in Appendix B. A short version of the framework described in this article has already been published (Pralet, Verfaillie, & Schiex, 2006c).

## 2. Background Notations and Definitions

The essential objects used in the article are variables, domains, and local functions (called scoped functions).

**Definition 1.** *The* domain *of values of a variable $x$ is denoted $dom(x)$ and for every $a \in dom(x)$, $(x, a)$ denotes the assignment of value $a$ to $x$. By extension, for a set of variables $S$, we denote by $dom(S)$ the Cartesian product of the domains of the variables in $S$, i.e. $dom(S) = \prod_{x \in S} dom(x)$. An element $A$ of $dom(S)$ is called an* assignment *of $S$.*[1]

*If $A_1$, $A_2$ are assignments of disjoint subsets $S_1$, $S_2$, then $A_1.A_2$, called the concatenation of $A_1$ and $A_2$, is the assignment of $S_1 \cup S_2$ where variables in $S_1$ are assigned as in $A_1$ and variables in $S_2$ are assigned as in $A_2$. If $A$ is an assignment of a set of variables $S$, the projection of $A$ onto $S' \subseteq S$ is the assignment of $S'$ where variables are assigned to their value in $A$.*

**Definition 2.** *(Scoped function) A* scoped function *is a pair $(S, \varphi)$ where $S$ is a set of variables and $\varphi$ is a function mapping elements in $dom(S)$ to a given set $E$. In the following, we will often consider that $S$ is implicit and denote a scoped function $(S, \varphi)$ as $\varphi$ alone. The set of variables $S$ is called the* scope *of $\varphi$ and is denoted $sc(\varphi)$. If $A$ is an assignment of a superset of $sc(\varphi)$ and $A'$ is the projection of $A$ onto $sc(\varphi)$, we define $\varphi(A)$ by $\varphi(A) = \varphi(A')$.*

For example, a scoped function $\varphi$ mapping assignments of $sc(\varphi)$ to elements of the boolean lattice $\mathbb{B} = \{t, f\}$ is analogous to a constraint describing the subset of $dom(sc(\varphi))$ of authorized tuples in constraint networks.

From this, the general notion of graphical model can be defined:

**Definition 3.** *(Graphical model) A* graphical model *is a pair $(V, \Phi)$ where $V = \{x_1, \ldots, x_n\}$ is a finite set of variables and $\Phi = \{\varphi_1, \ldots, \varphi_m\}$ is a finite set of scoped functions whose scopes are included in $V$.*

The terminology of *graphical* models is used here simply because a set of scoped functions can be represented as a hypergraph that contains one hyperedge per function scope. As we will see, this hypergraph captures a form of independence (see Section 5) and induces parameters for the time and space complexity of our algorithms (see Section 8). This definition of graphical models generalizes the usual one used in statistics, defining a graphical

1. An assignment of $S = \{x_1, \ldots, x_k\}$ is actually a set of variable-value pairs $\{(x_1, a_1), \ldots, (x_k, a_k)\}$; we assume that variables are implicit when using a tuple of values $(a_1, \ldots, a_k) \in dom(S)$.

model as a (directed or not) graph where the nodes represent random variables and where the structure captures probabilistic independence relations.

"Local" scoped functions in a graphical model give a space-tractable definition of a global function defined by their aggregation. For example, in a Bayesian network (Pearl, 1988) a global probability distribution $P_{x,y,z}$ over $x, y, z$ may be defined as the product (using operator $\times$) of a set of scoped functions $\{P_x, P_{y|x}, P_{z|y}\}$. Local scoped functions can also facilitate the projection of the information expressed by a graphical model onto a smaller scope. For example, in order to compute a marginal probability distribution $\mathcal{P}_{y,z}$ from the previous network, we can compute $\sum_x P_{x,y,z} = (\sum_x P_x \times P_{y|x}) \times P_{z|y}$ and avoid taking $P_{z|y}$ into account. Here the operator $\sum$ is used to project information onto a smaller scope by eliminating variable $x$. Operators used to combine scoped functions will be called *combination* operators, while operators used to project information onto smaller scopes will be called *elimination* operators.

**Definition 4.** *(Combination) Let $\varphi_1$, $\varphi_2$ be scoped functions to $E_1$ and $E_2$ respectively. Let $\otimes : E_1 \times E_2 \rightarrow E$ be a binary operator. The combination of $\varphi_1$ and $\varphi_2$, denoted by $\varphi_1 \otimes \varphi_2$, is the scoped function to $E$ with scope $sc(\varphi_1) \cup sc(\varphi_2)$ defined by $(\varphi_1 \otimes \varphi_2)(A) = \varphi_1(A) \otimes \varphi_2(A)$ for all assignments $A$ of $sc(\varphi_1) \cup sc(\varphi_2)$. $\otimes$ is called the* combination operator *of $\varphi_1$ and $\varphi_2$.*

In the rest of the article, all *combination operators* will be denoted $\otimes$.

**Definition 5.** *(Elimination) Let $\varphi$ be a scoped function to $E$. Let op be an on $E$ which is associative, commutative, and has an identity element . The elimination of variable $x$ from $\varphi$ with op is a scoped function whose scope is $sc(\varphi) - \{x\}$ and whose value for an assignment $A$ of its scope is $(op_x\, \varphi)(A) = op_{a \in dom(x)}\, \varphi(A.(x, a))$. In this context, op is called the* elimination operator *for $x$. The elimination of a set of variables $S = \{x_1, \ldots, x_k\}$ from $\varphi$ is a function with scope $sc(\varphi) - S$ defined by $op_S\, \varphi(A) = op_{A' \in dom(S)}\, \varphi(A.A')$.*

Hence, when computing $\sum_x (P_x \times P_{y|x} \times P_{z|x})$, scoped functions are aggregated using the combination operator $\otimes = \times$ and the information is projected by eliminating $x$ using the elimination operator $+$. In this article, $\oplus$ denotes elimination operators.

In some cases, the elimination of a set of variables $S$ with an operator *op* from a scoped function $\varphi$ should only be performed on a subset of $dom(S)$ containing assignments that satisfy some property denoted by a boolean scoped function $F$. Then, we must compute for every $A \in dom(sc(\varphi) - S)$ the value $op_{A' \in dom(S), F(A')=t}\, \varphi(A.A')$. For simplicity and homogeneity, and in order to always use elimination over $dom(S)$, we can equivalently truncate $\varphi$ so that elements of $dom(S)$ which do not satisfy $F$ are mapped to a special value (denoted $\diamondsuit$) which is itself defined as a new identity for *op*.

**Definition 6.** *(Truncation operator) The* unfeasible value *$\diamondsuit$ is a new special element that is supposed to be outside of the domain $E$ of every elimination operator $op : E \times E \rightarrow E$. We explicitly extend every elimination operator $op : E \times E \rightarrow E$ on $E \cup \{\diamondsuit\}$ by taking the convention $op(\diamondsuit, e) = op(e, \diamondsuit) = e$ for all $e \in E \cup \{\diamondsuit\}$.*

*Let $\{t, f\}$ be the boolean lattice. For any boolean $b$ and any $e \in E$, we define $b \star e$ to be equal to $e$ if $b = t$ and $\diamondsuit$ otherwise. $\star$ is called the* truncation operator.

Given a boolean scoped function $F$, $\Diamond$ and $\star$ make it possible to write quantities like $op_{A' \in dom(S), F(A')=t}\, \varphi$ as the elimination $op_S\,(F \star \varphi)$.

In order to solve decision problems, one usually wants to compute functions mapping the available information to a decision. The notion of *decision rules* will be used to formalize this:

**Definition 7.** *(Decision rule, policy) A* decision rule *for a variable $x$ given a set of variables $S'$ is a function $\delta : dom(S') \rightarrow dom(x)$ mapping each assignment of $S'$ to a value in $dom(x)$. By extension, a decision rule for a set of variables $S$ given a set of variables $S'$ is a function $\delta : dom(S') \rightarrow dom(S)$. A set of decision rules is called a* policy*.*

**Definition 8.** *(Optimal decision rule) Consider a totally $\preceq$-ordered set $E$, a scoped function $\varphi$ from $dom(sc(\varphi))$ to $E$, and a set of variables $S \subseteq sc(\varphi)$. A decision rule $\delta : dom(sc(\varphi) - S) \rightarrow dom(S)$. is said to be* optimal *iff, for all $(A, A') \in dom(sc(\varphi) - S) \times dom(S)$, $\varphi(A.\delta(A)) \succeq \varphi(A.A')$ (resp. $\varphi(A.\delta(A)) \preceq \varphi(A.A')$). Such a decision rule always exists when $dom(sc(\varphi))$ is finite.*

In other words, optimal decision rules are examples of decision rules given by argmin and argmax (in this article, we consider that optimality on decision rules is always given by min or max on some totally ordered set).

**Definition 9.** *(Directed Acyclic Graph (DAG)) A directed graph $G$ is a DAG if it contains no directed cycle. When variables are used as vertices, $pa_G(x)$ denotes the set of parents of variable $x$ in $G$.*

Last, $[1, n]$ will denote the set of integers $i$ such that $1 \leq i \leq n$.

## 3. From Examples of Graphical Models to a Generic Framework

We now present different AI formalisms for expressing and solving decision problems. In the most simple case, a single decision which maximizes utility is sought. The introduction of plausibilities (uncertainties), unfeasible actions (feasibilities), and sequential decision (several decision steps with some observations between decision steps) only appears in the most sophisticated frameworks. The goal of this section is to show that these formalisms can all be viewed as graphical models where specific elimination and combination operators are used.

### 3.1 Examples of Graphical Models

The examples used cover various AI formalisms, which are briefly described. A wider and more accurate review of existing graphical models could be provided (Pralet, 2006).

#### 3.1.1 Constraint Networks

Constraint networks (CNs, Mackworth, 1977), often called *constraint satisfaction problems* (CSPs), are graphical models $(V, \Phi)$ where the scoped functions in $\Phi$ are constraints mapping assignments onto $\{t, f\}$. The usual query on a CN is to determine the existence of an assignment of $V$ that satisfies all constraints. By setting $f \prec t$, this decision problem can be answered by computing:

$$\max_V \left( \wedge_{\varphi \in \Phi} \varphi \right). \tag{1}$$

If this quantity equals true, then an optimal decision rule for $V$ defines a solution. This query can be answered by performing eliminations (using max) on a combination of scoped functions (using $\wedge$). Replacing the hard constraints in $\Phi$ by soft constraints (boolean scoped functions being replaced by cost functions) and replacing $\wedge$ by an abstract operator $\otimes$ equal to $+$, min, $\times$, ... leads to queries on a *valued* and *totally ordered semiring CN* (Bistarelli et al., 1999).

### 3.1.2 Bayesian Networks

Bayesian networks (BNs, Pearl, 1988) can model problems with plausibilities expressed as probabilities. A BN is a graphical model $(V, \Phi)$ in which $\Phi$ is a set of local conditional probability distributions: $\Phi = \{P_{x \mid pa_G(x)}, x \in V\}$, where $G$ is a DAG with vertices in $V$. A BN represents a joint probability distribution $P_V$ over all variables as a combination of local conditional probability distributions ($P_V = \prod_{x \in V} P_{x \mid pa_G(x)}$), just as the combination of local constraints in a CN defines a global constraint over all variables. One possible query on a BN is to compute the marginal probability distribution of a variable $y \in V$:

$$P_y = \sum_{V - \{y\}} P_V = \sum_{V - \{y\}} \left( \prod_{x \in V} P_{x \mid pa_G(x)} \right). \tag{2}$$

Equation 2 corresponds to variable eliminations (with $+$) on a product of scoped functions. In other queries on BNs such as MAP (Maximum A Posteriori hypothesis), eliminations with max are also performed.

### 3.1.3 Quantified Boolean Formulas and Quantified CNs

Quantified boolean formulas (QBFs) and quantified CNs (Bordeaux & Monfroy, 2002) can model sequential decision problems. Let $x_1$, $x_2$, $x_3$ be boolean variables. A QBF using the so-called prenex conjunctive normal form looks like (with $f \prec t$):

$$\exists x_1 \forall x_2 \exists x_3 ((\neg x_1 \vee x_3) \wedge (x_2 \vee x_3)) = \max_{x_1} \min_{x_2} \max_{x_3} ((\neg x_1 \vee x_3) \wedge (x_2 \vee x_3)). \tag{3}$$

Thus, the query "Is there a value for $x_1$ such that for all values of $x_2$, there exists a value for $x_3$ such that the clauses $\neg x_1 \vee x_3$ and $x_2 \vee x_3$ are satisfied?" can be answered as in Equation 3 using a sequence of eliminations (max over $x_1$, min over $x_2$, and max over $x_3$) on a conjunction of clauses. In a quantified CN, clauses are replaced by constraints.

### 3.1.4 Stochastic CNs

A stochastic CN (Walsh, 2002) can model sequential decision problems with probabilities as plausibilities and hard requirements as utilities, provided that the decisions do not influence the environment (the so-called *contingency* assumption). In a stochastic CN, two types of variables are defined: decision variables $d_i$ and environment (stochastic) variables $s_j$. A global probability distribution over the environment variables is expressed as a combination of local probability distributions. If the environment variables are mutually independent, these local probability distributions are simply unary probability distributions $P_{s_j}$. Finally, a stochastic CN defines a set of constraints $\{C_1, \ldots, C_m\}$ mapping tuples of values onto $\{0, 1\}$ (instead of $\{t, f\}$). This allows constraints to be multiplied with probabilities.

Consider a situation where first two decisions $d_1$ and $d_2$ are made, then an environment variable $s_1$ is observed, then decisions $d_3$ and $d_4$ are made, while an environment variable $s_2$ remains unobserved. A possible query on a stochastic CN is to compute decision rules for $d_1, d_2, d_3$, and $d_4$ which maximize the expected constraint satisfaction, through Equation 4:

$$\max_{d_1,d_2} \sum_{s_1} \max_{d_3,d_4} \sum_{s_2} (P_{s_1} \times P_{s_2}) \times \Big(\prod\nolimits_{i\in[1,m]} C_i\Big). \tag{4}$$

The answer to the query defined by Equation 4 is determined by a sequence of eliminations (max over the decision variables, + over the environment ones) on a combination of scoped functions (probabilities are combined using $\times$, constraints are combined using $\times$, since they are expressed onto $\{0, 1\}$ instead of $\{t, f\}$, and probabilities are combined with constraints using $\times$).

#### 3.1.5 Influence Diagrams

Influence diagrams (Howard & Matheson, 1984) can model sequential decision problems with probabilities as plausibilities together with gains and costs as utilities. They can be seen as an extension of BNs including the notions of decision and utility. More precisely, an *influence diagram* is a composite graphical model defined by three sets of variables organized in a DAG $G$: (1) a set $S$ of *chance variables*; for each $s \in S$, a conditional probability distribution $P_{s\,|\,pa_G(s)}$ of $s$ given its parents in $G$ is specified; (2) a set $D$ of *decision variables*; for each $d \in D$, $pa_G(d)$ is the set of variables observed before decision $d$ is made; (3) a set $\Gamma = \{u_1, \ldots, u_m\}$ of *utility variables*, each of which is associated with a utility function $U_i = U_{pa_G(u_i)}$ of scope $pa_G(u_i)$. Utility variables must be leaves in the DAG, and the utility functions define a global additive utility $U_G = \sum_{i\in[1,m]} U_i$.

The usual problem associated with an influence diagram is to compute decision rules maximizing the global expected utility. If we modify the example used for stochastic CNs by replacing $P_{s_1}$ by $P_{s_1\,|\,d_2}$, $P_{s_2}$ by $P_{s_2\,|\,d_1,d_3}$, and the constraints $C_1, \ldots, C_m$ by the additive utility functions $U_1, \ldots, U_m$, then an optimal policy can be obtained by computing optimal decision rules for $d_1$, $d_2$, $d_3$, and $d_4$ in Equation 5:

$$\max_{d_1,d_2} \sum_{s_1} \max_{d_3,d_4} \sum_{s_2} \left(P_{s_1\,|\,d_2} \times P_{s_2\,|\,d_1,d_3}\right) \times \Big(\sum\nolimits_{i\in[1,m]} U_i\Big). \tag{5}$$

Again, the answer to the query defined by Equation 5 can be computed by a sequence of eliminations (alternating max- and sum-eliminations) on a combination of scoped functions (plausibilities combined using $\times$, utilities combined using $+$, plausibilities and utilities combined using $\times$).

### 3.1.6 Valuation Networks

Valuation networks (Shenoy, 1992) can model sequential decision problems with plausibilities, feasibilities, and utilities, where plausibilities are combined using $\times$ and where utilities are additive. A valuation network is composed of several sets of *nodes* and *valuations*: (1) a set $D$ of decision nodes, (2) a set $S$ of chance nodes, (3) a set $F$ of indicator valuations, which specify unfeasible assignments of decision and chance variables, (4) a set $P$ of probability valuations, which are multiplicative factors of a joint probability distribution over the chance variables, and (5) a set $U$ of utility valuations, representing additive factors of a joint utility function $U_G = \sum_{U_i \in U} U_i$. Arcs between nodes are also used to define the order in which decisions are made and chance variables are observed. If this order is $d_1 \prec d_2 \prec s_1 \prec d_3 \prec d_4 \prec s_2$, it can be shown that optimal decision rules for $d_1$, $d_2$, $d_3$, $d_4$ are defined through Equation 6:

$$\max_{d_1,d_2} \sum_{s_1} \max_{d_3,d_4} \sum_{s_2} \left( \left( \bigwedge_{F_i \in F} F_i \right) \star \left( \prod_{P_i \in P} P_i \right) \times \left( \sum_{U_i \in U} U_i \right) \right). \tag{6}$$

Local feasibility constraints are combined using $\wedge$, and combined with other scoped functions using the truncation operator $\star$ (cf. Definition 6).

### 3.1.7 Finite Horizon Markov Decision Processes

Finite horizon Markov Decision Processes (MDPs, Puterman, 1994; Monahan, 1982; Sabbadin, 1999; Boutilier et al., 1999, 2000) model sequential decision problems with plausibilities and utilities over a horizon of $T$ time-steps. For every time-step $t$, a variable $s_t$ represents the state of the environment and a variable $d_t$ represents the decision made after observing $s_t$. In factored MDPs, several state variables may be used at each time-step.

In a *probabilistic* finite horizon MDP, plausibilities over the environment are described by local probability distributions $P_{s_{t+1} \mid s_t, d_t}$ of being in state $s_{t+1}$ given $s_t$ and $d_t$. The utilities over states and decisions are local additive rewards $R_{s_t,d_t}$, and boolean functions $F_{d_t \mid s_t}$ can express whether a decision $d_t$ is feasible in state $s_t$. An optimal policy for each initial state $s_1$ can be computed by the following equation (which is a bit unusual for defining optimal policies for a MDP, but is equivalent to the usual form):

$$\max_{d_1} \oplus_{u\, s_2} \max_{d_2} \ldots \oplus_{u\, s_T} \max_{d_T} \left( \bigwedge_{t \in [1,T]} F_{d_t|s_t} \right) \star \left( \underset{t \in [1,T[}{\otimes_p} P_{s_{t+1}|s_t,d_t} \right) \otimes_{pu} \left( \underset{t \in [1,T]}{\otimes_u} R_{s_t,d_t} \right). \tag{7}$$

Plausibilities are combined using $\otimes_p = \times$, utilities are combined using $\otimes_u = +$, plausibilities and utilities are combined using $\otimes_{pu} = \times$, decision variables are eliminated using max, and environment variables are eliminated using $\oplus_u = +$. The truncation operator $\star$ enables the elimination operators to ignore unfeasible decisions.

In a *pessimistic possibilistic* finite horizon MDP (Sabbadin, 1999), probability distributions $P_{s_{t+1} \mid s_t, d_t}$ are replaced by possibility distributions $\pi_{s_{t+1} \mid s_t, d_t}$, rewards $R_{s_t,d_t}$ are replaced by preferences $\mu_{s_t,d_t}$, and the operators used are $\oplus_u = \otimes_p = \otimes_u = \min$ and $\otimes_{pu} : (p,u) \to \max(1-p, u)$.

### 3.2 Towards a Generic Framework

The previous section shows that usual queries in various existing formalisms can be reduced to a sequence of variable eliminations on a combination of scoped functions.

This observation has led to the definition of *algebraic MDPs* (Perny, Spanjaard, & Weng, 2005) or to the definition of *valuation algebras* (Shenoy, 1991; Kolhas, 2003), a generic algebraic framework in which eliminations can be performed on a combination of scoped functions. However, valuation algebras use only one combination operator, whereas several combination operators may be needed to manipulate different types of scoped functions (as previously shown). Moreover, valuation algebras can deal with only one type of elimination, whereas several elimination operators may be required for handling the different types of variables. In *valuation networks* (Shenoy, 2000), plausibilities are necessarily represented as probabilities, and eliminations with min cannot be performed. Essentially, a more powerful framework is needed.

In order to be simple and yet general enough to cover the queries defined by Equations 1 to 7, the generic form we should consider is:

$$Sov\left(\left(\bigwedge_{F_i \in F} F_i\right) \star \left(\bigotimes_{p\,P_i \in P} P_i\right) \otimes_{pu} \left(\bigotimes_{u\,U_i \in U} U_i\right)\right). \tag{8}$$

where (1) $\wedge$ is used to combine local feasibilities, $\otimes_p$ is used to combine plausibilities, $\otimes_u$ is used to combine utilities, $\otimes_{pu}$ is used to combine plausibilities and utilities, and the truncation operator $\star$ is used to ignore unfeasible decisions without having to deal with elimination operations over restricted domains;[2] (2) $F$, $P$, $U$ are (possibly empty) sets of local feasibility, plausibility, and utility functions respectively; (3) $Sov$ is an operator-variable(s) sequence, indicating how to eliminate variables. $Sov$ involves min or max to eliminate decision variables and an operator $\oplus_u$ to eliminate environment variables.

Equation 8 is still very informal. To define it formally, and to provide it with clear semantics, we need to define three key elements:

1. We must capture the essential properties of the combination operators $\otimes_p$, $\otimes_u$, $\otimes_{pu}$ used respectively to combine plausibilities, utilities, and plausibilities with utilities. We must also characterize the elimination operators $\oplus_u$ and $\oplus_p$ used to project information coming from utilities and plausibilities. These operators will define the *algebraic structure* of the PFU (Plausibility-Feasibility-Utility) framework.

2. On this algebraic structure, we must define a generic form of graphical model, involving a set of *variables* and sets of *scoped functions* expressing plausibilities, feasibilities, and utilities (sets $P$, $F$, $U$). Together, they will define a *PFU network*. The factored form offered by such graphical models must also be analyzed in order to understand when and how it can be applied to concisely represent global functions (using the notion of conditional independence).

2. In Equation 8, all plausibilities are combined using the same operator $\otimes_p$ and all utilities are combined using the same operator $\otimes_u$; we denote such models as composite graphical models because they include different types of scoped functions (plausibilities, feasibilities, and utilities). Beyond this, Equation 8 also allows for heterogeneous information among each type of scoped functions. For example, in order to manipulate both probabilities and possibilities, we can use $\otimes_p$ defined over probability-possibility pairs by $(p_1, \pi_1) \otimes_p (p_2, \pi_2) = (p_1 \times p_2, \min(\pi_1, \pi_2))$.

3. Finally, we must define *queries* on PFU networks capturing interesting decision problems. As Equation 8 shows, such queries will be defined by a sequence *Sov* of operator-variable(s) pairs, applied on the combination of the scoped functions in the network. The fact that the answer to such queries represents meaningful values from the decision theory point of view will be proved by relating approach.

### 3.3 Summary

We have informally shown that several queries in various formalisms dealing with plausibilities and/or feasibilities and/or utilities reduce to sequences of variable eliminations applied to combinations of scoped functions, using various operators. They can *intuitively* be covered by Equation 8.

The three key elements (an algebraic structure, a PFU network, and a sequence of variable eliminations) needed to *formally* define and give sense to this equation are introduced in Sections 4, 5, and 6.

## 4. The PFU Algebraic Structures

The first element of the PFU framework is an algebraic structure specifying how the information provided by plausibilities, feasibilities, and utilities is combined and synthesized. This algebraic structure is obtained by adapting previous structures defined by Friedman, Chu, and Halpern (Friedman & Halpern, 1995; Halpern, 2001; Chu & Halpern, 2003a) for representing uncertainties and expected utilities.

### 4.1 Definitions

**Definition 10.** $(E, \circledast)$ *is a* commutative monoid *iff $E$ is a set and $\circledast$ is a binary operator on $E$ which is associative ($x \circledast (y \circledast z) = (x \circledast y) \circledast z$), commutative ($x \circledast y = y \circledast x$), and has an identity $1_E \in E$ ($x \circledast 1_E = 1_E \circledast x = x$).*

**Definition 11.** $(E, \oplus, \otimes)$ *is a* commutative semiring *iff*

- *$(E, \oplus)$ is a commutative monoid, with an identity denoted $0_E$,*
- *$(E, \otimes)$ is a commutative monoid, with an identity denoted $1_E$,*
- *$0_E$ is annihilator for $\otimes$ ($x \otimes 0_E = 0_E$),*
- *$\otimes$ distributes over $\oplus$ ($x \otimes (y \oplus z) = (x \otimes y) \oplus (x \otimes z)$).*

**Definition 12.** *Let $(E_a, \oplus_a, \otimes_a)$ be a commutative semiring. Then, $(E_b, \oplus_b, \otimes_{ab})$ is a semi-module on $(E_a, \oplus_a, \otimes_a)$ iff*

- *$(E_b, \oplus_b)$ is a commutative monoid, with an identity denoted $0_{E_b}$,*
- *$\otimes_{ab} : E_a \times E_b \to E_b$ satisfies*
  - *$\otimes_{ab}$ distributes over $\oplus_b$ ($a \otimes_{ab} (b_1 \oplus_b b_2) = (a \otimes_{ab} b_1) \oplus_b (a \otimes_{ab} b_2)$),*
  - *$\otimes_{ab}$ distributes over $\oplus_a$ ($(a_1 \oplus_a a_2) \otimes_{ab} b = (a_1 \otimes_{ab} b) \oplus_b (a_2 \otimes_{ab} b)$),*

– *linearity property:* $a_1 \otimes_{ab} (a_2 \otimes_{ab} b) = (a_1 \otimes_a a_2) \otimes_{ab} b$,

– *for all* $b \in E_b$, $0_{E_a} \otimes_{ab} b = 0_{E_b}$ *and* $1_{E_a} \otimes_{ab} b = b$.

**Definition 13.** *Let $E$ be a set with a partial order $\preceq$. An operator $\circledast$ on $E$ is monotonic iff $(x \preceq y) \rightarrow (x \circledast z \preceq y \circledast z)$ for all $x, y, z \in E$.*

### 4.2 Plausibility Structure

Various forms of plausibilities exist (Halpern, 2003). The most usual one is *probabilities.* As shown previously, for example with Equation 2, probabilities are aggregated using $\otimes_p = \times$ as a combination operator, and projected using $\oplus_p = +$ as an elimination operator.

But plausibilities can also be expressed as *possibility degrees* in $[0, 1]$. Possibilities are eliminated using $\oplus_p = \max$ and *usually* combined using $\otimes_p = \min$. An interesting case appears when possibility degrees are booleans describing which states of the environment are completely possible or impossible. Plausibilities are then combined using $\otimes_p = \wedge$ and eliminated using $\oplus_p = \vee$.

Another example is Spohn's epistemic beliefs, also known as $\kappa$-rankings (kappa rankings, Spohn, 1990; Wilson, 1995; Giang & Shenoy, 2000). In this case, plausibilities are elements of $\mathbb{N} \cup \{+\infty\}$ called *surprise degrees*, 0 is associated with non-surprising situations, $+\infty$ is associated with completely surprising (impossible) situations, and more generally a surprise degree $k$ can be viewed as a probability of $\epsilon^k$ for an infinitesimal $\epsilon$. Surprise degrees are combined using $\otimes_p = +$ and eliminated using $\oplus_p = \min$.

To capture these various plausibility modeling frameworks, we start from Friedman-Halpern's work on *plausibility measures* (Friedman & Halpern, 1995; Halpern, 2001). Weydert (1994) and Darwiche-Ginsberg (1992) developed similar approaches.

**Friedman-Halpern's structure** Assume we want to express plausibilities over the assignments of a set of variables $S$. Each subset of $dom(S)$ is called an *event*. Friedman and Halpern (1995) define plausibilities as elements of a set $E_p$ called the plausibility domain. $E_p$ is equipped with a partial order $\preceq_p$ and with two special elements $0_p$ and $1_p$ satisfying $0_p \preceq_p p \preceq_p 1_p$ for all $p \in E_p$. A function $Pl : 2^{dom(S)} \rightarrow E_p$ is a plausibility measure over $S$ iff it satisfies $Pl(\emptyset) = 0_p$, $Pl(dom(S)) = 1_p$, and $(W_1 \subseteq W_2) \rightarrow (Pl(W_1) \preceq_p Pl(W_2))$. This means that $0_p$ is associated with impossibility, $1_p$ is associated with the highest plausibility degree, and the plausibility degree of a set is as least as high as the plausibility degree of each of its subsets.

Among all plausibility measures, we focus on so-called *algebraic conditional plausibility measures*, which use abstract functions $\oplus_p$ and $\otimes_p$ which are analogous to $+$ and $\times$ for probabilities. These measures satisfy properties such as *decomposability*: for all disjoint events $W_1$, $W_2$, $Pl(W_1 \cup W_2) = Pl(W_1) \oplus_p Pl(W_2)$. As $\cup$ is associative and commutative, it follows that $\oplus_p$ is associative and commutative on representations of disjoint events, i.e. $(a \oplus_p b) \oplus_p c = a \oplus_p (b \oplus_p c)$ and $a \oplus_p b = b \oplus_p a$ if there exist pairwise disjoint sets $W_1, W_2, W_3$ such that $Pl(W_1) = a$, $Pl(W_2) = b$, $Pl(W_3) = c$. More details are available in Friedman-Halpern's references (Friedman & Halpern, 1995; Halpern, 2001).

**Restriction of Friedman-Halpern's structure** An important aspect in Friedman-Halpern's work is that the algebraic properties of $\oplus_p$ and $\otimes_p$ hold only on the domains of definition of $\oplus_p$ and $\otimes_p$. Although this is sufficient to express and manipulate plausibilities, it can be algorithmically restrictive. Indeed, consider a Bayesian network involving two boolean variables $\{x_1, x_2\}$ and define $P_{x_1,x_2}$ as $P_{x_1} \times P_{x_2 \,|\, x_1}$. Assume that $P_{x_1}$ is a constant factor $\varphi_0 = 0.5$. In order to evaluate $P_{x_2}((x_2, t))$, the quantity $\sum_{x_1} \varphi_0 \times P_{x_2 \,|\, x_1}((x_2, t))$ must be computed. To do so, it is simpler to factor it and compute $\varphi_0 \times \sum_{x_1} P_{x_2 \,|\, x_1}((x_2, t))$. If $P_{x_2 \,|\, x_1}((x_2, t).(x_1, t)) = 0.6$ and $P_{x_2 \,|\, x_1}((x_2, t).(x_1, f)) = 0.8$, the answer is $0.5 \times (0.6 + 0.8) = 0.7$. Performing $0.6 + 0.8$ requires applying addition outside of the range of usual probabilities, for which $a \oplus_p b$ is defined only if $a + b \leq 1$, since two probabilities whose sum exceeds 1 cannot be associated with disjoint events.

To take such issues into account, we adapt Friedman-Halpern's $E_p$, $\oplus_p$, $\otimes_p$ so that $\oplus_p$ and $\otimes_p$ become closed in $E_p$ and so that Friedman-Halpern's axioms hold in the closed structure. Once this closure is performed, we obtain a *plausibility structure*.

**Definition 14.** *A* plausibility structure *is a tuple* $(E_p, \oplus_p, \otimes_p)$ *such that*

- $(E_p, \oplus_p, \otimes_p)$ *is a commutative semiring (identities for* $\oplus_p$ *and* $\otimes_p$ *are denoted* $0_p$ *and* $1_p$ *respectively),*
- $E_p$ *is equipped with a partial order* $\preceq_p$ *such that* $0_p = \min(E_p)$ *and such that* $\oplus_p$ *and* $\otimes_p$ *are monotonic with respect to* $\preceq_p$*.*

*Elements of* $E_p$ *are called* plausibility degrees

Note that $1_p$ is not necessarily the maximal element of $E_p$. For probabilities, Friedman-Halpern's structure would be $([0, 1], +', \times)$, where $a +' b = a + b$ if $a + b \leq 1$ and is undefined otherwise. In order to get closed operators, we take $(E_p, \oplus_p, \otimes_p) = (\mathbb{R}^+, +, \times)$ and therefore $1_p = 1$ is not the maximal element in $E_p$. In some cases, Friedman-Halpern's structure is already closed. This is the case with $\kappa$-rankings (where $(E_p, \oplus_p, \otimes_p) = (\mathbb{N} \cup \{+\infty\}, \min, +)$) and with possibilities (where $(E_p, \oplus_p, \otimes_p)$ is typically $([0, 1], \max, \min)$, although other choices like $([0, 1], \max, \times)$ are possible).

Given two plausibility structures $(E_p, \oplus_p, \otimes_p)$ and $(E'_p, \oplus'_p, \otimes'_p)$, if we define $E = E_p \times E'_p$, $(p_1, p'_1) \oplus (p_2, p'_2) = (p_1 \oplus_p p_2, p'_1 \oplus'_p p'_2)$ and $(p_1, p'_1) \otimes (p_2, p'_2) = (p_1 \otimes_p p_2, p'_1 \otimes'_p p'_2)$, then $(E, \oplus, \otimes)$ is a plausibility structure too. This allows us to deal with different kinds of plausibilities (such as probabilities and possibilities) or with families of probability distributions.

#### 4.2.1 From Plausibility Measures to Plausibility Distributions

Let us consider a plausibility measure (Friedman & Halpern, 1995; Halpern, 2001) $Pl : 2^{dom(S)} \rightarrow E_p$ over a set of variables $S$. Assume that $Pl(W_1 \cup W_2) = Pl(W_1) \oplus_p Pl(W_2)$ for all disjoint sets $W_1, W_2 \in 2^{dom(S)}$, as is the case with Friedman-Halpern's *algebraic plausibility measures*. This assumption entails that $Pl(W) = \oplus_{p\,A \in W}\, Pl(\{A\})$ for all $W \in 2^{dom(S)}$. This holds even for $W = \emptyset$ since $0_p$ is the identity for $\oplus_p$. Hence, defining $Pl(\{A\})$ for all complete assignments $A$ of $S$ suffices to describe $Pl$. Moreover, in this case, the three conditions defining plausibility measures ($Pl(dom(S)) = 1_p$, $Pl(\emptyset) = 0_p$, and $(W_1 \subseteq W_2) \rightarrow (Pl(W_1) \preceq_p Pl(W_2))$) are equivalent to just $\oplus_{p\,A \in dom(S)}\, Pl(\{A\}) = 1_p$, using the monotonicity of $\oplus_p$ for the third condition. This means that we can deal with *plausibility distributions* instead of plausibility measures:

**Definition 15.** *A* plausibility distribution *over $S$ is a function $\mathcal{P}_S : dom(S) \rightarrow E_p$ such that $\oplus_{p A \in dom(S)} \mathcal{P}_S(A) = 1_p$.*

The normalization condition imposed on plausibility distributions is simply a generalization of the convention that probabilities sum up to 1. It captures the fact that the disjunction of all the assignments of $S$ has $1_p$ as a plausibility degree.

**Proposition 1.** *A plausibility distribution $\mathcal{P}_S$ can be extended to give a plausibility distribution $\mathcal{P}_{S'}$ over every $S' \subset S$, defined by $\mathcal{P}_{S'} = \oplus_{p S-S'} \mathcal{P}_S$.*

### 4.3 Feasibility Structure

Feasibilities define whether a decision is possible or not, and are therefore expressed as booleans in $\{t, f\}$. This set is equipped with the total order $\preceq_{bool}$ satisfying $f \prec_{bool} t$.

Boolean scoped functions expressing feasibilities are combined using the operator $\wedge$, since an assignment of decision variables is feasible iff all feasibility functions agree that this assignment is feasible.

Given a scoped function $F_i$ expressing feasibilities, we can compute whether an assignment $A$ of a set $S$ of variables is feasible according to $F_i$ by computing $\vee_{sc(F_i)-S} F_i(A)$, since $A$ is feasible according to $F_i$ iff one of its extensions over $sc(F_i)$ is feasible. This means that the projection of feasibility functions onto a smaller scope uses the elimination operator $\vee$.

As a result, feasibilities are expressed using the *feasibility structure* $S_f = (\{t, f\}, \vee, \wedge)$. $S_f$ is not only a commutative semiring, but also a plausibility structure. Therefore, all plausibility notions and properties apply to feasibility. We may therefore speak of feasibility distributions, and the normalization condition $\vee_S \mathcal{F}_S = t$ imposed on a feasibility distribution $\mathcal{F}_S$ over $S$ means that at least one decision must be feasible.

### 4.4 Utility Structure

Utilities express preferences and can take various forms. Typically, utilities can be combined with $+$. But utilities can also model priorities combined using min. Also, when utilities represent hard requirements such as goals to be achieved or properties to be satisfied, they can be modeled as booleans combined using $\wedge$. More generally, utility degrees are defined as elements of a set $E_u$ equipped with a partial order $\preceq_u$. Smaller utility degrees are associated with less preferred events. Utility degrees are combined using an operator $\otimes_u$ which is assumed to be associative and commutative. This guarantees that combined utilities do not depend on the way combination is performed. We also assume that $\otimes_u$ admits an identity $1_u \in E_u$, representing indifference. This ensures the existence of a default utility degree when there are no utility scoped functions. These properties are captured in the following notion of *utility structure*.

**Definition 16.** *$(E_u, \otimes_u)$ is a* utility structure *iff it is a commutative monoid and $E_u$ is equipped with a partial order $\preceq_u$. Elements of $E_u$ are called utility degrees.*

$E_u$ may have a minimum element $\perp_u$ representing unacceptable events and which will be an annihilator for $\otimes_u$ (the combination of any event with an unacceptable one must be unacceptable too). $\otimes_u$ is also usually monotonic. But these properties are not necessary to establish the forthcoming results.

The distinction between plausibilities, feasibilities, and utilities is important and can be justified using algebraic arguments. Since $\otimes_p$ and $\otimes_u$ may be different operators (for example, $\otimes_p = \times$ and $\otimes_u = +$ in usual probabilities with additive utilities), we must distinguish plausibilities and utilities. It is also necessary to distinguish feasibilities from utilities or plausibilities. Indeed, imagine a simple card game involving two players $P_1$ and $P_2$, each having three cards: a jack $J$, a queen $Q$, and a king $K$. $P_1$ must first play one card $x \in \{J, Q, K\}$, then $P_2$ must play a card $y \in \{J, Q, K\}$, and last $P_1$ must play a card $z \in \{J, Q, K\}$. A rule forbids to play the same card consecutively (feasibility functions $F_{xy}$ : $x \neq y$ and $F_{yz} : y \neq z$). The goal for $P_1$ is that his two cards $x$ and $z$ have a value strictly better than $P_2$'s card $y$. By setting $J < Q < K$, this requirement corresponds to two utility functions $U_{xy} : x > y$ and $U_{yz} : z > y$. In order to compute optimal decisions in presence of unfeasibilities, we must restrict optimizations (eliminations of decision variables with max or min) to feasible values: instead of $\max_x \min_y \max_z (U_{xy} \wedge U_{yz})$, we must compute:

$$\max_{a \in dom(x)} \left( \min_{b \in dom(y), F_{xy}(a,b)=t} \left( \max_{c \in dom(z), F_{yz}(b,c)=t} (U_{xy}(a,b) \wedge U_{yz}(b,c)) \right) \right),$$

which, by setting $f \prec t$, is logically equivalent to

$$\max_x \min_y \left( F_{xy} \rightarrow \max_z (F_{yz} \wedge (U_{xy} \wedge U_{yz})) \right).$$

In the latter quantity, feasibility functions concerning $P_2$'s play ($y$) are taken into account using logical connective $\rightarrow$, so that $P_2$'s unfeasible decisions are ignored in the set of all scenarios considered. Feasibility functions concerning $P_1$'s last move ($z$) are taken into account using $\wedge$, so that $P_1$ does not consider scenarios in which he achieves a forbidden move. Therefore, feasibility functions cannot be handled simply by using the same combination operator as for utility functions: we need to dissociate what is unfeasible for all decision makers (unfeasibility is absolute) from what is unacceptable or required for one decision maker only (utility is relative), i.e. what a decision maker *wants* from what a decision maker *can* do.

At a more general level, for example when $U_{xy}$ and $U_{yz}$ are soft requirements or when we do not know exactly in advance who controls which variable, the logical connectives $\wedge$ and $\rightarrow$ cannot be used anymore. In order to ignore unfeasible values in decision variables elimination, we use the truncation operator $\star$ introduced in Definition 6. In order to eliminate a variable $x$ from a scoped function $\varphi$ while ignoring unfeasibilities indicated by a feasibility function $F_i$, we simply perform the elimination of $x$ on $(F_i \star \varphi)$ instead of $\varphi$. This maps unfeasibilities to value $\Diamond$, which is ignored by elimination operators (see Definition 6). On the example above, if $U_{xy}$ and $U_{yz}$ were additive gains and costs, we would compute

$$\max_x \min_y \left( F_{xy} \star \max_z (F_{yz} \star (U_{xy} + U_{yz})) \right).$$

### 4.5 Combining Plausibilities and Utilities via Expected Utility

To define expected utilities, plausibilities and utilities must be combined. Consider a situation where a utility $u_i$ is obtained with a plausibility $p_i$ for all $i \in [1, N]$, with $p_1 \oplus_p \ldots \oplus_p p_N = 1_p$. $\mathcal{L} = ((p_1, u_1), \ldots, (p_N, u_N))$ is classically called a lottery (von Neumann & Morgenstern, 1944). When we speak of expected utility, we implicitly speak of the expected utility $EU(\mathcal{L})$ of a lottery $\mathcal{L}$.

The standard way to combine plausibilities and utilities is to use the probabilistic expected utility (von Neumann & Morgenstern, 1944) defining $EU(\mathcal{L})$ as $\sum_{i\in[1,N]}(p_i \times u_i)$: it aggregates plausibilities and utilities using the combination operator $\otimes_{pu} = \times$ and projects the aggregated information using the elimination operator $\oplus_u = +$. However, alternative definitions exist:

- If plausibilities are possibilities, then $EU(\mathcal{L}) = \min_{i\in[1,N]} \max(1 - p_i, u_i)$ with the possibilistic *pessimistic* expected utility (Dubois & Prade, 1995) (i.e. $\oplus_u = \min$ and $\otimes_{pu} : (p, u) \to \max(1-p, u)$) and $EU(\mathcal{L}) = \max_{i\in[1,N]} \min(p_i, u_i)$ with the possibilistic *optimistic* expected utility (Dubois & Prade, 1995) (i.e. $\oplus_u = \max$ and $\otimes_{pu} = \min$).

- If plausibilities are $\kappa$-rankings and utilities are positive integers (Giang & Shenoy, 2000), then $EU(\mathcal{L}) = \min_{i\in[1,N]}(p_i + u_i)$ (i.e. $\oplus_u = \min$ and $\otimes_{pu} = +$).

To generalize these definitions of $EU(\mathcal{L})$, we start from Chu-Halpern's work on generalized expected utility (Chu & Halpern, 2003a, 2003b).

**Chu-Halpern's structure** Generalized expected utility is defined in an *expectation domain*, which is a tuple $(E_p, E_u, E'_u, \oplus_u, \otimes_{pu})$ such that: (1) $E_p$ is a set of plausibility degrees and $E_u$ is a set of utility degrees; (2) $\otimes_{pu} : E_p \times E_u \to E'_u$ combines plausibilities with utilities and satisfies $1_p \otimes_{pu} u = u$; (3) $\oplus_u : E'_u \times E'_u \to E'_u$ is a commutative and associative operator which can aggregate the information combined using $\otimes_{pu}$.

When a decision problem is *additive*, i.e. when, for all plausibility degrees $p_1, p_2$ associated with disjoint events, $(p_1 \oplus_p p_2) \otimes_{pu} u = (p_1 \otimes_{pu} u) \oplus_u (p_2 \otimes_{pu} u)$, the generic definition of the expected utility of a lottery is:

$$EU(\mathcal{L}) = \underset{i\in[1,N]}{\oplus_u} (p_i \otimes_{pu} u_i).$$

Classical expectation domains also satisfy additional properties such as "$\oplus_u$ is monotonic" and "$0_p \otimes_{pu} u = 0_u$, where $0_u$ is the identity of $\oplus_u$".

**Restriction of Chu-Halpern's structure for sequential decision making** If we use $\otimes_{pu} : E_p \times E_u \to E'_u$ and $\oplus_u : E'_u \times E'_u \to E'_u$ to compute expected utilities at the first decision step, then we need to introduce operators $\otimes'_{pu} : E_p \times E'_u \to E''_u$ and $\oplus'_u : E''_u \times E''_u \to E''_u$ to compute expected utilities at the second decision step. In the end, if there are $T$ decision steps, we must define $T$ operators $\otimes_{pu}$ and $T$ operators $\oplus_u$. In order to avoid the definition of an algebraic structure that would depend on the number of decision steps, we take $E_u = E'_u$ and work with only one operator $\otimes_{pu} : E_p \times E_u \to E_u$ and one operator $\oplus_u : E_u \times E_u \to E_u$.

As for plausibilities, and for the sake of the future algorithms, we restrict Chu-Halpern's expectation domains $(E_p, E_u, E_u, \oplus_u, \otimes_{pu})$ so that $\oplus_u$ and $\otimes_{pu}$ become closed and generalize properties of the initial $\oplus_u$ and $\otimes_{pu}$. However, this closure is not sufficient to deal with *sequential* decision making, because Chu-Halpern's expected utility is designed for *one-step* decision processes only. This is why we introduce three additional axioms for $\oplus_u$ and $\otimes_{pu}$:

- The first axiom is similar to a standard axiom for lotteries (von Neumann & Morgenstern, 1944) defining compound lotteries. It states that if a lottery $\mathcal{L}_2$ involves a utility $u$ with plausibility $p_2$, and if one of the utilities of a lottery $\mathcal{L}_1$ is the expected utility of $\mathcal{L}_2$ with plausibility $p_1$, then it is as if utility $u$ had been obtained with plausibility $p_1 \otimes_p p_2$. This gives the axiom $p_1 \otimes_{pu} (p_2 \otimes_{pu} u) = (p_1 \otimes_p p_2) \otimes_{pu} u$.

- We further require that $\otimes_{pu}$ distributes over $\oplus_u$. To justify this point, assume that a lottery $\mathcal{L} = ((p_1, u_1), (p_2, u_2))$ is obtained with plausibility $p$. Two different versions of the contribution of $\mathcal{L}$ to the global utility degree can be derived: the first is $p \otimes_{pu} ((p_1 \otimes_{pu} u_1) \oplus_u (p_2 \otimes_{pu} u_2))$, and the second, which uses compound lotteries, is $((p \otimes_p p_1) \otimes_{pu} u_1) \oplus_u ((p \otimes_p p_2) \otimes_{pu} u_2)$. We want these two quantities to be equal for all $p, p_1, p_2, u_1, u_2$. This can be shown to be equivalent to a simpler property $p \otimes_{pu} (u_1 \oplus_u u_2) = (p \otimes_{pu} u_1) \oplus_u (p \otimes_{pu} u_2)$, i.e. that $\otimes_{pu}$ distributes over $\oplus_u$.

- Finally, we assume that $\otimes_{pu}$ is right monotonic (i.e. $(u_1 \preceq_u u_2) \rightarrow (p \otimes_{pu} u_1 \preceq_u p \otimes_{pu} u_2)$). This means that if an agent prefers (strictly or not) an event $ev_2$ to another event $ev_1$, and if both events have the same plausibility degree $p$, then the contribution of $ev_2$ to the global expected utility degree must not be lesser than the contribution of $ev_1$.

These axioms define the notion of expected utility structure.

**Definition 17.** *Let $(E_p, \oplus_p, \otimes_p)$ be a plausibility structure and let $(E_u, \otimes_u)$ be a utility structure. $(E_p, E_u, \oplus_u, \otimes_{pu})$ is an* expected utility structure *iff*

- *$(E_u, \oplus_u, \otimes_{pu})$ is a semimodule on $(E_p, \oplus_p, \otimes_p)$ (cf. Definition 12),*
- *$\oplus_u$ is monotonic for $\preceq_u$ and $\otimes_{pu}$ is right monotonic for $\preceq_u$ $((u_1 \preceq_u u_2) \rightarrow (p \otimes_{pu} u_1 \preceq_u p \otimes_{pu} u_2))$.*

Many structures considered in the literature are instances of expected utility structures, as shown in Proposition 2. The results presented in the remaining of the article hold not only for these usual expected utility structures, but more generally for all structures satisfying the axioms specified in Definitions 14, 16, and 17.

**Proposition 2.** *The structures in Table 1 are expected utility structures.*

It is possible to define more complex expected utility structures from existing ones. For example, from two expected utility structures $(E_p, E_u, \oplus_u, \otimes_{pu})$ and $(E'_p, E'_u, \oplus'_u, \otimes'_{pu})$, it is possible to build a compound expected utility structure $(E_p \times E'_p, E_u \times E'_u, \oplus''_u, \otimes''_{pu})$. This can be used to deal simultaneously with probabilistic and possibilistic expected utility or more generally to deal with tuples of expected utilities.

**The business dinner example** To flesh out these definitions, we consider the following toy example, which will be referred to in the sequel. It does not correspond to a concrete real-life problem, but is used for its simplicity. *Peter invites John and Mary (a divorced couple) to a business dinner in order to convince them to invest in his company. Peter knows that if John is present at the end of the dinner, he will invest \$10K. The same holds for Mary with \$50K. Peter knows that John and Mary will not come together (one of them*

| | $E_p$ | $\preceq_p$ | $\oplus_p$ | $\otimes_p$ | $0_p, 1_p$ | $E_u$ | $\preceq_u$ | $\otimes_u$ | $\oplus_u$ | $\otimes_{pu}$ | $0_u, 1_u$ |
|---|---|---|---|---|---|---|---|---|---|---|---|
| 1 | $\mathbb{R}^+$ | $\leq$ | $+$ | $\times$ | $0, 1$ | $\mathbb{R} \cup \{-\infty\}$ | $\leq$ | $+$ | $+$ | $\times$ | $0, 0$ |
| 2 | $\mathbb{R}^+$ | $\leq$ | $+$ | $\times$ | $0, 1$ | $\mathbb{R}^+$ | $\leq$ | $\times$ | $+$ | $\times$ | $0, 1$ |
| 3 | $[0, 1]$ | $\leq$ | max | min | $0, 1$ | $[0, 1]$ | $\leq$ | min | max | min | $0, 1$ |
| 4 | $[0, 1]$ | $\leq$ | max | min | $0, 1$ | $[0, 1]$ | $\leq$ | min | min | $\max(1-p, u)$ | $1, 1$ |
| 5 | $\mathbb{N} \cup \{\infty\}$ | $\geq$ | min | $+$ | $\infty, 0$ | $\mathbb{N} \cup \{\infty\}$ | $\geq$ | $+$ | min | $+$ | $\infty, 0$ |
| 6 | $\{t, f\}$ | $\preceq_{bool}$ | $\vee$ | $\wedge$ | $f, t$ | $\{t, f\}$ | $\preceq_{bool}$ | $\wedge$ | $\vee$ | $\wedge$ | $f, t$ |
| 7 | $\{t, f\}$ | $\preceq_{bool}$ | $\vee$ | $\wedge$ | $f, t$ | $\{t, f\}$ | $\preceq_{bool}$ | $\wedge$ | $\wedge$ | $\rightarrow$ | $t, t$ |
| 8 | $\{t, f\}$ | $\preceq_{bool}$ | $\vee$ | $\wedge$ | $f, t$ | $\{t, f\}$ | $\preceq_{bool}$ | $\vee$ | $\vee$ | $\wedge$ | $f, f$ |
| 9 | $\{t, f\}$ | $\preceq_{bool}$ | $\vee$ | $\wedge$ | $f, t$ | $\{t, f\}$ | $\preceq_{bool}$ | $\vee$ | $\wedge$ | $\rightarrow$ | $t, f$ |

Table 1: Expected utility structures for: 1. probabilistic expected utility with additive utilities (allows the probabilistic expected utility of a cost or a gain to be computed); 2. probabilistic expected utility with multiplicative utilities, also called probabilistic expected satisfaction (allows the probability of satisfaction of some constraints to be computed); 3. possibilistic optimistic expected utility; 4. possibilistic pessimistic expected utility; 5. qualitative utility with $\kappa$-rankings and with only positive utilities; 6. boolean optimistic expected utility with conjunctive utilities (allows one to know whether there exists a possible world in which all goals of a set of goals $G$ are satisfied); $\preceq_{bool}$ denotes the order on booleans such that $f \prec_{bool} t$; 7. boolean pessimistic expected utility with conjunctive utilities (allows one to know whether in all possible worlds, all goals of a set of goals $G$ are satisfied); 8. boolean optimistic expected utility with disjunctive utilities (allows one to know whether there exists a possible world in which at least one goal of a set of goals $G$ is satisfied); 9. boolean pessimistic expected utility with disjunctive utilities (allows one to know whether in all possible worlds, at least one goal of a set of goals $G$ is satisfied).

*has to baby-sit their child), that at least one of them will come, and that the case "John comes and Mary does not" occurs with a probability of* 0.6*. As for the menu, Peter can order fish or meat for the main course, and white or red for the wine. However, the restaurant does not serve fish and red wine together. John does not like white wine and Mary does not like meat. If the menu does not suit them, they will leave the dinner. If John comes, Peter does not want him to leave the dinner because he is his best friend.*

**Example.** *The dinner problem uses the expected utility structure representing probabilistic expected additive utility (row 1 in Table 1): the plausibility structure is* $(\mathbb{R}^+, +, \times)$*,* $\oplus_u = +$*,* $\otimes_{pu} = \times$*, and utilities are additive gains* $((E_u, \otimes_u) = (\mathbb{R} \cup \{-\infty\}, +)$*, with the convention that* $u + (-\infty) = -\infty$*).*

### 4.6 Relation with Existing Structures

If we compare the algebraic structures defined with existing ones (Friedman & Halpern, 1995; Halpern, 2001; Chu & Halpern, 2003a), we can observe that:

- The structures defined here are less general than Friedman-Chu-Halpern's, since additional axioms are introduced. For example, plausibility structures are not able to model *belief functions* (Shafer, 1976), which are not decomposable, whereas this is possible using Friedman-Halpern's plausibility measures (however, the authors are not aware of existing schemes for decision theory using belief functions). Moreover, for one-step decision processes, Chu-Halpern's generalized expected utility is more general, since it assumes that $\otimes_{pu} : E_p \times E_u \rightarrow E'_u$ whereas we consider $\otimes_{pu} : E_p \times E_u \rightarrow E_u$.

- Conversely, the structures defined here can deal with multi-step decision processes whereas Chu-Halpern's generalized expected utility is designed for one-shot decision processes. Beyond this, other axioms, such as the use of closed operators, are essentially motivated by operational reasons. We use a less expressive structure for the sake of future algorithms (cf. Section 8).

As a set $E_p$ of plausibility degrees and a set $E_u$ of utility degrees are defined, plausibilities and utilities must be *cardinal*. *Purely ordinal* approaches such as CP-nets (Boutilier, Brafman, Domshlak, Hoos, & Poole, 2004), which, like Bayesian networks, exploit the notion of conditional independence to express a network of purely ordinal preference relations, are not covered.

As $\otimes_{pu}$ takes values in $E_u$, it is implicitly assumed that plausibilities and utilities are *commensurable*: works from Fargier and Perny (1999), describing a purely ordinal approach, where qualitative preferences and plausibilities are not necessarily commensurable, are not captured either. Also, works from Giang and Shenoy (2005), which satisfy all required associativity, commutativity, identity, annihilator, and distributivity properties, are not covered because they implicitly use $\otimes_{pu} : E_p \times E_u \rightarrow E'_u$ with $E_u \neq E'_u$ (even if the expected utility $EU(\mathcal{L}) = (p_1 \otimes_{pu} u_1) \oplus_u (p_2 \otimes_{pu} u_2)$ of a lottery $\mathcal{L} = ((p_1, u_1), (p_2, u_2))$ stays in $E_u$).

Furthermore, some axioms entail that only *distributional plausibilities* are covered (the plausibility of a set of variable assignments is determined by the plausibilities of each covered complete assignment): Dempster-Shafer *belief functions* (Shafer, 1976) or Choquet expected utility (Schmeidler, 1989) are not encompassed. Finally, as only one partial order $\preceq_u$ on $E_u$ is defined, it is assumed that the decision makers share the same preferences over utilities.

### 4.7 Summary

In this section, we have introduced *expected utility structures*, which are the first element of the PFU framework. They specify how plausibilities are combined and projected (using $\otimes_p$ and $\oplus_p$ respectively), how utilities are combined (using $\otimes_u$), and how plausibilities and utilities are aggregated to define generalized expected utility (using $\oplus_u$ and $\otimes_{pu}$). The structure chosen is inspired by Friedman-Chu-Halpern's plausibility measures and generalized expected utility. The main differences lie in the addition of axioms to deal with multi-step decision processes and in the use of extended domains to have closed operators, motivated by operational reasons.

## 5. Plausibility-Feasibility-Utility Networks

The second element of the PFU framework is a network of scoped functions $P_i$, $F_i$, and $U_i$ (cf. Equation 8) over a set of variables $V$. This network defines a compact and structured representation of the state of the environment, of the decisions, and of the global plausibilities, feasibilities, and utilities which hold over them.

In the rest of the article, a *plausibility function* denotes a scoped function to $E_p$ (the set of plausibility degrees), a *feasibility function* is a scoped function to $\{t, f\}$ (the set of feasibility degrees), and a *utility function*, a scoped function to $E_u$ (the set of utility degrees).

### 5.1 Variables

In structured representations, decisions are represented using *decision variables*, which are directly controlled by an agent, and the state of the environment is represented by *environment variables*, which are not directly controlled by an agent. The notion of agent used here is restricted to cooperative and adversarial decision makers (if there is an uncertainty on the way a decision maker behaves, then the decisions he controls will be modeled as environment variables). We use $V_D$ to denote the set of decision variables and $V_E$ to denote the set of environment variables. $V_D$ and $V_E$ form a partition of $V$.

**Example.** *The dinner problem can be modeled using six variables: $bp_J$ and $bp_M$ (value $t$ or $f$), representing John's and Mary's presence at the beginning, $ep_J$ and $ep_M$ (value $t$ or $f$), representing their presence at the end, $mc$ (value $fish$ or $meat$), representing the main course choice, and $w$ (value $white$ or $red$), representing the wine choice. Thus, we have $V_D = \{mc, w\}$ and $V_E = \{bp_J, bp_M, ep_J, ep_M\}$.*

### 5.2 Decomposing Plausibilities and Feasibilities into Local Functions

Using combined local functions to represent a global one raises some considerations: how and when such local functions can be obtained from a global one, and conversely, when such local functions are directly used, which implicit assumptions on the global function are made. We now show that all these questions boil down to the notion of conditional independence. In the following definitions and propositions, $(E_p, \oplus_p, \otimes_p)$ corresponds to a plausibility structure.

#### 5.2.1 Preliminaries: Generalization of Bayesian Networks Results

Assume that we want to express a global plausibility distribution $\mathcal{P}_S$ (cf. Definition 15) as a combination of local plausibility functions $P_i$. As work on Bayesian networks (Pearl, 1988) has shown, the factorization of a joint distribution is essentially related to the notion of conditional independence. To introduce conditional independence, we first define *conditional plausibility distributions*.

**Definition 18.** *A plausibility distribution $\mathcal{P}_S$ over $S$ is said to be* conditionable *iff there exists a set of functions denoted $\mathcal{P}_{S_1 \mid S_2}$ (one function for each pair $S_1, S_2$ of disjoint subsets of $S$) such that if $S_1, S_2$, $S_3$ are disjoint subsets of $S$, then*

(a) *for all assignments A of* $S_2$ *such that* $\mathcal{P}_{S_2}(A) \neq 0_p$, $\mathcal{P}_{S_1 \mid S_2}(A)$ *is a plausibility distribution over* $S_1$,[3]

(b) $\mathcal{P}_{S_1 \mid \emptyset} = \mathcal{P}_{S_1}$,

(c) $\oplus_{p S_1} \mathcal{P}_{S_1,S_2 \mid S_3} = \mathcal{P}_{S_2 \mid S_3}$,

(d) $\mathcal{P}_{S_1,S_2 \mid S_3} = \mathcal{P}_{S_1 \mid S_2,S_3} \otimes_p \mathcal{P}_{S_2 \mid S_3}$,

(e) $(\mathcal{P}_{S_1,S_2,S_3} = \mathcal{P}_{S_1 \mid S_3} \otimes_p \mathcal{P}_{S_2 \mid S_3} \otimes_p \mathcal{P}_{S_3}) \to (\mathcal{P}_{S_1,S_2 \mid S_3} = \mathcal{P}_{S_1 \mid S_3} \otimes_p \mathcal{P}_{S_2 \mid S_3})$.

$\mathcal{P}_{S_1 \mid S_2}$ *is called the* conditional plausibility distribution of $S_1$ given $S_2$.

Condition (a) means that conditional plausibility distributions must be normalized. Condition (b) means that the information given by an empty set of variables does not change the plausibilities over the states of the environment. Condition (c) means that conditional plausibility distributions are consistent from the marginalization point of view. Condition (d) is the analog of the so-called chain rule with probabilities. Condition (e) is a kind of weak division axiom.[4]

Proposition 3 gives simple conditions on a plausibility structure, satisfied in all usual frameworks, that suffice for plausibility distributions to be conditionable.

**Definition 19.** *A plausibility structure* $(E_p, \oplus_p, \otimes_p)$ *is called a* conditionable plausibility structure *iff it satisfies the axioms:*

- *if* $p_1 \preceq_p p_2$ *and* $p_2 \neq 0_p$, *then* $\max\{p \in E_p \mid p_1 = p \otimes_p p_2\}$ *exists and is* $\preceq_p 1_p$,
- *if* $p_1 \prec_p p_2$, *then there exists a unique* $p \in E_p$ *such that* $p_1 = p \otimes_p p_2$,
- *if* $p_1 \prec_p p_2$, *then there exists a unique* $p \in E_p$ *such that* $p_2 = p \oplus_p p_1$.

**Proposition 3.** *If* $(E_p, \oplus_p, \otimes_p)$ *is a conditionable plausibility structure, then all plausibility distributions are conditionable: it suffices to define* $\mathcal{P}_{S_1 \mid S_2}$ *by* $\mathcal{P}_{S_1 \mid S_2}(A) = \max\{p \in E_p \mid \mathcal{P}_{S_1,S_2}(A) = p \otimes_p \mathcal{P}_{S_2}(A)\}$ *for all* $A \in dom(S_1 \cup S_2)$ *satisfying* $\mathcal{P}_{S_2}(A) \neq 0_p$.

The systematic definition of conditional plausibility distributions given in Proposition 3 fits with the usual definitions of conditional distributions, which are, with probabilities, "$\mathcal{P}_{S_1 \mid S_2}(A) = \mathcal{P}_{S_1,S_2}(A)/\mathcal{P}_{S_2}(A)$", with $\kappa$-rankings, "$\mathcal{P}_{S_1 \mid S_2}(A) = \mathcal{P}_{S_1,S_2}(A) - \mathcal{P}_{S_2}(A)$", and with possibility degrees combined using min, "$\mathcal{P}_{S_1 \mid S_2}(A) = \mathcal{P}_{S_1,S_2}(A)$ if $\mathcal{P}_{S_1,S_2}(A) < \mathcal{P}_{S_2}(A)$, 1 otherwise". In the following, every conditioning statement $\mathcal{P}_{S_1 \mid S_2}$ for conditionable plausibility structures will refer to the canonical notion of conditioning given in Proposition 3. Conditional independence can now be defined.

---

3. To avoid specifying that properties of $\mathcal{P}_{S_1 \mid S_2}$ hold only for assignments $A$ of $S_1 \cup S_2$ satisfying $\mathcal{P}_{S_2}(A) \neq 0_p$, we use expressions such as "$\mathcal{P}_{S_1 \mid S_2} = \varphi$" to denote "$\forall A \in dom(S_1 \cup S_2)$, $(\mathcal{P}_{S_2}(A) \neq 0_p) \to (\mathcal{P}_{S_1 \mid S_2}(A) = \varphi(A))$".
4. Compared to Friedman and Halpern's conditional plausibility measures (Friedman & Halpern, 1995; Halpern, 2001), (c) is the analog of axiom (Alg1), (d) is the analog of axiom (Alg2), (e) is the analog of axiom (Alg4), and axiom (Alg3) corresponds to the distributivity of $\otimes_p$ over $\oplus_p$.

**Definition 20.** *Let $(E_p, \oplus_p, \otimes_p)$ be a conditionable plausibility structure. Let $\mathcal{P}_S$ be a plausibility distribution over $S$ and $S_1, S_2, S_3$ be disjoint subsets of $S$. $S_1$ is said to be conditionally independent of $S_2$ given $S_3$, denoted $I(S_1, S_2 \,|\, S_3)$, iff $\mathcal{P}_{S_1,S_2 \,|\, S_3} = \mathcal{P}_{S_1 \,|\, S_3} \otimes_p \mathcal{P}_{S_2 \,|\, S_3}$.*

This means that $S_1$ is conditionally independent of $S_2$ given $S_3$, iff the problem can be split into one part depending on $S_1$ and $S_3$, and another part depending on $S_2$ and $S_3$.[5] This definition satisfies the usual properties of conditional independence, as Proposition 4 shows.

**Proposition 4.** *$I(.,.\,|\,.)$ satisfies the semigraphoid axioms:*

1. *symmetry: $I(S_1, S_2 \,|\, S_3) \rightarrow I(S_2, S_1 \,|\, S_3)$,*
2. *decomposition: $I(S_1, S_2 \cup S_3 \,|\, S_4) \rightarrow I(S_1, S_2 \,|\, S_4)$,*
3. *weak union: $I(S_1, S_2 \cup S_3 \,|\, S_4) \rightarrow I(S_1, S_2 \,|\, S_3 \cup S_4)$,*
4. *contraction: $(I(S_1, S_2 \,|\, S_4) \wedge I(S_1, S_3 \,|\, S_2 \cup S_4)) \rightarrow I(S_1, S_2 \cup S_3 \,|\, S_4)$.*

Proposition 4 makes it possible to use Bayesian network techniques to express information in a compact way. With Bayesian networks, a DAG of variables is used to represent conditional independences between the variables (Pearl, 1988). In some cases, such as image processing and statistical physics, it is more natural to express conditional independences between sets of variables. If probabilities are used, such situations can be modeled using *chain graphs* (Frydenberg, 1990). In a chain graph, the DAG defined is not a DAG of variables, but a DAG of sets of variables, called components. Conditional probability distributions $P_{x \,|\, pa_G(x)}$ of variables are replaced by conditional probability distributions $P_{c \,|\, pa_G(c)}$ of components, each $P_{c \,|\, pa_G(c)}$ being expressed in a factored form $\varphi_1^c \times \varphi_2^c \times \ldots \times \varphi_{k_c}^c$. *Markov random fields* (Chellappa & Jain, 1993) correspond to the case in which there is a unique component equal to $V$, and in which the factored form of $P_V$ looks like $1/Z \times \prod_{j \in J} e^{H_j}$ (Gibbs distribution).

We now formally introduce DAGs over sets of variables, called *DAGs of components*, and then use them to factor plausibility distributions.

**Definition 21.** *A DAG $G$ is said to be a* DAG of components *over a set of variables $S$ iff the vertices of $G$ form a partition of $S$. $\mathcal{C}(G)$ denotes the set of components of $G$. For each $c \in \mathcal{C}(G)$, $pa_G(c)$ denotes the set of variables included in the parents of $c$ in $G$, and $nd_G(c)$ denotes the set of variables included in the non-descendant components of $c$ in $G$.*

**Definition 22.** *Let $(E_p, \oplus_p, \otimes_p)$ be a conditionable plausibility structure. Let $\mathcal{P}_S$ be a plausibility distribution over $S$ and let $G$ be a DAG of components over $S$. $G$ is said to be* compatible *with $\mathcal{P}_S$ iff $I(c, nd_G(c) - pa_G(c) \,|\, pa_G(c))$ for all $c \in \mathcal{C}(G)$ ($c$ is conditionally independent of its non-descendants given its parents).*

5. Definition 20 differs from Halpern's, which is "$S_1$ is conditionally independent (CI) of $S_2$ given $S_3$ iff $\mathcal{P}_{S_1 \,|\, S_2,S_3} = \mathcal{P}_{S_1 \,|\, S_3}$ and $\mathcal{P}_{S_2 \,|\, S_1,S_3} = \mathcal{P}_{S_2 \,|\, S_3}$". Halpern (2001) called the definition we adopt non-interactivity (NI) and showed that NI is weaker than CI. This implies that NI is satisfied more often and may lead to more factorizations. Halpern gave a simple axiom (axiom (Alg4')) under which CI and NI are equivalent. Though this axiom holds in many usual frameworks, it does not hold with possibility degrees combined using min, a case covered by the PFU algebraic structure.

**Theorem 1.** *(Conditional independence and factorization) Let* $(E_p, \oplus_p, \otimes_p)$ *be a conditionable plausibility structure and let* $G$ *be a DAG of components over* $S$*.*

*(a) If* $G$ *is compatible with a plausibility distribution* $\mathcal{P}_S$ *over* $S$*, then* $\mathcal{P}_S = \otimes_{p\, c \in \mathcal{C}(G)} \mathcal{P}_{c \,|\, pa_G(c)}$.

*(b) If, for all* $c \in \mathcal{C}(G)$*, there is a function* $\varphi_{c,pa_G(c)}$ *such that* $\varphi_{c,pa_G(c)}(A)$ *is a plausibility distribution over* $c$ *for all assignments* $A$ *of* $pa_G(c)$*, then* $\gamma_S = \otimes_{p\, c\in\mathcal{C}(G)} \varphi_{c,pa_G(c)}$ *is a plausibility distribution over* $S$ *with which* $G$ *is compatible.*

Theorem 1 links conditional independence and factorization. Theorem 1(a) is a generalization of the usual result of Bayesian networks (Pearl, 1988) which says that if a DAG of variables is compatible with a probability distribution $P_S$, then $P_S$ can be factored as $P_S = \prod_{x\in S} P_{x \,|\, pa_G(x)}$. Theorem 1(b) is a generalization of the standard result of Bayesian networks (Pearl, 1988) which says that, given a DAG $G$ of variables in $S$, if conditional probabilities $P_{x \,|\, pa_G(x)}$ are defined for each variable $x \in S$, then $\prod_{x\in S} P_{x \,|\, pa_G(x)}$ defines a probability distribution over $S$ with which $G$ is compatible. Both results are generalizations since they hold for arbitrary plausibility distributions (and not for probability distributions only). Results similar in spirit are provided by Halpern (2001), who gives some conditions under which a plausibility measure can be represented by a Bayesian network.

Theorem 1(a) entails that, in order to factor a global plausibility distribution $\mathcal{P}_S$, it suffices to define a DAG of components compatible with it, i.e. to express conditional independences. To define such a DAG, the following systematic procedure can be used. The initial DAG of components is an empty DAG $G$. While $\mathcal{C}(G) = \{c_1, \ldots, c_{k-1}\}$ is not a partition of $S$, do:

1. Let $S_k = c_1 \cup \ldots \cup c_{k-1}$; choose a subset $c_k$ of the set $S - S_k$ of variables not already considered.

2. Add $c_k$ as a component to $G$ and find a minimal subset $pa_k$ of $S_k$ such that $I(c_k, S_k - pa_k \,|\, pa_k)$. Add edges directed from components containing at least one variable in $pa_k$ to $c_k$, so that $pa_G(c_k) = \cup_{(c\in\{c_1,\ldots,c_{k-1}\})/(c\cap pa_k \neq \emptyset)} c$.

The resulting DAG of components is guaranteed to be compatible with $\mathcal{P}_S$, which implies, using Theorem 1(a), that the local functions $P_i$ representing $\mathcal{P}_S$ can simply be defined as the functions in the set $\{\mathcal{P}_{c \,|\, pa_G(c)}, c \in \mathcal{C}(G)\}$. In practice, if there is a reasonable notion of causes and effects, then networks that are smaller or somehow easier to build can be obtained by using the following two heuristics in order to choose $c_k$ at each step of the procedure above:

(R1) *Consider causes before effects*: in the dinner problem, this suggests not putting $ep_J$ in $c_k$ if its causes $bp_J$ and $w$ are not in $S_k$.

(R2) *Gather in a component variables that are correlated even when all variables in* $S_k$ *are assigned*: $bp_J$ and $bp_M$ are correlated and (R1) does not apply. Indeed, we cannot say that $bp_J$ has a causal influence on $bp_M$, or that $bp_M$ has a causal influence on $bp_J$, since which of Mary or John chooses first if (s)he baby-sits is not specified. We can even assume that $bp_J$ and $bp_M$ are correlated via an unmodeled common cause, such as a coin toss that determines the baby-sitter. Hence, $bp_J$ and $bp_M$ can be put in the same component $c = \{bp_J, bp_M\}$.[6]

We say that (R1) and (R2) build a DAG respecting causality. They must be seen just as *possible mechanisms* that help in identifying conditional independences by using the notions of causes and effects.

All the previous results extending Bayesian networks results to plausibility distributions also apply to feasibilities. Indeed, the feasibility structure $S_f = (\{t, f\}, \vee, \wedge)$ is a particular case of a conditionable plausibility structure, since it satisfies the axioms of Definition 19. We may therefore speak of conditional feasibility distribution. If $S$ is a set of decision variables, the construction of a DAG compatible with a feasibility distribution $\mathcal{F}_S$ leads to the factorization $\mathcal{F}_S = \wedge_{c \in \mathcal{C}(G)} \mathcal{F}_{c \,|\, pa_G(c)}$.

### 5.2.2 Taking the Differenty Types of Variables into Account

The material defined in the previous subsection enables us to factor one plausibility distribution $\mathcal{P}_{V_E}$ defined over the set $V_E$ of environment variables and one feasibility distribution $\mathcal{F}_{V_D}$ defined over the set $V_D$ of decision variables. However, dealing with just one plausibility distribution over $V_E$ and one feasibility distribution over $V_D$ is not sufficient.

Indeed, for plausibilities, decision variables can influence the environment (for example, the health state of a patient depends on the treatment chosen for him by a doctor). Rather than expressing one plausibility distribution over $V_E$, we want to express a family of plausibility distributions over $V_E$, one for each assignment of $V_D$. To make this clear, we define *controlled plausibility distributions.*

**Definition 23.** *A plausibility distribution over $V_E$ controlled by $V_D$ (or just a* controlled plausibility distribution*), denoted $\mathcal{P}_{V_E \,||\, V_D}$, is a function $dom(V_E \cup V_D) \rightarrow E_p$, such that for all assignments $A_D$ of $V_D$, $\mathcal{P}_{V_E \,||\, V_D}(A_D)$ is a plausibility distribution over $V_E$.*

For feasibilities, it goes the other way around: the values of environment variables can constrain the possible decisions (for example, an unmanned aerial vehicle which is flying cannot take off). Thus, we want to express a family of feasibility distributions over $V_D$, one for each assignment of $V_E$. In other words, we want to express a controlled feasibility distribution $\mathcal{F}_{V_D \,||\, V_E}$.

In order to directly reuse Theorem 1 for controlled distributions, we introduce the notion of the completion of a controlled distribution. This allows us to extend a distribution to the full set of variables $V$ by assigning the same plausibility (resp. feasibility) degree to every assignment of $V_D$ (resp. $V_E$), and to work with only one plausibility (resp. feasibility) distribution.

6. Components such as $\{bp_J, bp_M\}$ could be broken by assuming for example that $bp_M$ causally influences $bp_J$, i.e. that Mary chooses if she baby-sits first. We can (and prefer to) keep the component as $\{bp_J, bp_M\}$ because, in general, "breaking" components can increase the scopes of the functions involved. For example, assume that we want to model plausibilities over variables representing colors of pixels of an $N \times N$ image, such that the color of a pixel probabilistically depends on the colors of its 4 neighbors only. With a component approach, results of Markov random fields (Chellappa & Jain, 1993) show that the local functions obtained have scopes of size 5 only, whereas with a component-breaking mechanism, the size of the largest scope is linear in $N$.

**Proposition 5.** *Let $(E_p, \oplus_p, \otimes_p)$ be a conditionable plausibility structure. Then, for all $n \in \mathbb{N}^*$, there exists a unique $p_0$ such that $\oplus_{p_{i\in\{1,\dots,n\}}} p_0 = 1_p$.*

**Definition 24.** *Let $(E_p, \oplus_p, \otimes_p)$ be a conditionable plausibility structure and let $\mathcal{P}_{V_E \,||\, V_D}$ be a controlled plausibility distribution. Then, the* completion of $\mathcal{P}_{V_E \,||\, V_D}$ *is a function denoted $\mathcal{P}_{V_E,V_D}$, defined by $\mathcal{P}_{V_E,V_D} = \mathcal{P}_{V_E \,||\, V_D} \otimes_p p_0$, where $p_0$ is the unique element of $E_p$ such that $\oplus_{p_{i\in[1,|dom(V_D)|]}} p_0 = 1_p$ (the cardinality of a set $\Gamma$ is denoted $|\Gamma|$).*

In other words, $\mathcal{P}_{V_E,V_D}$ is defined from $\mathcal{P}_{V_E \,||\, V_D}$ by assigning the same plausibility degree $p_0$ to all assignments of $V_D$. In the case of probability theory, it corresponds to saying that the assignments of $V_D$ are equiprobable. The definition of the completion of a controlled plausibility distribution could be made more flexible: instead of defining a uniform plausibility distribution over $V_D$, we could define a plausibility distribution such that no assignment of $V_D$ has $0_p$ as a plausibility degree. We arbitrarily choose the uniform distribution, the goal being just to introduce some prior plausibilities over decision variables, for the sake of factorization.

**Proposition 6.** *Let $\mathcal{P}_{V_E,V_D}$ be the completion of a controlled plausibility distribution $\mathcal{P}_{V_E \,||\, V_D}$. Then, $\mathcal{P}_{V_E,V_D}$ is a plausibility distribution over $V_E \cup V_D$ and $\mathcal{P}_{V_E \,|\, V_D} = \mathcal{P}_{V_E \,||\, V_D}$.*

As a result, we use $\mathcal{P}_{V_E \,|\, V_D}$ to denote $\mathcal{P}_{V_E \,||\, V_D}$ (and this is equivalent). Similarly, it is possible to complete a controlled feasibility distribution $\mathcal{F}_{V_D \,||\, V_E}$.

#### 5.2.3 A First Factorization

Proposition 7 below, entailed by Theorem 1(a), shows how to obtain a first factorization of $\mathcal{P}_{V_E \,|\, V_D}$ and $\mathcal{F}_{V_D \,|\, V_E}$.

**Definition 25.** *A DAG $G$ is a* typed DAG of components over $V_E \cup V_D$ *iff the vertices of $G$ form a partition of $V_E \cup V_D$ such that each element of this partition is a subset of either $V_D$ or $V_E$. Each vertex of $G$ is called a* component*. The set of components contained in $V_E$ (environment components) is denoted $\mathcal{C}_E(G)$ and the set of components contained in $V_D$ (decision components) is denoted $\mathcal{C}_D(G)$.*

**Proposition 7.** *Let $G$ be a typed DAG of components over $V_E \cup V_D$. Let $G_p$ be the partial graph of $G$ induced by the arcs of $G$ incident to environment components. Let $G_f$ be the partial graph of $G$ induced by the arcs of $G$ incident to decision components. If $G_p$ is compatible with the completion of $\mathcal{P}_{V_E \,||\, V_D}$ (cf. Definition 22) and $G_f$ is compatible with the completion of $\mathcal{F}_{V_D \,||\, V_E}$, then*

$$\mathcal{P}_{V_E \,|\, V_D} = \bigotimes_{p\; c\in\mathcal{C}_E(G)} \mathcal{P}_{c \,|\, pa_G(c)} \text{ and } \mathcal{F}_{V_D \,|\, V_E} = \bigwedge_{c\in\mathcal{C}_D(G)} \mathcal{F}_{c \,|\, pa_G(c)}.$$

This allows us to specify local $P_i$ and $F_i$ functions: it suffices to express each $\mathcal{P}_{c \,|\, pa_G(c)}$ and each $\mathcal{F}_{c \,|\, pa_G(c)}$ to express $\mathcal{P}_{V_E \,|\, V_D}$ and $\mathcal{F}_{V_D \,|\, V_E}$ in a compact way. In fact, we could have defined two DAGs, one for the factorization of $\mathcal{P}_{V_E \,|\, V_D}$ and the other for the factorization of $\mathcal{F}_{V_D \,|\, V_E}$, but these two DAGs can actually always be merged as soon as we make the (undemanding) assumption that it is impossible, given $x \in V_D$ and $y \in V_E$, that both $x$ influences $y$ and $y$ constrains the possible decision values for $x$. This assumption ensures that the union of the two DAGs does not create cycles. We use just one DAG for simplicity.

**Example.** *Consider the dinner problem to illustrate the first factorization step. One way to obtain $G$ is to use the causality-based reasoning described after Theorem 1. We start with an empty DAG. As $ep_J$ and $ep_M$ are both effects of $bp_J$, $bp_M$, $w$, or $mc$, they are not considered in the first component $c_1$. $bp_J$ can be chosen as a variable to add to $c_1$, because we cannot say that $bp_J$ is necessarily an effect of another variable. As previously explained, $bp_J$ can be a cause of $bp_M$, or an effect of $bp_M$, or $bp_J$ may be correlated with $bp_M$ via an unmodeled cause. As a result, we get $c_1 = \{bp_J, bp_M\}$ as a first component. Obviously, $c_1$ has no parents in the DAG because it is the first added component.*

*Then, as $ep_J$ and $ep_M$ are effects of $w$ or $mc$, we do not consider $ep_J$ or $ep_M$ in the second component $c_2$. Since $w$ is not necessarily an effect of $mc$, we can add $w$ to $c_2$. The dinner problem specifies that ordering fish and red wine simultaneously is not feasible, but we do not know whether the wine is chosen before or after the main course, i.e. $w$ can be a cause or an effect of $mc$. As a result, we take $c_2 = \{mc, w\}$. As the menu choice is independent from who is present at the beginning, $c_2$ has no parent in the temporary DAG.*

*As $ep_J$ is a direct effect of $bp_J$ and $w$ only (John leaves the dinner if white wine is chosen), we can add $ep_J$ to $c_3$. Moreover, $ep_J$ is not correlated with $ep_M$ when $c_1 \cup c_2 = \{bp_J, bp_M, mc, w\}$ is assigned. Therefore, we take $c_3 = \{ep_J\}$. Given that $ep_J$ depends both on $bp_J$ and $w$, $c_3$ gets $\{bp_J, bp_M\}$ and $\{mc, w\}$ as parents. Finally, $c_4 = \{ep_M\}$, and as $ep_M$ depends on $bp_M$ and $mc$ (Mary leaves if meat is chosen) and is independent of $ep_J$ given $bp_M$ and $mc$, we have that $I(\{ep_M\}, \{ep_J, bp_J, w\} \,|\, \{bp_M, mc\})$. This entails that $\{ep_M\}$ is added to the DAG with $\{bp_J, bp_M\}$ and $\{mc, w\}$ as parents. Therefore, we get $\mathcal{C}_D(G) = \{\{mc, w\}\}$ as the set of decision components and $\mathcal{C}_E(G) = \{\{bp_J, bp_M\}, \{ep_J\}, \{ep_M\}\}$ as the set of environment components. The DAG of components is shown in Figure 1a.*

*Using Proposition 7, we know that the joint probability distribution factors as $\mathcal{P}_{V_E \,|\, V_D} = \mathcal{P}_{bp_J, bp_M} \times \mathcal{P}_{ep_J \,|\, bp_J, bp_M, mc, w} \times \mathcal{P}_{ep_M \,|\, bp_J, bp_M, mc, w}$ and that the joint feasibility distribution can be factored as $\mathcal{F}_{V_D \,|\, V_E} = \mathcal{F}_{mc,w}$.*

#### 5.2.4 Further Factorization Steps

Proposition 7 provides us with a decomposition of $\mathcal{P}_{V_E \,|\, V_D}$ and $\mathcal{F}_{V_D \,|\, V_E}$ based on the conditional independence relation $I(.,.\,|\,.)$ of Definition 20. It may be possible to perform further factorization steps by factoring each $\mathcal{P}_{c \,|\, pa_G(c)}$ as a set of local plausibility functions $P_i$ and factoring each $\mathcal{F}_{c \,|\, pa_G(c)}$ as a set of local feasibility functions $F_i$.

- In some cases, expressing factors of $\mathcal{P}_{c \,|\, pa_G(c)}$ or $\mathcal{F}_{c \,|\, pa_G(c)}$ is quite natural. For example, if $\otimes_p = \wedge$, if variables in an environment component $c = \{x_{i,j} \,|\, i, j \in [1, n]\}$ without parents represent pixel colors, and if we want to model in $\mathcal{P}_c$ that two adjacent pixels have different colors, it is natural to define a set of binary difference constraints $\delta_{x_{i,j}, x_{k,l}}$ and to factor $\mathcal{P}_c$ as $\mathcal{P}_c = \left(\wedge_{i \in [1,n-1]} \wedge_{j \in [1,n]} \delta_{x_{i,j}, x_{i+1,j}}\right) \wedge \left(\wedge_{i \in [1,n]} \wedge_{j \in [1,n-1]} \delta_{x_{i,j}, x_{i,j+1}}\right)$. Such a decomposition cannot be obtained based only on the conditional independence relation $I(.,.\,|\,.)$ of Definition 20.

- In some settings, as in Markov random fields (Chellappa & Jain, 1993), systematic techniques exist to obtain such factorizations. The Bayesian network community also offers systematic techniques: with hybrid networks (Dechter & Larkin, 2001), we can extract the deterministic information contained in conditional probability distributions. More precisely, a conditional probability distribution $P_{x \,|\, pa_G(x)}$ can be expressed as $P_{x\,|\,pa_G(x)} = P_{x\,|\,pa_G(x)} \times \Gamma$, where $\Gamma$ is the 0-1 function defined by $\Gamma(A) = \begin{cases} 0 \text{ if } P_{x\,|\,pa_G(x)}(A) = 0 \\ 1 \text{ otherwise} \end{cases}$. The factorization of $P_{x\,|\,pa_G(x)}$ as $P_{x\,|\,pa_G(x)} \times \Gamma$ can be computationally relevant when constraint propagation techniques on 0-1 functions are used to solve hybrid networks.

- We may use another weaker definition of conditional independence: in valuation-based systems (Shenoy, 1994), $S_1$ and $S_2$ are said to be conditionally independent given $S_3$ with regard to a function $\gamma_{S_1,S_2,S_3}$ if this function factors into two scoped functions with scopes $S_1 \cup S_3$ and $S_2 \cup S_3$. This definition is not used for the first factorization step because it destroys the normalization conditions which may be useful from a computational point of view.

These additional factorization steps are of interest because decreasing the size of the scopes of the functions involved or adding redundant information in the problem can be computationally useful.

For every environment component $c$, if "$P_i \in Fact(c)$" stands for "$P_i$ is a factor of $\mathcal{P}_{c\,|\,pa_G(c)}$", the second factorization gives us

$$\mathcal{P}_{c\,|\,pa_G(c)} = \underset{P_i \in Fact(c)}{\otimes_p} P_i.$$

As $\oplus_{p_c} \mathcal{P}_{c\,|\,pa_G(c)} = 1_p$, the $P_i$ functions in $Fact(c)$ satisfy the normalization condition $\oplus_{p_c}(\otimes_{p_{P_i \in Fact(c)}} P_i) = 1_p$. Their scopes $sc(P_i)$ are contained in $sc(\mathcal{P}_{c\,|\,pa_G(c)}) = c \cup pa_G(c)$.

For every decision component $c$, if "$F_i \in Fact(c)$" stands for "$F_i$ is a factor of $\mathcal{F}_{c\,|\,pa_G(c)}$", the second factorization gives us

$$\mathcal{F}_{c\,|\,pa_G(c)} = \underset{F_i \in Fact(c)}{\wedge} F_i.$$

Given that $\vee_c \mathcal{F}_{c\,|\,pa_G(c)} = t$, the $F_i$ functions in $Fact(c)$ satisfy the normalization condition $\vee_c \left(\wedge_{F_i \in Fact(c)} F_i\right) = t$. Moreover, $sc(F_i) \subseteq c \cup pa_G(c)$.

Other factorizations, which do not decrease the scopes of the functions involved, could also be exploited. Indeed, each scoped function $P_i$ or $F_i$ can itself have an internal *local structure*, as for instance when $P_i$ is a noisy-OR gate (Pearl, 1988) in a Bayesian network, or in presence of context-specific independence (Boutilier, Friedman, Goldszmidt, & Koller, 1996). Such internal local structures can be made explicit by representing functions with tools such as Algebraic Decision Diagrams (Bahar, Frohm, Gaona, Hachtel, Macii, Pardo, & Somenzi, 1993). In the rest of the article, we do not make any assumption on the way each scoped function is represented.

**Example.** *$\mathcal{P}_{bp_J,bp_M}$ can be expressed in terms of a first plausibility function $P_1$ specifying the probability of John and Mary being present at the beginning. $P_1$ is defined by $P_1((bp_J,t).(bp_M,f)) = 0.6$, $P_1((bp_J,f).(bp_M,t)) = 0.4$, and $P_1((bp_J,t).(bp_M,t)) = P_1((bp_J,f).(bp_M,f)) = 0$. We can also add redundant deterministic information with a second plausibility function $P_2$ defined as the constraint $bp_J \neq bp_M$ ($P_2(A) = 1$ if the constraint is satisfied, 0 otherwise). We get $\mathcal{P}_{bp_J,bp_M} = P_1 \otimes_p P_2$ and $Fact(\{bp_J, bp_M\}) = \{P_1, P_2\}$.*

$\mathcal{P}_{ep_J \mid bp_J, bp_M, mc, w}$ *can be specified as a combination of two plausibility functions* $P_3$ *and* $P_4$*.* $P_3$ *expresses that if John is absent at the beginning, he is absent at the end:* $P_3$ *is the hard constraint* $(bp_J = f) \rightarrow (ep_J = f)$ *(*$P_3(A) = 1$ *if the constraint is satisfied,* 0 *otherwise). Then,* $P_4 : (bp_J = t) \rightarrow ((ep_J = t) \leftrightarrow (w \neq white))$ *is a hard constraint specifying that John leaves iff white wine is chosen. Hence, we have* $\mathcal{P}_{ep_J \mid bp_J, bp_M, mc, w} = P_3 \otimes_p P_4$ *and* $Fact(\{ep_J\}) = \{P_3, P_4\}$*. Similarly,* $\mathcal{P}_{ep_M \mid bp_J, bp_M, mc, w} = P_5 \otimes_p P_6$*, with* $P_5$*,* $P_6$ *defined as constraints, and* $Fact(\{ep_M\}) = \{P_5, P_6\}$*.*

*As for feasibilities,* $\mathcal{F}_{mc,w}$ *can be specified by a feasibility function* $F_1$ *expressing that ordering fish with red wine is not allowed:* $F_1 : \neg((mc = fish) \wedge (w = red))$ *and* $Fact(\{mc, w\}) = \{F_1\}$*. The association of local functions with components appears in Figure 1a.*

### 5.3 Local Utilities

Local utilities can be defined over the states of the environment only (as in the utility of the health state of a patient), over decisions only (as in the utility of the decision of buying a car or not), or over the states of the environment and decisions (as in the utility of the result of a horse race and a bet on the race).[7]

In order to specify local utilities, one standard approach, used in CSPs and influence diagrams, is to directly define a set $U$ of local utility functions, modeling preferences or hard requirements, over decision and environment variables. This set implicitly defines a global utility $\mathcal{U}_V = \otimes_{u U_i \in U} U_i$ over all variables. If this factored form is obtained from a global joint utility, one may rely, when $\otimes_u = +$, on the work of Fishburn (1982) and Bacchus-Grove (1995), who introduced a notion of conditional independence for utilities.

No normalization condition is imposed on local utilities.

**Example.** *In the dinner problem, three local utility functions can be defined. A binary utility function* $U_1$ *expresses that Peter does not want John to leave the dinner:* $U_1$ *is the hard constraint* $(bp_J = t) \rightarrow (ep_J = t)$ *(*$U_1(A) = 0$ *if the constraint is satisfied,* $-\infty$ *otherwise). Two unary utility functions* $U_2$ *and* $U_3$ *over* $ep_J$ *and* $ep_M$ *respectively express the gains expected from the presences at the end:* $U_2((ep_J, t)) = 10$ *and* $U_2((ep_J, f)) = 0$ *(John invests \$*$10K$ *if he is present at the end), while* $U_3((ep_M, t)) = 50$ *and* $U_3((ep_M, f)) = 0$ *(Mary invests \$*$50K$ *if she is present at the end).* $U_2$ *and* $U_3$ *can be viewed as soft constraints. All the local functions are represented in a graphical model in Figure 1b.*

### 5.4 Formal Definition of PFU Networks

We can now formally define Plausibility-Feasibility-Utility networks. The definition is justified by the previous construction process, but it holds even if the plausibility structure is not conditionable.

7. In influence diagrams, special nodes called value nodes are introduced to represent the outcome of decisions, and one utility function is associated with each of these value nodes (the utility of the outcome). In the PFU framework, we directly represent such utility functions as scoped functions which hold on the parents of value nodes. This explicitly express that utility functions are scoped functions, just as plausibility and feasibility functions. In other words, utility functions are directly utilities of the outcome of decision and environment variables assignments.

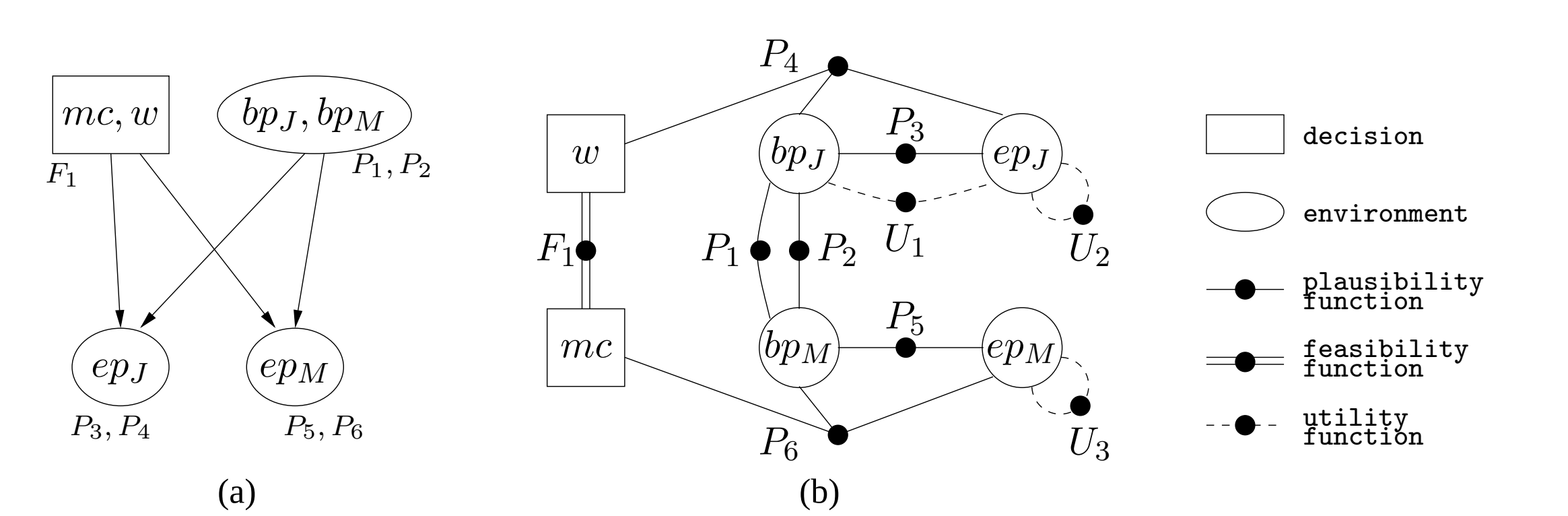

**Figure 1:** (a) DAG of components (b) Network of scoped functions.

**Definition 26.** *A* Plausibility-Feasibility-Utility network *on an expected utility structure is a tuple* $\mathcal{N} = (V, G, P, F, U)$ *such that the following conditions hold:*

- $V = \{x_1, x_2, \ldots\}$ *is a finite set of finite domain* variables. $V$ *is partitioned into* $V_D$ *(decision variables) and* $V_E$ *(environment variables);*
- $G$ *is a typed DAG of components over* $V_E \cup V_D$ *(cf. Definition 25);*
- $P = \{P_1, P_2, \ldots\}$ *is a finite set of* plausibility functions. *Each* $P_i \in P$ *is associated with a unique component* $c \in \mathcal{C}_E(G)$ *such that* $sc(P_i) \subseteq c \cup pa_G(c)$. *The set of* $P_i \in P$ *associated with a component* $c \in \mathcal{C}_E(G)$ *is denoted* $Fact(c)$ *and must satisfy* $\bigoplus_{p\,c}(\otimes_{p\,P_i \in Fact(c)} P_i) = 1_p$*;*
- $F = \{F_1, F_2, \ldots\}$ *is a finite set of* feasibility functions. *Each function* $F_i$ *is associated with a unique component* $c \in \mathcal{C}_D(G)$ *such that* $sc(F_i) \subseteq c \cup pa_G(c)$. *The set of* $F_i \in F$ *associated with a component* $c \in \mathcal{C}_D(G)$ *is denoted* $Fact(c)$ *and must satisfy* $\bigvee_c \left(\wedge_{F_i \in Fact(c)} F_i\right) = t$*;*
- $U = \{U_1, U_2, \ldots\}$ *is a finite set of* utility functions.

### 5.5 From PFU Networks to Global Functions

We have seen how to obtain a PFU network expressing a global controlled plausibility distribution $\mathcal{P}_{V_E \,||\, V_D}$, a global controlled feasibility distribution $\mathcal{F}_{V_D \,||\, V_E}$, and a global utility $\mathcal{U}_V$. Conversely, let $\mathcal{N} = (V, G, P, F, U)$ be a PFU network, i.e. a set of variables, a typed DAG of components, and sets of scoped functions. Then

- the global function $\Psi = \otimes_{p\,P_i \in P} P_i$ is a controlled plausibility distribution of $V_E$ given $V_D$. Moreover, by Theorem 1(b), if the plausibility structure is conditionable and if $G_p$ is the partial DAG of $G$ induced by the arcs incident to environment components, then $G_p$ is compatible with the completion of $\Psi$;
- the global function $\Phi = \wedge_{F_i \in F} F_i$ is a controlled feasibility distribution of $V_D$ given $V_E$. Moreover, by Theorem 1(b), if $G_f$ is the partial DAG of $G$ induced by the arcs of $G$ incident to decision components, then $G_f$ is compatible with the completion of $\Phi$;

- $\mu = \otimes_{u U_i \in U} U_i$ is necessarily a global utility.

We can therefore denote $\Psi$ by $\mathcal{P}_{V_E \,||\, V_D}$, $\Phi$ by $\mathcal{F}_{V_D \,||\, V_E}$, and $\mu$ by $\mathcal{U}_V$.

### 5.6 Back to Existing Frameworks

Let us consider the formalisms described in Section 3 again.

- A CSP (hard or soft) can easily be represented as a PFU network $\mathcal{N} = (V, G, \emptyset, \emptyset, U)$: all variables in $V$ are decision variables, $G$ is reduced to a single decision component containing all variables, and constraints are represented by utility functions. Using feasibility functions to represent constraints, it would be impossible to represent inconsistent networks because of the normalization conditions on feasibilities. SAT is modeled similarly; the only difference is that constraints are replaced by clauses.

- The same PFU network as above is used to represent the local functions of a quantified boolean formula or of a quantified CSP. The differences with CSPs or SAT appear when we consider queries on the network (see Section 6).

- A Bayesian network can be modeled as $\mathcal{N} = (V, G, P, \emptyset, \emptyset)$: all variables in $V$ are environment variables, $G$ is the DAG of the BN, and $P = \{P_{x \,|\, pa_G(x)}, x \in V\}$. There is no feasibility or utility function. A chain graph is also modeled as $\mathcal{N} = (V, G, P, \emptyset, \emptyset)$, with $G$ the DAG of components of the chain graph and $P$ the set of factors of each $P_{c \,|\, pa_G(c)}$.

- A stochastic CSP is represented by a PFU network $\mathcal{N} = (V, G, P, \emptyset, U)$, where $V$ is partitioned into $V_D$, the set of decision variables, and $V_E$, the set of stochastic variables, $G$ is a DAG which depends on the relations between the stochastic variables, $P$ is the set of probability distributions over the stochastic variables, and $U$ is the set of constraints.

- An influence diagram can be modeled by $\mathcal{N} = (V, G, P, \emptyset, U)$ such that $V_D$ contains the decision variables, $V_E$ contains the chance variables, $G$ is the DAG of the influence diagram without the utility nodes and with arcs into random variables only (i.e. we keep only the so-called *influence* arcs), and $P = \{P_{x \,|\, pa_G(x)}, x \in V_E\}$. There are no feasibilities, and one utility function $U_i$ is defined per utility variable $u$, the scope of $U_i$ being $pa_G(u)$. To represent valuation networks, a set $F$ of feasibility functions is added. Note that the business dinner example could not have been modeled using a standard influence diagram, since influence diagrams cannot model feasibilities (suitable extensions exist however, Shenoy, 2000).

- A finite horizon probabilistic MDP can be modeled as $\mathcal{N} = (V, G, P, F, U)$. If there are $T$ time-steps, then $V_D = \{d_t, t \in [1, T]\} \cup \{s_1\}$ and $V_E = \{s_t, t \in [2, T]\}$;[8] $G$ is a DAG of components such that (a) each component contains one variable, (b) the unique parent of a decision component $\{d_t\}$ is $\{s_t\}$, (c) the parents of an environment component $\{s_{t+1}\}$ are $\{s_t\}$ and $\{d_t\}$; $P = \{P_{s_{t+1}|s_t, d_t}, t \in [1, T-1]\}$, $F = \{F_{d_t \,|\, s_t}, t \in [1, T]\}$, and $U = \{R_{s_t, d_t}, t \in [1, T]\}$. Modeling a finite horizon possibilistic MDP is similar.

8. As there is no plausibility distribution over the initial state $s_1$, $s_1$ is not viewed as an environment variable. This corresponds to the special case where decision variables model problem parameters.

### 5.7 Summary

In this section, we have introduced the second element of the PFU framework: a network of variables linked by local plausibility, feasibility, and utility functions, with a DAG capturing normalization conditions. The factorization of global plausibilities, feasibilities, and utilities into scoped functions was linked to conditional independence.

## 6. Queries on a PFU Network

A query will correspond to a reasoning task on the information expressed by a PFU network. If decision variables are involved in the PFU network considered, answering a query may provide decision rules. Examples of informal queries about the dinner problem are

1. "What is the best menu choice if Peter does not know who is present at the beginning?"
2. "What is the best menu choice if Peter knows who is present at the beginning?"
3. "How should we maximize the expected investment if the restaurant chooses the main course first and Peter is pessimistic about this choice, then who is present at the beginning is observed, and last Peter chooses the wine?"

Dissociating PFU networks from queries is consistent with the trend in the influence diagram community to relax the so-called *information links*, as in Unconstrained Influence Diagrams (Jensen & Vomlelova, 2002) or Limited Memory Influence Diagrams (Lauritzen & Nilsson, 2001): it explains the intuition that queries do not change the local relations between variables.

In this section, we define a simple class of queries on PFU networks. We assume that a *sequence of decisions* must be performed, and that the order in which decisions and observations are made is known. We also make a *no-forgetting* assumption, that is, when making a decision, an agent is aware of all previous decisions and observations. From now on, the set of utility degrees $E_u$ is assumed to be *totally* ordered. This total order assumption, which holds in most of the standard frameworks, implies that there always exists an optimal decision rule. See Subsection 6.7 for a discussion of how to extend the results to a partial order.

Two definitions of the answer to a query are given, the first based on decision trees, and the second more operational. An equivalence between these two definitions is then established.

### 6.1 Query Definition

In order to formulate reasoning tasks on a PFU network, we use a sequence $Sov$ of operator-variable(s) pairs. This sequence captures different aspects of the query:

- *Partial observabilities*: $Sov$ specifies the order in which decisions are made and environment variables are observed. If $x \in V_E$ appears to the left of $y \in V_D$ (for example $Sov = \ldots (\oplus_u, \{x\}) \ldots (\max, \{y\}) \ldots$), this means that the value of $x$ is known (observed) when a value for $y$ is chosen. Conversely, $Sov = \ldots (\max, \{y\}) \ldots (\oplus_u, \{x\}) \ldots$ if $x$ is not observed when choosing $y$.

- *Optimistic/pessimistic attitude* concerning the decision makers: $(\max, \{y\})$ is inserted in the elimination sequence if the decision maker is optimistic about the behavior of the agent controlling a decision variable $y$ (i.e. if this agent is cooperative), and $(\min, \{y\})$ if one is pessimistic (i.e. if the agent controlling $y$ is an adversary). The operator used for environment variables will always be $\oplus_u$, to model that expected utilities are sought.[9]

- *Parameters of the decision making problem*: the set $S$ of variables that are not involved in $Sov$ are kind of parameters. Their absence indicates that we want to obtain optimal expected utilities and/or optimal policies for each assignment of $S$. This is useful in order to evaluate several scenarios simultaneously.

**Example.** *The sequence corresponding to the informal query: "How should we maximize the expected investment if the restaurant chooses the main course first and Peter is pessimistic about this choice, then those present at the beginning of the dinner are observed, and last Peter chooses the wine before knowing who is present at the end?" is*

$$Sov = (\min, \{mc\}).(\oplus_u, \{bp_J, bp_M\}).(\max, \{w\}).(\oplus_u, \{ep_J, ep_M\}).$$

*It models the fact that: (1) Peter is pessimistic about the main course (*min *over $mc$), which is chosen without observing any variable (no variable to the left of $mc$ in $Sov$); (2) Peter makes the best choice of wine (*max *over $w$) after the main course has been chosen and after knowing who is present at the beginning ($w$ appears to the right of $mc$, $bp_J$, and $bp_M$ in $Sov$), but before knowing who is present at the end ($w$ appears to the left of $ep_J, ep_M$). Specifically, $bp_J$ and $bp_M$ are partially observable, whereas $ep_J$ and $ep_M$ are unobservable. If the query becomes "What should Peter do if he observes those present at the beginning of the dinner and then chooses the wine before knowing who is present at the end?", then the sequence to use is $Sov = (\oplus_u, \{bp_J, bp_M\}).(\max, \{w\}).(\oplus_u, \{ep_J, ep_M\})$. In this case, variable $mc$ does not appear in the sequence anymore, which means that $mc$ is a parameter and that an answer for each value of this parameter is sought.*

**Definition 27.** *A query on a PFU network is a pair $Q = (\mathcal{N}, Sov)$ where $\mathcal{N}$ is a PFU network and $Sov = (op_1, S_1) \cdot (op_2, S_2) \cdots (op_k, S_k)$ is a sequence of* operator-set of variables *pairs such that*

*(1) all the $S_i$ are disjoint;*

*(2) either $S_i \subseteq V_D$ and $op_i = \min$ or $\max$, or $S_i \subseteq V_E$ and $op_i = \oplus_u$;*

*(3) variables not involved in any of the $S_i$, called* free variables*, are decision variables;*

*(4) for all variables $x, y$ of different types (one is a decision variable, the other is an environment variable), if there is a directed path from the component which contains $x$ to the component which contains $y$ in the DAG of the PFU network $\mathcal{N}$, then $x$ does not appear to the right of $y$ in $Sov$, i.e. either $x$ appears to the left of $y$, or $x$ is a free variable.*

9. If a decision is made by nature and if we have a plausibility distribution on that decision, then this decision will be viewed as an environment variable.

Condition (1) ensures that each variable is eliminated at most once. Condition (2) means that optimal decisions are sought for decision variables (either maximized if the decision maker who controls a decision variable is cooperative, or minimized if he is adversarial), whereas expected utilities are sought for environment variables. Condition (3) means that variables which are not eliminated in $Sov$ act as problem parameters and are viewed as decision variables. Condition (4) means that if $x$ and $y$ are of different types and $x$ is an ancestor of $y$, then $x$ is assigned before $y$. This ensures that causality is respected for variables of different types: for example, $(\mathcal{N}, (\oplus_u, \{bp_J, bp_M, ep_J, ep_M\}).(\max, \{mc, w\}))$, which violates condition (4), violates causality since the menu cannot be chosen after knowing who is present at the end.

Variables appearing in $Sov$ are called *quantified variables*, by analogy with quantified boolean formulas. The set of free variables is denoted by $V_{fr}$. Notice that the definition of queries does not prevent an environment variable from being "quantified" by min or max, because we may have $\oplus_u = \min$ or $\oplus_u = \max$. Note also that it is straightforward that for every PFU network $\mathcal{N}$, there exists at least one query on $\mathcal{N}$ without free variables.

For all $i \in [1, k]$, we define

- the set $l(S_i)$ of variables appearing in $V_{fr}$ or to the left of $S_i$ in $Sov$ by $l(S_i) = V_{fr} \cup (\cup_{j \in [1,i-1]} S_j)$;
- the set $r(S_i)$ of variables appearing to the right of $S_i$ in $Sov$ by $r(S_i) = \cup_{j \in [i+1,k]} S_j$.

### 6.2 Semantic Answer to a Query

In this subsection, we assume that the plausibility structure is conditionable (cf. Definition 19). The controlled plausibility distribution $\mathcal{P}_{V_E \,||\, V_D} = \otimes_{p P_i \in P} P_i$ can then be completed (cf. Definition 24) to give a plausibility distribution $\mathcal{P}_{V_E, V_D}$ over $V_E \cup V_D$. Similarly, the controlled feasibility distribution $\mathcal{F}_{V_D \,||\, V_E} = \wedge_{F_i \in F} F_i$ can be completed to give a feasibility distribution $\mathcal{F}_{V_E, V_D}$ over $V_E \cup V_D$. We also use the global utility $\mathcal{U}_V = \otimes_{u U_i \in U} U_i$ defined by the PFU network.

Imagine that we want to answer the query $Q = (\mathcal{N}, Sov)$, where $\mathcal{N}$ is the network of the dinner problem and $Sov = (\min, \{mc\}).(\oplus_u, \{bp_J, bp_M\}).(\max, \{w\}).(\oplus_u, \{ep_J, ep_M\})$.

To answer such a query, we can use a decision tree. First, the restaurant chooses the worst possible main course, taking into account the feasibility distribution of $mc$. Here, $\mathcal{F}_{mc}((mc, meat)) = \mathcal{F}_{mc,w}((mc, meat).(w, white)) \vee \mathcal{F}_{mc,w}((mc, meat).(w, red)) = t \vee t = t$. Similarly, $\mathcal{F}_{mc}((mc, fish)) = t$. Both choices are feasible. Then, if $A_1$ denotes the assignment of $mc$, the uncertainty over those present at the beginning given the main course choice is described by the probability distribution $\mathcal{P}_{bp_J, bp_M \,|\, mc}(A_1)$. For each possible assignment $A_2$ of $\{bp_J, bp_M\}$, i.e. for each $A_2$ such that $\mathcal{P}_{bp_J, bp_M \,|\, mc}(A_1.A_2) \neq 0_p$, Peter chooses the best wine while taking into account the feasibility $\mathcal{F}_{w \,|\, mc, bp_J, bp_M}(A_1.A_2)$: if the restaurant chooses meat, Peter chooses an optimal value between red and white, and if the restaurant chooses fish, Peter can choose white wine only. Then, for each feasible assignment $A_3$ of $w$, the uncertainty regarding the presence of John and Mary at the end of the dinner is given by $\mathcal{P}_{ep_J, ep_M \,|\, bp_J, bp_M, mc, w}(A_1.A_2.A_3)$.

Note that the conditional probabilities used in the decision tree above are not directly defined by the network. They must be computed from the global distributions. This computation can be a challenge in large problems.

The utility $\mathcal{U}_V(A_1.A_2.A_3.A_4)$ can be associated with each possible complete assignment $A_1.A_2.A_3.A_4$ of the variables. For each possible assignment $A_1.A_2.A_3$ of $\{bp_J, bp_M, mc, w\}$, the last stage, i.e. the one in which $ep_J$ and $ep_M$ are assigned, can be seen as a *lottery* (von Neumann & Morgenstern, 1944) whose expected utility is $\sum_{A_4 \in dom(\{ep_J, ep_M\})} p(A_4) \times u(A_4)$, where $p(A_4) = \mathcal{P}_{ep_J, ep_M \,|\, bp_J, bp_M, mc, w}(A_1.A_2.A_3.A_4)$ and $u(A_4) = \mathcal{U}_V(A_1.A_2.A_3.A_4)$. This expected utility becomes the reward of the scenario over $\{bp_M, bp_J, mc, w\}$ described by $A_1.A_2.A_3$. It provides us with a criterion for choosing an optimal value for $w$. The step in which $bp_J$ and $bp_M$ are assigned can then be seen as a lottery, which provides us with a criterion for choosing a worst value for $mc$. The computation associated with the previously described process is:

$$\min_{A_1 \in dom(mc), \mathcal{F}_{mc}(A_1)=t} \Big( \sum_{A_2 \in dom(\{bp_J, bp_M\}), \mathcal{P}_{bp_J, bp_M \,|\, mc}(A_1.A_2) \neq 0} \mathcal{P}_{bp_J, bp_M \,|\, mc}(A_1.A_2) \times \Big( \max_{A_3 \in dom(w), \mathcal{F}_{w \,|\, mc, bp_J, bp_M}(A_1.A_2.A_3)=t} \Big( \sum_{\substack{A_4 \in dom(\{ep_J, ep_M\}) \\ \mathcal{P}_{ep_J, ep_M \,|\, bp_J, bp_M, mc, w}(A_1.A_2.A_3.A_4) \neq 0}} \mathcal{P}_{ep_J, ep_M \,|\, bp_J, bp_M, mc, w}(A_1.A_2.A_3.A_4) \times \mathcal{U}_V(A_1.A_2.A_3.A_4) \Big) \Big) \Big).$$

Decision rules for the decision variables (argmin and argmax) can be recorded during the computation. This formulation represents the decision process as a decision tree in which each internal level corresponds to variable assignments. Arcs associated with the assignment of a set of decision variables are weighted by the feasibility of the decision given the previous assignments. Arcs associated with the assignment of environment variables are weighted by the plausibility degree of the assignment given the previous assignments. Leaf nodes correspond to the utilities of complete assignments, and a node collects the values of its children to compute its own value.

#### 6.2.1 Formalization of the Decision Tree Procedure

In order to formalize the decision tree procedure, some technical results are first introduced in Proposition 8. These results and the definitions preceding them can be skipped for a first reading.

**Definition 28.** *Let $\mathcal{P}_{S_1 \,|\, S_2}$ be the conditional plausibility distribution of $S_1$ given $S_2$ and let $A \in dom(S_2)$. The function $\mathcal{P}_{S_1 \,|\, S_2}(A)$ is said to be well-defined iff $\mathcal{P}_{S_2}(A) \neq 0_p$. Similarly, if $\mathcal{F}_{S_1 \,|\, S_2}$ is the conditional feasibility distribution of $S_1$ given $S_2$, then, for all $A \in dom(S_2)$, $\mathcal{F}_{S_1 \,|\, S_2}(A)$ is said to be well-defined iff $\mathcal{F}_{S_2}(A) = t$.*

Next, the conditioning can be defined directly for controlled plausibility distributions because for all $A \in dom(V_D)$, $\mathcal{P}_{V_E \,||\, V_D}(A)$ is a plausibility distribution over $V_E$:

**Definition 29.** *Assume that the plausibility structure used is conditionable. Let $\mathcal{P}_{V_E \,||\, V_D}$ be a controlled plausibility distribution and $S, S'$ be two disjoint subsets of $V_E$. We define conditional controlled plausibility distributions by: for all $A \in dom(S \cup S' \cup V_D)$ such that $\mathcal{P}_{S' \,||\, V_D}(A) \neq 0_p$, $\mathcal{P}_{S \,|\, S' \,||\, V_D}(A) = \max\{p \in E_p \,|\, \mathcal{P}_{S,S' \,||\, V_D}(A) = p \otimes_p \mathcal{P}_{S' \,||\, V_D}(A)\}$, as in the canonical definition of conditioning given in Proposition 3. Given a controlled feasi-*

*bility distribution* $\mathcal{F}_{V_D \,||\, V_E}$*, the definition of conditional controlled feasibility distributions* $\mathcal{F}_{S \,|\, S' \,||\, V_E}$ *for* $S, S'$ *disjoint subsets of* $V_D$ *is similar.*

**Proposition 8.** *Assume that the plausibility structure used is conditionable. Let* $Q = (\mathcal{N}, Sov)$ *be a query where* $Sov = (op_1, S_1) \cdot (op_2, S_2) \cdots (op_k, S_k)$*. Let* $V_{fr}$ *denote the set of free variables of* $Q$*.*

*(1) If* $S_i \subseteq V_E$ *and* $\mathcal{P}_{S_i \,|\, l(S_i)}(A)$ *is well-defined, then there exists at least one* $A' \in dom(S_i)$ *satisfying* $\mathcal{P}_{S_i \,|\, l(S_i)}(A.A') \neq 0_p$*.*

*(2) If* $S_i \subseteq V_D$ *and* $\mathcal{F}_{S_i \,|\, l(S_i)}(A)$ *is well-defined, then there exists at least one* $A' \in dom(S_i)$ *satisfying* $\mathcal{F}_{S_i \,|\, l(S_i)}(A.A') = t$*.*

*(3) If* $V_E \neq \emptyset$ *and* $S_i$ *is the leftmost set of environment variables appearing in Sov, then, for all* $A \in dom(l(S_i))$*,* $\mathcal{P}_{S_i \,|\, l(S_i)}(A)$ *is well-defined.*

*(4) If* $i, j \in [1, k]$*,* $i < j$*,* $S_i \subseteq V_E$*,* $S_j \subseteq V_E$*,* $r(S_i) \cap l(S_j) \subseteq V_D$ *(*$S_j$ *is the first set of environment variables appearing to the right of* $S_i$ *in Sov),* $(A, A') \in dom(l(S_i)) \times dom(S_i)$*,* $\mathcal{P}_{S_i \,|\, l(S_i)}(A)$ *is well-defined, and* $\mathcal{P}_{S_i \,|\, l(S_i)}(A.A') \neq 0_p$*, then, for all* $A''$ *extending* $A.A'$ *over* $l(S_j)$*,* $\mathcal{P}_{S_j \,|\, l(S_j)}(A'')$ *is well-defined.*

*(5) If* $i, j \in [1, k]$*,* $i < j$*,* $S_i \subseteq V_D$*,* $S_j \subseteq V_D$*,* $r(S_i) \cap l(S_j) \subseteq V_E$ *(*$S_j$ *is the first set of decision variables appearing to the right of* $S_i$ *in Sov),* $(A, A') \in dom(l(S_i)) \times dom(S_i)$*,* $\mathcal{F}_{S_i \,|\, l(S_i)}(A)$ *is well-defined, and* $\mathcal{F}_{S_i \,|\, l(S_i)}(A.A') = t$*, then, for all* $A''$ *extending* $A.A'$ *over* $l(S_j)$*,* $\mathcal{F}_{S_j \,|\, l(S_j)}(A'')$ *is well-defined.*

*(6) For all* $i \in [1, k]$ *such that* $S_i \subseteq V_E$*,* $\mathcal{P}_{S_i \,|\, l(S_i)} = \mathcal{P}_{S_i \,|\, l(S_i) \cap V_E \,||\, V_D}$*.*

*(7) For all* $i \in [1, k]$ *such that* $S_i \subseteq V_D$*,* $\mathcal{F}_{S_i \,|\, l(S_i)} = \mathcal{F}_{S_i \,|\, l(S_i) \cap V_D \,||\, V_E}$*.*

The technical results of Proposition 8 ensure that, in the following semantic answer to a query (see Definition 30),

- all quantities $\mathcal{P}_{S \,|\, l(S)}(A.A')$ and $\mathcal{F}_{S \,|\, l(S)}(A.A')$ used are defined (thanks to items 3 to 5 in Proposition 8);
- all eliminations over restricted domains are defined because the restricted domains used are never empty (items 1 and 2 in Proposition 8);
- the conditional distributions used coincide with a conditioning defined directly from the controlled plausibility and feasibility distributions $\mathcal{P}_{V_E \,||\, V_D}$ and $\mathcal{F}_{V_D \,||\, V_E}$ (items 6 and 7 in Proposition 8). This is useful because this guarantees that $\mathcal{P}_{S \,|\, l(S)}(A.A')$ and $\mathcal{F}_{S \,|\, l(S)}(A.A')$, which *a priori* require the notion of completion to be written, are actually independent of the notion of completion, which is arbitrarily added to the basic information expressed in a PFU network. We use $\mathcal{P}_{S \,|\, l(S)}$ and $\mathcal{F}_{S \,|\, l(S)}$ instead of conditional controlled distributions $\mathcal{P}_{S \,|\, l(S) \cap V_E \,||\, V_D}$ and $\mathcal{F}_{S \,|\, l(S) \cap V_D \,||\, V_E}$ for notation convenience and to explicitly represent that $\mathcal{P}_{S \,|\, l(S) \cap V_E \,||\, V_D}$ and $\mathcal{F}_{S \,|\, l(S) \cap V_D \,||\, V_E}$ do not depend on the assignment of $V_D - l(S)$ and $V_E - l(S)$ respectively.

**Definition 30.** *The semantic answer Sem-Ans(Q) to a query $Q = (\mathcal{N}, Sov)$ is a function of the set $V_{fr}$ of free variables of Q defined by*[10]

$$Sem\text{-}Ans(Q)(A) = \begin{cases} \Diamond \text{ if } \mathcal{F}_{V_{fr}}(A) = f \\ Qs(\mathcal{N}, Sov, A) \text{ otherwise,} \end{cases}$$

*with Qs inductively defined by:*

$$(1) \quad Qs(\mathcal{N}, \emptyset, A) = \mathcal{U}_V(A)$$

$$(2) \quad Qs(\mathcal{N}, (op, S) . Sov, A) =$$

$$\begin{cases} \min\limits_{\substack{A' \in dom(S) \\ \mathcal{F}_{S|l(S)}(A.A') = t}} Qs(\mathcal{N}, Sov, A.A') & \text{if } (S \subseteq V_D) \wedge (op = \min), \\ \max\limits_{\substack{A' \in dom(S) \\ \mathcal{F}_{S|l(S)}(A.A') = t}} Qs(\mathcal{N}, Sov, A.A') & \text{if } (S \subseteq V_D) \wedge (op = \max), \\ \bigoplus\limits_{\substack{u \\ A' \in dom(S) \\ \mathcal{P}_{S|l(S)}(A.A') \neq 0_p}} \left(\mathcal{P}_{S|l(S)}(A.A') \otimes_{pu} Qs(\mathcal{N}, Sov, A.A')\right) & \text{if } (S \subseteq V_E). \end{cases}$$

In other words, each step involving decision variables (first two cases) corresponds to an optimization step among the feasible choices, and each step involving environment variables (third case) corresponds to a lottery (von Neumann & Morgenstern, 1944) such that the rewards are the $Qs(\mathcal{N}, Sov, A.A')$, and such that the plausibility attributed to a reward is $\mathcal{P}_{S\,|\,l(S)}(A.A')$ (the formula looking like $\oplus_{ui}(p_i \otimes_{pu} u_i)$ is the expected utility of this lottery). When a set of decision variables $S$ is eliminated, a decision rule for $S$ can be recorded, using an argmax (resp. an argmin) if max (resp. min) is performed.

**Example.** *What is the maximum investment Peter can expect, and which associated decision(s) should he make if he chooses the menu without knowing who will attend? To answer this question, we can use a query in which $bp_J$, $bp_M$, $ep_J$, and $ep_M$ are eliminated before mc and w to represent the fact that their values are not known when the menu is chosen. This query is:*

$$(\mathcal{N}, (\max, \{mc, w\}).(\oplus_u, \{bp_J, bp_M, ep_J, ep_M\})).$$

*The answer is \$6K, with (mc, meat).(w, red) as a decision. If Peter knows who comes, the query becomes*

$$(\mathcal{N}, (\oplus_u, \{bp_J, bp_M\}).(max, \{mc, w\}).(\oplus_u, \{ep_J, ep_M\})).$$

*and optimal values for mc and w can depend on $bp_J$ and $bp_M$. The answer is \$26K with a \$20K gain from the observability of who is present. The decision rule for $\{mc, w\}$ is (mc, meat).(w, red) if John is present and Mary is not, (mc, fish).(w, white) otherwise. Consider the query introduced at the beginning of Section 6.1:*

$$(\mathcal{N}, (\min, \{mc\}).(\oplus_u, \{bp_J, bp_M\}).(\max, \{w\}).(\oplus_u, \{ep_J, ep_M\})).$$

*The answer is $-\infty$: in the worst main course case, even if Peter chooses the wine, the situation can be unacceptable. In order to compute the expected utility for each menu choice, we can use a query in which mc and w are free variables:*

10. $\Diamond$ is the unfeasible value, cf. Definition 6.

$$(\mathcal{N}, (\oplus_u, \{bp_J, bp_M, ep_J, ep_M\})).$$

*The answer is a function of* $\{mc, w\}$*. These examples show how queries can capture various situations in terms of partial observabilities, optimistic/pessimistic attitude, and parameters in the decision process.*

### 6.3 Operational Answer to a Query

The quantities $\mathcal{P}_{S\,|\,l(S)}(A.A')$ and $\mathcal{F}_{S\,|\,l(S)}(A.A')$ involved in the definition of the semantic answer to a query are not directly available from the local functions and can be very expensive to compute. For instance, with probabilities, $\mathcal{P}_{S\,|\,l(S)}(A.A')$ equals $\mathcal{P}_{S,l(S)}(A.A')/\mathcal{P}_{l(S)}(A)$. Computing $\mathcal{P}_{S,l(S)}(A.A') = \sum_{A''\in dom(V-(S\cup l(S)))} \mathcal{P}_{V_E,V_D}(A.A'.A'')$ typically requires time exponential in $|V - (S \cup l(S))|$. Moreover, such quantities must be computed at each node of the decision tree. Fortunately, there exists an alternative definition of the answer to a query, which can be directly expressed using a PFU instance, i.e. the expressed local plausibility, feasibility, and utility functions.

**Definition 31.** *The operational answer* $Op\text{-}Ans(Q)$ *to a query* $Q = (\mathcal{N}, Sov)$ *is a function of the free variables of* $Q$*: if* $A$ *is an assignment of the free variables, then* $(Op\text{-}Ans(Q))(A)$ *is defined inductively as follows:*

$$(Op\text{-}Ans(Q))(A) = Qo\,(\mathcal{N}, Sov, A)$$

$$Qo(\mathcal{N}, (op, S)\,.\,Sov, A) = op_{A'\in dom(S)}\, Qo\,(\mathcal{N}, Sov, A.A') \tag{9}$$

$$Qo(\mathcal{N}, \emptyset, A) = \left(\left(\bigwedge_{F_i\in F} F_i\right) \star \left(\underset{P_i\in P}{\otimes_p} P_i\right) \otimes_{pu} \left(\underset{U_i\in U}{\otimes_u} U_i\right)\right)(A). \tag{10}$$

By Equation 10, if all the problem variables are assigned, the answer to the query is the combination of the plausibility degree, the feasibility degree, and the utility degree of the corresponding complete assignment. By Equation 9, if the variables are not all assigned and $(op, S)$ is the leftmost operator-variable(s) pair in $Sov$, the answer to the query is obtained by eliminating $S$ using $op$ as an elimination operator. Again, optimal decision rules for the decision variables can be recorded if needed, using argmin and argmax. Equivalently, by considering a sequence of operator-variable(s) pairs as a sequence of variable eliminations, $Op\text{-}Ans(Q)$ can be written:

$$Op\text{-}Ans(Q) = Sov\left(\left(\bigwedge_{F_i\in F} F_i\right) \star \left(\underset{P_i\in P}{\otimes_p} P_i\right) \otimes_{pu} \left(\underset{U_i\in U}{\otimes_u} U_i\right)\right).$$

It shows that $Op\text{-}Ans(Q)$ actually corresponds to the generic form of Equation 8.

### 6.4 Equivalence Theorem

Theorem 2 proves that the semantic definition $Sem\text{-}Ans(Q)$ gives semantic foundations to what is computed with the operational definition $Op\text{-}Ans(Q)$.

**Theorem 2.** *If the plausibility structure is conditionable, then, for all queries* $Q$ *on a PFU network,* $Sem\text{-}Ans(Q) = Op\text{-}Ans(Q)$ *and the optimal policies for the decisions are the same with* $Sem\text{-}Ans(Q)$ *and* $Op\text{-}Ans(Q)$*.*

In other words, Theorem 2 shows that it is possible to perform computations in a completely generic algebraic framework, while providing the result of the computations with decision-theoretic foundations. Due to this equivalence theorem, *Op-Ans*$(Q)$ is denoted simply by *Ans*$(Q)$ in the following. Note that the operational definition applies even in a non-conditionable plausibility structure. Giving a decision-theoretic-based semantics to *Op-Ans* when the plausibility structure is not conditionable is an open issue.

### 6.5 Bounded Queries

It may be interesting to relax the problem of computing the exact answer to a query. Assume that the leftmost operator-variable(s) pair in the sequence *Sov* is $(\max, \{x\})$, with $x$ a decision variable. From the decision maker point of view, computing decision rules providing an expected utility greater than a given threshold $\theta$ may be sufficient. This is the case with the E-MAJSAT problem, defined as "*Given a boolean formula over a set of variables* $V = V_D \cup V_E$*, does there exist an assignment of* $V_D$ *such that the formula is satisfied for at least half of the assignments of* $V_E$*?*" Extending the generic PFU framework to answer such queries is done in Definitions 32 and 33, which introduce bounded queries.

**Definition 32.** *A bounded query B-Q is a triple* $(\mathcal{N}, Sov, \theta)$*, such that* $(\mathcal{N}, Sov)$ *is a query and* $\theta \in E_u$ *(*$\theta$ *is the threshold).*

**Definition 33.** *The answer Ans(B-Q) to a bounded query B-Q* $= (\mathcal{N}, Sov, \theta)$ *is a boolean function of the free variables of the "unbounded" query* $Q = (\mathcal{N}, Sov)$*. For every assignment A of these free variables,*

$$(Ans(B\text{-}Q))(A) = \begin{cases} t & \text{if } Ans(Q)(A) \succeq_u \theta \\ f & \text{otherwise.} \end{cases}$$

As the threshold $\theta$ may be used to prune the search space during the resolution, computing the answer to a bounded query is easier than computing the answer to an unbounded one.

### 6.6 Back to Existing Frameworks

Let us consider again the frameworks of Section 3. Solving a CSP (Equation 1) or a totally ordered soft CSP corresponds to the query $Q = (\mathcal{N}, (\max, V))$, with $\mathcal{N}$ the PFU network corresponding to the CSP and $V$ the set of variables of the CSP. Computing the probability distribution of a variable $y$ for a Bayesian network (Equation 2) modeled as $\mathcal{N}$ corresponds to $Q = (\mathcal{N}, (+, V - \{y\})$. These examples are *mono-operator queries*, involving only one type of elimination operator.

Consider *multi-operator queries*. The search for an optimal policy for the stochastic CSP associated with Equation 4 is captured by a query such as $Q = (\mathcal{N}, (\max, \{d_1, d_2\}) .(+, \{s_1\}).(\max, \{d_3, d_4\}).(+, \{s_2\}))$. The query on influence diagrams of Equation 5 and the query on valuation networks of Equation 6 are captured the same way.

For a finite horizon MDP with $T$ time-steps (Equation 7), the query looks like $Q = (\mathcal{N}, (\max, \{d_1\}).(\oplus_u, \{s_2\}).(\max, \{d_2\}) \ldots (\oplus_u, \{s_T\}).(\max, \{d_T\}))$, where $\oplus_u = +$ with probabilistic MDPs and $\oplus_u = \min$ with pessimistic possibilistic MDPs. The initial state $s_1$ is a free variable. With a quantified CSP or a quantified boolean formula, elimination operators min and max are used to represent ∀ and ∃.

More formally, we can show:

**Theorem 3.** *Queries and bounded queries can be used to express and solve the following list of problems:*

1. *SAT framework: SAT, MAJSAT, E-MAJSAT, quantified boolean formula, stochastic SAT (SSAT) and extended-SSAT (Littman et al., 2001).*
2. *CSP (or CN) framework:*
   - *Check consistency for a CSP (Mackworth, 1977); find a solution to a CSP; count the number of solutions of a CSP.*
   - *Find a solution of a valued CSP (Bistarelli et al., 1999).*
   - *Solve a quantified CSP (Bordeaux & Monfroy, 2002).*
   - *Find a conditional decision or an unconditional decision for a mixed CSP or a probabilistic mixed CSP (Fargier et al., 1996).*
   - *Find an optimal policy for a stochastic CSP or a policy with a value greater than a threshold; solve a stochastic COP (Constraint Optimization Problem) (Walsh, 2002).*
3. *Integer Linear Programming (Schrijver, 1998) with finite domain variables.*
4. *Search for a solution plan with a length $\leq k$ in a classical planning problem (STRIPS planning, Fikes & Nilsson, 1971; Ghallab et al., 2004).*
5. *Answer classical queries on Bayesian networks (Pearl, 1988), Markov random fields (Chellappa & Jain, 1993), and chain graphs (Frydenberg, 1990), with plausibilities expressed as probabilities, possibilities, or $\kappa$-rankings:*
   - *Compute plausibility distributions.*
   - *MAP (Maximum A Posteriori hypothesis) and MPE (Most Probable Explanation).*
   - *Compute the plausibility of an evidence.*
   - *CPE task for hybrid networks (Dechter & Larkin, 2001) (CPE means CNF Probability Evaluation, a CNF being a formula in Conjunctive Normal Form).*
6. *Solve an influence diagram (Howard & Matheson, 1984).*
7. *With a finite horizon, solve a probabilistic MDP, a possibilistic MDP, a MDP based on $\kappa$-rankings, completely or partially observable (POMDP), factored or not (Puterman, 1994; Monahan, 1982; Sabbadin, 1999; Boutilier et al., 1999, 2000).*

### 6.7 Towards More Complex Queries

Queries can be made more complex by relaxing some assumptions:

- In the definition of queries, the order $\preceq_u$ on $E_u$ is assumed to be total. Extending the results to a *partial* order is possible if $(E_u, \preceq_u)$ defines a lattice (partially ordered set closed under least upper and greatest lower bounds) and if $\otimes_{pu}$ distributes over the least upper bound *lub* and greatest lower bound *glb* (i.e. $p \otimes_{pu} lub(u_1, u_2) = lub(p \otimes_{pu} u_1, p \otimes_{pu} u_2)$ and $p \otimes_{pu} glb(u_1, u_2) = glb(p \otimes_{pu} u_1, p \otimes_{pu} u_2)$). This allows semiring CSPs (Bistarelli et al., 1999) to be captured in the framework. We believe that other extensions to partial orders on utilities should allow algebraic MDPs (Perny et al., 2005) to be captured.
- We can try to relax the *no-forgetting* assumption, as in limited memory influence diagrams (LIMIDs, Lauritzen & Nilsson, 2001), which show that this can be relevant for decision processes involving multiple decision makers or memory constraints on the policy recording. In such cases, optimal decisions can become *nondeterministic* (decisions such as "choose $x = 0$ with probability $p$ and $x = 1$ with probability $1-p$").
- The order in which decisions are made and environment variables are observed is total and completely determined by the query. One may wish to compute not only an optimal policy for the decisions, but also an *optimal order* in which to perform decisions, without exactly knowing the steps at which other agents make decisions or the steps at which observations are made. Work on influence diagrams with unordered decisions (Jensen & Vomlelova, 2002) is good starting point to try and extend our work in this direction.

While it should be possible to relax the assumption that variables have a *finite domain*, doing this is nontrivial, since transforming $\oplus_u = +$ into integrals is not straightforward, and performing min- or max-eliminations over continuous domains requires the guarantee of existence of a supremum.

### 6.8 Summary

In Section 6, the last element of the PFU framework, a class of queries on PFU networks, has been introduced. A decision-tree based definition of the answer to a query has been provided. The first main result of the section is Theorem 2, which gives theoretical foundations to another equivalent operational definition, reducing the answer to a query to a sequence of eliminations on a combination of scoped functions. The latter is best adapted to future algorithms, because it directly handles the local functions defined by a PFU network. The second important result is Theorem 3, which shows that many standard queries are PFU queries. Overall, the PFU framework is captured by Definitions 14, 16, and 17 for algebraic structures, Definition 26 for PFU networks, and Definitions 27 and 31 for queries.

## 7. Gains and Costs

**A better understanding** Theorem 3 shows that many existing frameworks are instances of the PFU framework. Through this unification, similarities and differences between ex-isting formalisms can be analyzed. For instance, by comparing VCSPs and the optimistic version of finite horizon possibilistic MDPs through the operational definition of the answer to a query, it appears that a finite horizon optimistic possibilistic MDP (partially observable or not) is a fuzzy CSP: both can indeed be represented as a query $Q$ whose operational answer looks like $\max_V(\min_{\varphi \in \Phi} \varphi)$, where $V$ is a set of variables and $\Phi$ is a set of scoped functions. Techniques available for solving fuzzy CSPs can then be used to solve finite horizon optimistic possibilistic MDPs.

From the complexity theory point of view, studying the time and space complexity for answering queries of the form of Equation 8 can lead to upper bounds on the complexity for several frameworks simultaneously. One may also try to characterize which properties lead to a given theoretical complexity.

**Increased expressive power** The expressive power of PFU networks is the result of a number of features: (1) flexibility of the plausibility/utility model; (2) flexibility of the possible networks; (3) flexibility of the queries in terms of situation modeling. This enables queries on PFU networks to cover generic finite horizon sequential decision problems with plausibilities, feasibilities, and utilities, cooperative or adversarial decision makers, partial observabilities, and possible parameters in the decision process modeled through free variables.

As none of the frameworks indicated in Theorem 3 presents such flexibility, for every subsumed formalism $X$ indicated in Theorem 3, it is possible to find a problem which can be represented with PFUs but not directly with $X$. More specifically, compared to influence diagrams (Howard & Matheson, 1984; Jensen & Vomlelova, 2002; Smith et al., 1993; Nielsen & Jensen, 2003; Jensen et al., 2004) or valuation networks (VNs, Shenoy, 1992, 2000; Demirer & Shenoy, 2001), PFUs can deal with more than the probabilistic expected additive utility and allow us to perform eliminations with min to model the presence of adversarial agents. Thus, quantified boolean formulas cannot be represented with influence diagrams or VNs, but are covered by PFU networks (see Theorem 3). Moreover, PFU networks use a DAG which captures the normalization conditions of plausibilities or feasibilities, whereas with VNs, this information is lost. Compared to sequential influence diagrams (Jensen et al., 2004) or sequential VNs (Demirer & Shenoy, 2001), PFUs can express some so-called *asymmetric decision problems* (problems in which some variables may not even need to be considered in a decision process) by adding dummy values to variables.

Actually, some simple problems which can be expressed with PFUs cannot be apparently directly expressed in other frameworks. The simple instance "feasibilities *with normalization conditions* + hard requirements" is not captured by any of the subsumed frameworks. For example, using a CSP to model it would result in a loss of the information provided by the normalization conditions on feasibilities. The same occurs for influence diagrams - like sequential decision processes based on possibilistic expected utility, which could be called *possibilistic influence diagrams*. Similarly for stochastic CSPs without contingency assumption.

The cost of greater flexibility and increased expressive power is that the PFU framework cannot be described as simply and straightforwardly as, for example, constraint networks.

**Generic algorithms** Section 8 shows that generic algorithms can be built to answer queries on PFU networks. As previously said, building generic algorithms should facilitate cross-fertilization in the sense that any of the subsumed formalisms will directly benefit from the techniques developed on another subsumed formalism. This fits into a growing effort to generalize resolution methods used for different AI problems. For example, soft constraint propagation drastically improves the resolution of valued CSPs; integrating such a tool in a generic algorithm on PFUs could improve the resolution of influence diagrams. Using abstract operators may enable us to identify algorithmically interesting properties, or to infer necessary or sufficient conditions for a particular algorithm to be usable.

However, one could argue that some techniques are highly specific to one formalism or to one type of problem, and that, in this case, dedicated approaches certainly outperform a generic algorithm. A solution for this can be to characterize the actual properties used by a dedicated approach, in order to generalize it as much as possible. Moreover, even if specialized schemes usually improve over generic ones, there exist cases in which general tools can be more efficient than specialized algorithms, as shown by the use of SAT solvers for solving STRIPS planning problems (Sang, Beame, & Kautz, 2005).

## 8. Algorithms

The ability to design generic algorithms is one of the motivations for building the PFU framework, and some choices are justified by algorithmic considerations. We present generic algorithms that answer arbitrary PFU queries.

### 8.1 A Generic Tree Search Algorithm

The operational definition of the answer to a query $Q$ actually defines a naive exponential time algorithm to compute $Ans(Q)$ using a tree-exploration procedure, with a variable ordering given by $Sov$, that collects elementary plausibilities, feasibilities, and utilities. More precisely, for each assignment $A$ of the free variables of $Q$, a tree is explored. Each node in this tree corresponds to a partial assignment of the variables. The value of a leaf is provided by the combination of the scoped functions of the PFU network, applied to the complete assignment defined by the path from the root to the leaf. Depending on the operator used, the value of an internal node is computed by performing a min, max, or $\oplus_u$ operation on the values of its children. The root node returns $(Ans(Q))(A)$. The corresponding pseudo-code is given in Figure 2. For a query $(\mathcal{N}, Sov)$, the first call is **TreeSearchAnswerQ**$(\mathcal{N}, Sov)$. It returns a function of the free variables.

If we assume that every operator returns a result in a constant time, then the time complexity of the algorithm is $O(m \cdot n \cdot ln(d) \cdot d^n)$, where $d$ stands for the maximum domain size, $n$ stands for the number of variables of the PFU network, and $m$ stands for the number of scoped functions.[11]

The space complexity is polynomial (it can be shown to be linear in the entry data size). Hence, computing the answer to a bounded query is *PSPACE*. Moreover, given that the satisfiability of a QBF is a PSPACE-complete problem which can be expressed as a bounded query (cf. Theorem 3), it follows that computing the answer to a bounded query is *PSPACE-hard*. Being PSPACE and PSPACE-hard, the decision problem that consists of

11. The factor $n \cdot ln(d)$ corresponds to an upper bound on the time needed to get $\varphi(A)$ for a scoped function $\varphi$ represented as a table (of size $\leq d^n$).

```
TreeSearchAnswerQ((V, G, P, F, U), Sov)
begin
    foreach A ∈ dom(V_fr) do φ(A) ← AnswerQ((V, G, P, F, U), Sov, A)
    return φ
end

AnswerQ((V, G, P, F, U), Sov, A)
begin
    if Sov = ∅ then return ((∧_{F_i∈F} F_i) ⋆ (⊗_{p P_i∈P} P_i) ⊗_pu (⊗_{u U_i∈U} U_i))(A)
    else
        (op, S).Sov' ← Sov
        choose x ∈ S
        if S = {x} then Sov ← Sov' else Sov ← (op, S − {x}).Sov'
        dom ← dom(x)
        res ← ◊
        while dom ≠ ∅ do
            choose a ∈ dom
            dom ← dom − {a}
            res ← op (res, AnswerQ((V, G, P, F, U), Sov, A.(x, a)))
        return res
end
```

**Figure 2:** A generic tree search algorithm for answering a query $Q = ((V, G, P, F, U), Sov)$

answering a bounded query is *PSPACE-complete*. This result is not surprising, but it gives an idea of the level of expressiveness which can be reached by the PFU framework. More work is needed to identify subclasses of queries with a lower complexity, although many are already known.

### 8.2 A Generic Variable Elimination Algorithm

Quite naturally, a generic variable elimination algorithm (Bertelé & Brioschi, 1972; Shenoy, 1991; Dechter, 1999; Kolhas, 2003) can be defined to answer queries on a PFU network.

#### 8.2.1 A First Naive Scheme

This first naive variable elimination algorithm is given in Figure 3. It eliminates variables from the right to the left of the sequence $Sov$ of the query, whereas with the tree search procedure, variables are assigned from the left to the right. This right-to-left processing entails that the algorithm naturally returns a function of the free variables of the query. The first call is **VarElimAnswerQ**$((V, G, P, F, U), Sov)$.

The version presented in Figure 3 is actually a very naive variable elimination scheme with time and space complexities $O(m \cdot n \cdot ln(d) \cdot d^n)$ and $O(m \cdot d^n)$ respectively: it begins by combining all the scoped functions before eliminating variables, whereas the interest of a variable elimination algorithm is primarily to use the factorization into local functions.

**VarElimAnswerQ**$((V,G,P,F,U),Sov)$
**begin**
  $\varphi_0 \leftarrow ((\wedge_{F_i \in F} F_i) \star (\otimes_{p\,P_i \in P} P_i) \otimes_{pu} (\otimes_{u\,U_i \in U} U_i))$
  **while** $Sov \neq \emptyset$ **do**
    $Sov'.(op,S) \leftarrow Sov$
    choose $x \in S$
    **if** $S = \{x\}$ **then** $Sov \leftarrow Sov'$ **else** $Sov \leftarrow Sov'.(op, S - \{x\})$
    $\varphi_0 \leftarrow op_x\, \varphi_0$
  **return** $\varphi_0$
**end**

**Figure 3:** A first generic variable elimination algorithm for answering a query $Q = ((V,G,P,F,U),Sov)$

#### 8.2.2 Improving the Basic Scheme

The algorithm of Figure 3 works on a unique global function defined by the combination of all the plausibility, feasibility, and utility functions (first line), whereas a factorization is available. To improve this scheme, the properties of the algebraic structure can be used. In the sequel, we denote by $\Phi^{+x}$ (resp. $\Phi^{-x}$) a scoped function which has (resp. does not have) $x$ in its scope. Moreover, we extend every combination operator $\otimes$ on $E \cup \{\diamondsuit\}$ by setting $\diamondsuit \otimes e = e \otimes \diamondsuit = \diamondsuit$ (combining anything with something unfeasible is unfeasible too).[12]

First, in order to use the factorization of plausibilities and feasibilities, we can use the properties below, which come from the right monotonicity of $\otimes_{pu}$, the distributivity of $\otimes_{pu}$ over $\oplus_u$, and the definition of the truncation operator $\star$:

$$\begin{cases} \min_x (P^{-x} \otimes_{pu} U) = P^{-x} \otimes_{pu} (\min_x U) \\ \max_x (P^{-x} \otimes_{pu} U) = P^{-x} \otimes_{pu} (\max_x U) \\ \oplus_{ux} (P^{-x} \otimes_{pu} U) = P^{-x} \otimes_{pu} (\oplus_{ux} U) \\ \min_x (F^{-x} \star U) = F^{-x} \star (\min_x U) \\ \max_x (F^{-x} \star U) = F^{-x} \star (\max_x U) \\ \oplus_{ux} (F^{-x} \star U) = F^{-x} \star (\oplus_{ux} U)\,. \end{cases}$$

They express that when a variable $x$ is eliminated, it is not necessary to consider plausibility functions or feasibility functions that do not have $x$ in their scope.

However, it is necessary to add some axioms on the expected utility structure, since in the general case, an expression such as $\oplus_{ux} (P^{+x} \otimes_{pu} (U^{-x} \otimes_u U^{+x}))$ cannot be decomposed. We give two axioms, $Ax1$ and $Ax2$, each of which is a sufficient additional condition to exploit the factorization of utility functions.

(Ax1) $(E_p, \preceq_p) = (E_u, \preceq_u)$, $\oplus_p = \oplus_u$, $\otimes_p = \otimes_u = \otimes_{pu}$
(Ax2) $\oplus_u = \otimes_u$ on $E_u$ (but not on $E_u \cup \{\diamondsuit\}$).

12. An operator $op$ can be used both as a combination operator between scoped functions and as an elimination operator over some variables. In this case, the extension of $op$ used as a combination operator creates an operator $op'$ such that $op'(e, \diamondsuit) = \diamondsuit$, whereas the extension of $op$ used as an elimination operator creates an operator $op''$ such that $op''(e, \diamondsuit) = e$. $op'$ and $op''$ coincide on $E$ but differ on $E \cup \{\diamondsuit\}$.

Among the cases in Table 1, rows 2, 3, 5, 6 satisfy Ax1, whereas rows 1, 4, 7, 8 satisfy Ax2. Ax1 and Ax2 enable us to write:

$$\begin{cases} \min_x (F^{+x} \star (U^{-x} \otimes_u U^{+x})) = U^{-x} \otimes_u (\min_x F^{+x} \star U^{+x}) \\ \max_x (F^{+x} \star (U^{-x} \otimes_u U^{+x})) = U^{-x} \otimes_u (\max_x F^{+x} \star U^{+x}) . \end{cases}$$

and

$$\bigoplus_{u\,x} \left(P^{+x} \otimes_{pu} (U^{-x} \otimes_u U^{+x})\right) = \begin{cases} U^{-x} \otimes_u (\oplus_{ux} P^{+x} \otimes_{pu} U^{+x}) \text{ with Ax1} \\ ((\oplus_{p_x} P^{+x}) \otimes_{pu} U^{-x}) \otimes_u (\oplus_{ux} P^{+x} \otimes_{pu} U^{+x}) \text{ with Ax2.} \end{cases}$$

Hence, when eliminating a variable $x$, it is not necessary to consider utility functions which do not have $x$ in their scope.

We present an algorithm when Ax1 is satisfied. When Ax2 holds, working on plausibility/utility pairs $(p, u)$ allows Ax1 to be recovered: this is used, for example, to solve influence diagrams (Ndilikilikesha, 1994). When Ax1 is satisfied, there is actually only one set $E = E_p = E_u$, one order $\preceq = \preceq_p = \preceq_u$, one combination operator $\otimes = \otimes_p = \otimes_u = \otimes_{pu}$, and one elimination operator $\oplus = \oplus_p = \oplus_u$. Rather than express feasibilities on $\{t, f\}$, we can express them on $\{1_E, \Diamond\}$ by mapping $t$ onto $1_E$ and $f$ onto $\Diamond$: this preserves the value of the answer to a query, since $f \star u = \Diamond \otimes u$ and $t \star u = 1_E \otimes u$.

The improved variable elimination algorithm is shown in Figure 4. To answer a query $Q = ((V, G, P, F, U), Sov)$, the first call is **Ax1-VarElimAnswerQ**$(P \cup F \cup U, Sov)$. It returns a set of scoped functions whose $\otimes$-combination equals $Ans(Q)$. This time, the factorization available in a PFU network is exploited, since when eliminating a variable $x$, only scoped functions involving $x$ are considered.

```
Ax1-VarElimAnswerQ(Φ, Sov)
begin
    if Sov = ∅ then return Φ
    else
        Sov′.(op, S) ← Sov
        choose x ∈ S
        if S = {x} then Sov ← Sov′ else Sov ← Sov′.(op, S − {x})
        Φ^{+x} ← {φ ∈ Φ | x ∈ sc(φ)}
        φ0 ← op_x (⊗_{φ∈Φ^{+x}} φ)
        Φ ← (Φ − Φ^{+x}) ∪ {φ0}
        return Ax1-VarElimAnswerQ(Φ, Sov)
end
```

**Figure 4:** Variable elimination algorithm when $Ax1$ holds ($\Phi$: set of scoped functions)

When Ax1 holds, the algorithm is actually a standard variable elimination algorithm on a commutative semiring. As for classical variable elimination algorithms, the time complexity of this algorithm is in $O(m \cdot n \cdot ln(d) \cdot d^{w+1})$, where $w$ is the tree-width (Bodlaender, 1997; Dechter & Fattah, 2001) of the network of scoped functions, constrained by the elimination order imposed by $Sov$. Yet, its space complexity is also exponential in this tree-width.

### 8.3 Other Approaches

Starting from the generic tree-search algorithm of Section 8.1, bound computations and local consistencies (Mackworth, 1977; Cooper & Schiex, 2004; Larrosa & Schiex., 2003) can be integrated in order to prune the search space. Local consistencies can improve the quality of the bounds thanks to the use of smaller local functions. Techniques coming from quantified boolean formulas or from game algorithms (such as the $\alpha\beta$-algorithm) can be considered to more efficiently manage bounds when min and max operators alternate. Caching strategies exploiting the problem structure (Darwiche, 2001; Jégou & Terrioux, 2003) are also obvious candidates to improve the basic tree search scheme. Additional axioms Ax1 and Ax2 can be useful in this direction. Heuristics for the choice of the variable to assign when a pair $(op, S)$ is encountered, as well as heuristics for the value choices, may also speed up the search.

In another direction, approximate algorithms using sampling and local search could also be considered: sampling when eliminations with $+$ ($+$, and not $\oplus_u$) are performed, local search when eliminations with min or max are performed.

## 9. Conclusion

In the last decades, AI has witnessed the design and study of numerous formalisms for reasoning about decision problems. In this article, we have built a generic framework to model sequential decision making with plausibilities, feasibilities, and utilities. This framework covers many existing approaches, including hard, valued, quantified, mixed, and stochastic CSPs, Bayesian networks, finite horizon probabilistic or possibilistic MDPs, or influence diagrams. The result is an algebraic framework built upon decision-theoretic foundations: the PFU framework. The two facets of the PFU framework are explicit in Theorem 2, which states that the operational definition of the answer to a query is equivalent to the decision tree-based semantics. This is the result of a design that accounts both for expressiveness and for computational aspects.

Compared to related works (Shenoy, 1991; Dechter, 1999; Kolhas, 2003), the PFU framework is the only framework which directly deals with different types of variables (decision and environment variables), different types of local functions (plausibilities, feasibilities, utilities), and different types of combination and elimination operators.

From an algorithmic point of view, generic algorithms based on tree search and variable elimination have been described. They prove that the PFU framework is not just an abstraction. The next step is to explore ways of improving these algorithms, so as to generalize techniques that are used in formalisms subsumed by the PFU framework. Along this line, a generic approach to query optimization has lead to the definition of original architectures for answering queries, called *multi-operator cluster trees* and *multi-operator cluster DAGs*. These can be applied to QBFs and other structures compatible with Ax1 (Pralet, Schiex, & Verfaillie, 2006a), as well as influence diagrams and other structures satisfying Ax2 (Pralet, Schiex, & Verfaillie, 2006b).

## Acknowledgments

We would like to thank Jean-Loup Farges, Jérôme Lang, Régis Sabbadin, and the three anonymous reviewers for useful comments on previous versions of this article. The work described in this article was initiated when the first author was at LAAS-CNRS and INRA Toulouse. It was partially conducted within the EU Integrated Project COGNIRON ("The Cognitive Companion") and funded by the European Commission Division FP6-IST Future and Emerging Technologies under Contract FP6-002020.

## Appendix A. Notations

See Table 2.

| *Symbol* | *Meaning* |
|---|---|
| $\oplus$ | Elimination operator |
| $\oplus_p$ | Elimination operator on plausibilities |
| $\oplus_u$ | Elimination operator on utilities |
| $\otimes$ | Combination operator |
| $\otimes_p$ | Combination operator for plausibilities |
| $\otimes_u$ | Combination operator for utilities |
| $\otimes_{pu}$ | Combination operator between plausibilities and utilities |
| $\preceq_p$ | Partial order on plausibilities |
| $\preceq_u$ | Partial order on utilities |
| $\star$ | Truncation operator |
| $\Diamond$ | Unfeasible value |
| $V_E$ | Environment variables |
| $V_D$ | Decision variables |
| $dom(x)$ | Domain of values of a variable $x$ |
| $dom(S)$ | $\prod_{x \in S} dom(x)$ |
| $G$ | Directed Acyclic Graph (DAG) |
| $pa_G(x)$ | Parents of $x$ in the DAG $G$ |
| $nd_G(x)$ | Non-descendant of $x$ in the DAG $G$ |
| $\mathcal{C}_E(G)$ | Set of environment components of $G$ |
| $\mathcal{C}_D(G)$ | Set of decision components of $G$ |
| $P_i$ | Plausibility function |
| $F_i$ | Feasibility function |
| $U_i$ | Utility function |
| $Fact(c)$ | $P_i$ or $F_i$ factors associated with a component $c$ |
| $sc(\varphi)$ | Scope of a local function $\varphi$ |
| $\mathcal{P}_S$ | Plausibility distribution over $S$ |
| $\mathcal{P}_{S_1 \mid S_2}$ | Conditional plausibility distribution of $S_1$ given $S_2$ |
| $\mathcal{F}_S$ | Feasibility distribution over $S$ |
| $\mathcal{F}_{S_1 \mid S_2}$ | Conditional feasibility distribution of $S_1$ given $S_2$ |
| $Sov$ | Sequence of operator-variable(s) pairs |
| Sem-Ans($Q$) | Semantic answer to a query $Q$ (decision trees) |
| Op-Ans($Q$) | Operational answer to a query $Q$ |
| Ans($Q$) | Answer to a query $Q$ |

Table 2: Notation.

## Appendix B. Proofs

**Proposition 1** A plausibility distribution $\mathcal{P}_S$ can be extended to give a plausibility distribution $\mathcal{P}_{S'}$ over every $S' \subset S$, defined by $\mathcal{P}_{S'} = \oplus_{p_{S-S'}} \mathcal{P}_S$.

*Proof.* Given that $\oplus_p$ is associative and commutative, $\oplus_{p_{S'}} \mathcal{P}_{S'} = \oplus_{p_{S'}} (\oplus_{p_{S-S'}} \mathcal{P}_S) = \oplus_{p_S} \mathcal{P}_S = 1_p$. Thus, $\mathcal{P}_{S'} : dom(S') \to E_p$ is a plausibility distribution over $S'$. □

**Proposition 2** The structures presented in Table 1 are expected utility structures.

*Proof.* It is sufficient to verify each of the required axioms successively. □

**Proposition 3** If $(E_p, \oplus_p, \otimes_p)$ is a conditionable plausibility structure, then all plausibility distributions are conditionable: it suffices to define $\mathcal{P}_{S_1 \,|\, S_2}$ by $\mathcal{P}_{S_1 \,|\, S_2}(A) = \max\{p \in E_p \,|\, \mathcal{P}_{S_1,S_2}(A) = p \otimes_p \mathcal{P}_{S_2}(A)\}$ for all $A \in dom(S_1 \cup S_2)$ satisfying $\mathcal{P}_{S_2}(A) \neq 0_p$.

*Proof.* Let $\mathcal{P}_S$ be a plausibility distribution over $S$. For all $S_1$, $S_2$ disjoint subsets of $S$ and for all $A \in dom(S_1 \cup S_2)$ satisfying $\mathcal{P}_{S_2}(A) \neq 0_p$, let us define $\mathcal{P}_{S_1 \,|\, S_2}(A) = \max\{p \in E_p \,|\, \mathcal{P}_{S_1,S_2}(A) = p \otimes_p \mathcal{P}_{S_2}(A)\}$. We must show that the $\mathcal{P}_{S_1 \,|\, S_2}$ functions satisfy axioms a, b, c, d, e of Definition 18.

(a) By definition of $\mathcal{P}_{S_1 \,|\, S_2}$ and by distributivity of $\otimes_p$ over $\oplus_p$, we can write

$$\mathcal{P}_{S_2} = \oplus_{p_{S_1}} \mathcal{P}_{S_1,S_2} = \oplus_{p_{S_1}} (\mathcal{P}_{S_1 \,|\, S_2} \otimes_p \mathcal{P}_{S_2}) = (\oplus_{p_{S_1}} \mathcal{P}_{S_1 \,|\, S_2}) \otimes_p \mathcal{P}_{S_2}.$$

As $\mathcal{P}_{S_2} \preceq_p \mathcal{P}_{S_2}$, we can infer that $\oplus_{p_{S_1}} \mathcal{P}_{S_1 \,|\, S_2} \preceq_p 1_p$. Let $A_2$ be an assignment of $S_2$ satisfying $\mathcal{P}_{S_2}(A_2) \neq 0_p$. Assume that the hypothesis (H): "$\oplus_{p_{S_1}} \mathcal{P}_{S_1 \,|\, S_2}(A_2) \prec_p 1_p$" holds.

Then, for all $A_1 \in dom(S_1)$, $\mathcal{P}_{S_1,S_2}(A_1.A_2) \prec_p \mathcal{P}_{S_2}(A_2)$, since if $\mathcal{P}_{S_1,S_2}(A_1.A_2) = \mathcal{P}_{S_2}(A_2)$, then $\mathcal{P}_{S_1 \,|\, S_2}(A_1.A_2) = 1_p$, which implies that $\oplus_{p_{S_1}} \mathcal{P}_{S_1 \,|\, S_2}(A_2) \succeq_p 1_p$ by the monotonicity of $\oplus_p$. Moreover, (H) implies that there exists a unique $p \in E_p$ satisfying $(\oplus_{p_{S_1}} \mathcal{P}_{S_1 \,|\, S_2}(A_2)) \oplus_p p = 1_p$. Combining this equation by $\mathcal{P}_{S_2}(A_2)$ gives $\mathcal{P}_{S_2}(A_2) \oplus_p \mathcal{P}_{S_2}(A_2) \otimes_p p = \mathcal{P}_{S_2}(A_2)$, i.e. $\mathcal{P}_{S_2}(A_2) \otimes_p (1_p \oplus_p p) = \mathcal{P}_{S_2}(A_2)$. This implies that $1_p \oplus_p p \preceq_p 1_p$. Given that $1_p \oplus_p p \succeq_p 1_p$ (by monotonicity of $\oplus_p$), we obtain $1_p \oplus_p p = 1_p$. We analyze two cases.

- If $p \prec_p 1_p$, there exists a unique $p'$ satisfying $p' \oplus_p p = 1_p$. As both $(\oplus_{p_{S_1}} \mathcal{P}_{S_1 \,|\, S_2}(A_2)) \oplus_p p = 1_p$ and $1_p \oplus_p p = 1_p$, this entails that $\oplus_{p_{S_1}} \mathcal{P}_{S_1 \,|\, S_2}(A_2) = 1_p$, which contradicts (H).
- If $p = 1_p$, then $1_p \oplus_p 1_p = 1_p$. This entails that $\oplus_p$ is idempotent. Let $dom'$ be a subset of $dom(S_1)$ such that $\oplus_{p_{A_1 \in dom'}} \mathcal{P}_{S_1,S_2}(A_1.A_2) = \mathcal{P}_{S_2}(A_2)$. Let $A'_1 \in dom'$. We can write:

$$\begin{cases} \mathcal{P}_{S_1,S_2}(A'_1.A_2) \oplus_p (\oplus_{p_{A_1 \in dom' - \{A'_1\}}} \mathcal{P}_{S_1,S_2}(A_1.A_2)) = \mathcal{P}_{S_2}(A_2) \\ \mathcal{P}_{S_1,S_2}(A'_1.A_2) \oplus_p (\oplus_{p_{A_1 \in dom'}} \mathcal{P}_{S_1,S_2}(A_1.A_2)) = \mathcal{P}_{S_2}(A_2) \text{ (as } \oplus_p \text{ is idempotent)} \end{cases}.$$

  As $\mathcal{P}_{S_1,S_2}(A'_1.A_2) \prec_p \mathcal{P}_{S_2}(A_2)$, there exists a unique $p'' \in E_p$ such that $\mathcal{P}_{S_1,S_2}(A'_1.A_2) \oplus_p p'' = \mathcal{P}_{S_2}(A_2)$. Therefore, $\oplus_{p_{A_1 \in dom'}} \mathcal{P}_{S_1,S_2}(A_1.A_2) = \oplus_{p_{A_1 \in dom' - \{A'_1\}}} \mathcal{P}_{S_1,S_2}(A_1.A_2)$, which gives $\oplus_{p_{A_1 \in dom' - \{A'_1\}}} \mathcal{P}_{S_1,S_2}(A_1.A_2) = \mathcal{P}_{S_2}(A_2)$.

  The assumption $\oplus_{p_{A_1 \in dom'}} \mathcal{P}_{S_1,S_2}(A_1.A_2) = \mathcal{P}_{S_2}(A_2)$ holds for $dom' = dom(S_1)$. Recursively applying the previous mechanism by removing one assignment in $dom'$ at each iteration leads to $\oplus_{p_{A_1 \in dom'}} \mathcal{P}_{S_1,S_2}(A_1.A_2) = \mathcal{P}_{S_2}(A_2)$ with $|dom'| = 1$, i.e. it leads to $\mathcal{P}_{S_1,S_2}(A''_1.A_2) = \mathcal{P}_{S_2}(A_2)$ with $dom' = \{A''_1\}$. As a result, we obtain a contradiction.

In both cases, a contradiction with (H) is obtained, so $\oplus_{p_{S_1}} \mathcal{P}_{S_1 \,|\, S_2}(A_2) = 1_p$.

(b) $\mathcal{P}_{S_1} = \mathcal{P}_{S_1 \,|\, \emptyset} \otimes_p \mathcal{P}_\emptyset = \mathcal{P}_{S_1 \,|\, \emptyset} \otimes_p (\oplus_{p_S} \mathcal{P}_S) = \mathcal{P}_{S_1 \,|\, \emptyset} \otimes_p 1_p = \mathcal{P}_{S_1 \,|\, \emptyset}$.

(d) Let $A \in dom(S_1 \cup S_2 \cup S_3)$ satisfying $\mathcal{P}_{S_2,S_3}(A) \neq 0_p$. Then, $\mathcal{P}_{S_1,S_2 \,|\, S_3}(A) = \mathcal{P}_{S_1 \,|\, S_2,S_3}(A) \otimes_p \mathcal{P}_{S_2 \,|\, S_3}(A)$ holds, because:

- If $\mathcal{P}_{S_1,S_2,S_3}(A) \prec_p \mathcal{P}_{S_3}(A)$, then, there exists a unique $p \in E_p$ such that $\mathcal{P}_{S_1,S_2,S_3}(A) = p \otimes_p \mathcal{P}_{S_3}(A)$. As both $\mathcal{P}_{S_1,S_2,S_3}(A) = \mathcal{P}_{S_1,S_2 \,|\, S_3}(A) \otimes_p \mathcal{P}_{S_3}(A)$ (by definition of $\mathcal{P}_{S_1,S_2 \,|\, S_3}$) and $\mathcal{P}_{S_1,S_2,S_3}(A) = \mathcal{P}_{S_1 \,|\, S_2,S_3}(A) \otimes_p \mathcal{P}_{S_2 \,|\, S_3}(A) \otimes_p \mathcal{P}_{S_3}(A)$ (by definition of $\mathcal{P}_{S_1 \,|\, S_2,S_3}$ and $\mathcal{P}_{S_2 \,|\, S_3}$), this implies that $\mathcal{P}_{S_1,S_2 \,|\, S_3}(A) = \mathcal{P}_{S_1 \,|\, S_2,S_3}(A) \otimes_p \mathcal{P}_{S_2 \,|\, S_3}(A)$.
- Otherwise, $\mathcal{P}_{S_1,S_2,S_3}(A) = \mathcal{P}_{S_3}(A)$. This implies that $1_p \preceq_p \mathcal{P}_{S_1,S_2 \,|\, S_3}(A)$ and, as $\mathcal{P}_{S_1,S_2 \,|\, S_3}(A) \preceq_p 1_p$, that $\mathcal{P}_{S_1,S_2 \,|\, S_3}(A) = 1_p$. Similarly, this entails that $\mathcal{P}_{S_2 \,|\, S_3}(A) = 1_p$ and $\mathcal{P}_{S_1 \,|\, S_2,S_3}(A) = 1_p$ (the monotonicity of $\oplus_p$ implies that $\mathcal{P}_{S_1,S_2,S_3}(A) = \mathcal{P}_{S_2,S_3}(A) = \mathcal{P}_{S_3}(A)$). As $1_p = 1_p \otimes_p 1_p$, we get $\mathcal{P}_{S_1,S_2 \,|\, S_3}(A) = \mathcal{P}_{S_1 \,|\, S_2,S_3}(A) \otimes_p \mathcal{P}_{S_2 \,|\, S_3}(A)$.

(c)
$$\begin{aligned}
\oplus_{p\,S_1} \mathcal{P}_{S_1,S_2 \,|\, S_3} &= \oplus_{p\,S_1} (\mathcal{P}_{S_1 \,|\, S_2,S_3} \otimes_p \mathcal{P}_{S_2 \,|\, S_3}) \text{ (using (d))} \\
&= (\oplus_{p\,S_1} \mathcal{P}_{S_1 \,|\, S_2,S_3}) \otimes_p \mathcal{P}_{S_2 \,|\, S_3} \text{ (because } \otimes_p \text{ distributes over } \oplus_p) \\
&= \mathcal{P}_{S_2 \,|\, S_3} \text{ (using (a))}
\end{aligned}$$

(e) Assume that $\mathcal{P}_{S_1,S_2,S_3} = \mathcal{P}_{S_1 \,|\, S_3} \otimes_p \mathcal{P}_{S_2 \,|\, S_3} \otimes_p \mathcal{P}_{S_3}$. Let $A \in dom(S_1 \cup S_2 \cup S_3)$ such that $\mathcal{P}_{S_3}(A) \neq 0_p$. Then, $\mathcal{P}_{S_1,S_2 \,|\, S_3}(A) = \mathcal{P}_{S_1 \,|\, S_3}(A) \otimes_p \mathcal{P}_{S_2 \,|\, S_3}(A)$ holds, because:

- If $\mathcal{P}_{S_1,S_2,S_3}(A) \prec_p \mathcal{P}_{S_3}(A)$, there exists a unique $p \in E_p$ such that $\mathcal{P}_{S_1,S_2,S_3}(A) = p \otimes_p \mathcal{P}_{S_3}(A)$, and therefore $\mathcal{P}_{S_1,S_2 \,|\, S_3}(A) = \mathcal{P}_{S_1 \,|\, S_3}(A) \otimes_p \mathcal{P}_{S_2 \,|\, S_3}(A)$.
- Otherwise, we can write $\mathcal{P}_{S_1 \,|\, S_3}(A) = \mathcal{P}_{S_2 \,|\, S_3}(A) = \mathcal{P}_{S_1,S_2 \,|\, S_3}(A) = 1_p$ by using reasoning similar to that of (d), and therefore $\mathcal{P}_{S_1,S_2 \,|\, S_3}(A) = \mathcal{P}_{S_1 \,|\, S_3}(A) \otimes_p \mathcal{P}_{S_2 \,|\, S_3}(A)$.

□

**Proposition 4** $I(.,.\,|\,.)$ satisfies the semigraphoid axioms:

1. symmetry: $I(S_1, S_2 \,|\, S_3) \rightarrow I(S_2, S_1 \,|\, S_3)$,
2. decomposition: $I(S_1, S_2 \cup S_3 \,|\, S_4) \rightarrow I(S_1, S_2 \,|\, S_4)$,
3. weak union: $I(S_1, S_2 \cup S_3 \,|\, S_4) \rightarrow I(S_1, S_2 \,|\, S_3 \cup S_4)$,
4. contraction: $(I(S_1, S_2 \,|\, S_4) \wedge I(S_1, S_3 \,|\, S_2 \cup S_4)) \rightarrow I(S_1, S_2 \cup S_3 \,|\, S_4)$.

*Proof.*

1. *Symmetry axiom*: directly satisfied by commutativity of $\otimes_p$.
2. *Decomposition axiom*: assume that $I(S_1, S_2 \cup S_3 \,|\, S_4)$ holds. Then
$$\begin{aligned}
\mathcal{P}_{S_1,S_2 \,|\, S_4} &= \oplus_{p\,S_3} \mathcal{P}_{S_1,S_2,S_3 \,|\, S_4} \\
&= \oplus_{p\,S_3} (\mathcal{P}_{S_1 \,|\, S_4} \otimes_p \mathcal{P}_{S_2,S_3 \,|\, S_4}) \text{ (since } I(S_1, S_2 \cup S_3 \,|\, S_4)) \\
&= \mathcal{P}_{S_1 \,|\, S_4} \otimes_p (\oplus_{p\,S_3} \mathcal{P}_{S_2,S_3 \,|\, S_4}) \text{ (by distributivity of } \otimes_p \text{ over } \oplus_p) \\
&= \mathcal{P}_{S_1 \,|\, S_4} \otimes_p \mathcal{P}_{S_2 \,|\, S_4}.
\end{aligned}$$
Thus, $I(S_1, S_2 \,|\, S_4)$ holds.
3. *Weak union axiom*: assume that $I(S_1, S_2 \cup S_3 \,|\, S_4)$ holds. The decomposition axiom entails that $I(S_1, S_3 \,|\, S_4)$ is also satisfied. Then
$$\begin{aligned}
\mathcal{P}_{S_1,S_2,S_3,S_4} &= \mathcal{P}_{S_1,S_2,S_3 \,|\, S_4} \otimes_p \mathcal{P}_{S_4} \text{ (chain rule)} \\
&= \mathcal{P}_{S_1 \,|\, S_4} \otimes_p \mathcal{P}_{S_2,S_3 \,|\, S_4} \otimes_p \mathcal{P}_{S_4} \text{ (since } I(S_1, S_2 \cup S_3 \,|\, S_4)) \\
&= \mathcal{P}_{S_1 \,|\, S_4} \otimes_p \mathcal{P}_{S_3 \,|\, S_4} \otimes_p \mathcal{P}_{S_4} \otimes_p \mathcal{P}_{S_2 \,|\, S_3,S_4} \text{ (chain rule)} \\
&= \mathcal{P}_{S_1,S_3 \,|\, S_4} \otimes_p \mathcal{P}_{S_4} \otimes_p \mathcal{P}_{S_2 \,|\, S_3,S_4} \text{ (since } I(S_1, S_3 \,|\, S_4)) \\
&= \mathcal{P}_{S_1 \,|\, S_3,S_4} \otimes_p \mathcal{P}_{S_2 \,|\, S_3,S_4} \otimes_p \mathcal{P}_{S_3,S_4} \text{ (chain rule).}
\end{aligned}$$
From axiom (e) in Definition 18, we can infer that $\mathcal{P}_{S_1,S_2 \,|\, S_3,S_4} = \mathcal{P}_{S_1 \,|\, S_3,S_4} \otimes_p \mathcal{P}_{S_2 \,|\, S_3,S_4}$, i.e. $I(S_1, S_2 \,|\, S_3 \cup S_4)$ holds.

4. *Contraction axiom*: assume that $I(S_1, S_2 \,|\, S_4)$ and $I(S_1, S_3 \,|\, S_2 \cup S_4)$ hold. Then
$$\begin{aligned}\mathcal{P}_{S_1,S_2,S_3\,|\,S_4} &= \mathcal{P}_{S_1,S_3\,|\,S_2,S_4} \otimes_p \mathcal{P}_{S_2\,|\,S_4} \text{ (chain rule)}\\ &= \mathcal{P}_{S_1\,|\,S_2,S_4} \otimes_p \mathcal{P}_{S_3\,|\,S_2,S_4} \otimes_p \mathcal{P}_{S_2\,|\,S_4} \text{ (since } I(S_1,S_3\,|\,S_2\cup S_4))\\ &= \mathcal{P}_{S_1,S_2\,|\,S_4} \otimes_p \mathcal{P}_{S_3\,|\,S_2,S_4} \text{ (chain rule)}\\ &= \mathcal{P}_{S_1\,|\,S_4} \otimes_p \mathcal{P}_{S_2\,|\,S_4} \otimes_p \mathcal{P}_{S_3\,|\,S_2,S_4} \text{ (since } I(S_1,S_2\,|\,S_4))\\ &= \mathcal{P}_{S_1\,|\,S_4} \otimes_p \mathcal{P}_{S_2,S_3\,|\,S_4} \text{ (chain rule).}\end{aligned}$$
Thus, $I(S_1, S_2 \cup S_3 \,|\, S_4)$ holds.

□

**Theorem 1** (Conditional independence and factorization) Let $(E_p, \oplus_p, \otimes_p)$ be a conditionable plausibility structure and let $G$ be a DAG of components over $S$.

(a) If $G$ is compatible with a plausibility distribution $\mathcal{P}_S$ over $S$, then $\mathcal{P}_S = \otimes_{p\,c\in\mathcal{C}(G)} \mathcal{P}_{c\,|\,pa_G(c)}$.

(b) If, for all $c \in \mathcal{C}(G)$, there is a function $\varphi_{c,pa_G(c)}$ such that $\varphi_{c,pa_G(c)}(A)$ is a plausibility distribution over $c$ for all assignments $A$ of $pa_G(c)$, then $\gamma_S = \otimes_{p\,c\in\mathcal{C}(G)} \varphi_{c,pa_G(c)}$ is a plausibility distribution over $S$ with which $G$ is compatible.

*Proof.*

(a) First, if $|\mathcal{C}(G)| = 1$, $G$ contains a unique component $c_1$. Then, $\otimes_{p\,c\in\mathcal{C}(G)} \mathcal{P}_{c\,|\,pa_G(c)} = \mathcal{P}_{c_1\,|\,\emptyset} = \mathcal{P}_{c_1}$: the proposition holds for $|\mathcal{C}(G)| = 1$.

Assume that the proposition holds for all DAGs with $n$ components. Let $G$ be a DAG of components compatible with a plausibility distribution $\mathcal{P}_S$ and such that $|\mathcal{C}(G)| = n+1$. Let $c_0$ be a component labeling a leaf of $G$. As $G$ is compatible with $\mathcal{P}_S$, we can write $I(c_0, nd_G(c_0) - pa_G(c_0) \,|\, pa_G(c_0))$. As $c_0$ is a leaf, $nd_G(c_0) = S - c_0$, and consequently $I(c_0, (S - c_0) - pa_G(c_0) \,|\, pa_G(c_0))$. This means that $\mathcal{P}_{S-pa_G(c_0)\,|\,pa_G(c_0)} = \mathcal{P}_{c_0\,|\,pa_G(c_0)} \otimes_p \mathcal{P}_{(S-c_0)-pa_G(c_0)\,|\,pa_G(c_0)}$. Combining each side of the equation by $\mathcal{P}_{pa_G(c_0)}$ gives
$$\mathcal{P}_S = \mathcal{P}_{c_0\,|\,pa_G(c_0)} \otimes_p \mathcal{P}_{S-c_0}.$$
Let $G'$ be the DAG obtained from $G$ by deleting the node labeled with $c_0$. Then for every component $c \in \mathcal{C}(G')$, $pa_{G'}(c) = pa_G(c)$ (since the deleted component $c_0$ is a leaf). Moreover $nd_{G'}(c)$ equals either $nd_G(c)$ or $nd_G(c) - c_0$ (again, since the deleted component $c_0$ is a leaf). In the first case ($nd_{G'}(c) = nd_G(c)$), the property $I(c, nd_G(c) - pa_G(c) \,|\, pa_G(c))$ directly implies $I(c, nd_{G'}(c) - pa_{G'}(c) \,|\, pa_{G'}(c))$. In the second case ($nd_{G'}(c) = nd_G(c) - c_0$), the decomposition axiom allows us to write $I(c, nd_{G'}(c) - pa_{G'}(c) \,|\, pa_{G'}(c))$ from $I(c, nd_G(c) - pa_G(c) \,|\, pa_G(c))$. Consequently, $G'$ is a DAG compatible with $\mathcal{P}_{S-c_0}$. As $|\mathcal{C}(G')| = n$, the induction hypothesis gives $\mathcal{P}_{S-c_0} = \otimes_{p\,c\in\mathcal{C}(G')} \mathcal{P}_{c\,|\,pa_G(c)}$, which implies that $\mathcal{P}_S = \otimes_{p\,c\in\mathcal{C}(G)} \mathcal{P}_{c\,|\,pa_G(c)}$, as desired.

(b) Assume that for every component $c$, $\varphi_{c,pa_G(c)}(A)$ is a plausibility distribution over $c$ for all assignments $A$ of $pa_G(c)$. For $|\mathcal{C}(G)| = 1$, $\mathcal{C}(G) = \{c_1\}$. Then, $\gamma_S = \varphi_{c_1}$ is a plausibility distribution over $c_1$. Moreover, as $\gamma_{\emptyset\,|\,\emptyset} = 1_p$, we can write $\gamma_{c_1\cup\emptyset\,|\,\emptyset} = \gamma_{c_1\,|\,\emptyset} \otimes_p \gamma_{\emptyset\,|\,\emptyset}$, i.e. $I(c_1, \emptyset \,|\, \emptyset)$. Therefore, $G$ is compatible with $\gamma_{c_1}$: the proposition holds for $|\mathcal{C}(G)| = 1$.

Assume that the proposition holds for all DAGs with $n$ components. Consider a DAG $G$ with $n+1$ components. We first show that $\gamma_S$ is a plausibility distribution over $S$, i.e. $\oplus_{p_S}(\otimes_{p\,c\in\mathcal{C}(G)} \varphi_{c,pa_G(c)}) = 1_p$. Let $c_0$ be a leaf component in $G$. As $c_0$ is a leaf, the unique scoped function whose scope contains a variable in $c_0$ is $\varphi_{c_0,pa_G(c_0)}$. By the distributivity of $\otimes_p$ over $\oplus_p$, this implies that
$$\oplus_{p_{c_0}}(\otimes_{p\,c\in\mathcal{C}(G)} \varphi_{c,pa_G(c)}) = (\oplus_{p_{c_0}} \varphi_{c_0,pa_G(c_0)}) \otimes_p (\otimes_{p\,c\in\mathcal{C}(G)-\{c_0\}} \varphi_{c,pa_G(c)}).$$
Given that $\varphi_{c_0,pa_G(c_0)}(A)$ is a plausibility distribution over $c_0$ for all assignments $A$ of $pa_G(c_0)$,

$\oplus_{p_{c_0}} \varphi_{c_0,pa_G(c_0)} = 1_p$. Consequently,

$$\oplus_{p_{c_0}} (\otimes_{p_{c \in \mathcal{C}(G)}} \varphi_{c,pa_G(c)}) = \otimes_{p_{c \in \mathcal{C}(G)-\{c_0\}}} \varphi_{c,pa_G(c)}.$$

Applying the induction hypothesis to the DAG with $n$ components obtained from $G$ by deleting $c_0$, we can infer that $\oplus_{p_{S-c_0}} (\otimes_{p_{c \in \mathcal{C}(G)-\{c_0\}}} \varphi_{c,pa_G(c)}) = 1_p$. This allows us to write $\oplus_{p_{S-c_0}} (\oplus_{p_{c_0}} (\otimes_{p_{c \in \mathcal{C}(G)}} \varphi_{c,pa_G(c)})) = 1_p$, i.e. $\oplus_{p_S} \gamma_S = 1_p$: $\gamma_S$ is a plausibility distribution over $S$. It remains to prove that $G$ is a DAG of components compatible with $\gamma_S$. Let $c \in \mathcal{C}(G)$. We must show that $I(c, nd_G(c) - pa_G(c) \,|\, pa_G(c))$ holds. There are two cases:

1. If $c = c_0$, we must prove that

$$\gamma_{c_0, nd_G(c_0)-pa_G(c_0) \,|\, pa_G(c_0)} = \gamma_{c_0 \,|\, pa_G(c_0)} \otimes_p \gamma_{nd_G(c_0)-pa_G(c_0) \,|\, pa_G(c_0)}.$$

   First, note that

$$\begin{aligned}
\gamma_{c_0,pa_G(c_0)} &= \oplus_{p_{S-(c_0 \cup pa_G(c_0))}} (\otimes_{p_{c \in \mathcal{C}(G)}} \varphi_{c,pa_G(c)}) \\
&= (\oplus_{p_{S-(c_0 \cup pa_G(c_0))}} (\otimes_{p_{c \in \mathcal{C}(G)-\{c_0\}}} \varphi_{c,pa_G(c)})) \otimes_p \varphi_{c_0,pa_G(c_0)} \\
&\quad \text{(because } \otimes_p \text{ distributes over } \oplus_p \text{ and } sc(\varphi_{c_0,pa_G(c_0)}) \subseteq c_0 \cup pa_G(c_0)\text{)} \\
&= (\oplus_{p_{S-pa_G(c_0)}} (\otimes_{p_{c \in \mathcal{C}(G)}} \varphi_{c,pa_G(c)})) \otimes_p \varphi_{c_0,pa_G(c_0)} \\
&\quad \text{(because } \otimes_p \text{ distributes over } \oplus_p \text{ and } \oplus_{c_0} \varphi_{c_0,pa_G(c_0)} = 1_p\text{)} \\
&= \gamma_{pa_G(c_0)} \otimes_p \varphi_{c_0,pa_G(c_0)}.
\end{aligned}$$

   From this, it is possible to write:

$$\begin{aligned}
& \gamma_{nd_G(c_0)-pa_G(c_0) \,|\, pa_G(c_0)} \otimes_p \gamma_{c_0 \,|\, pa_G(c_0)} \otimes_p \gamma_{pa_G(c_0)} \\
= \; & \gamma_{nd_G(c_0)-pa_G(c_0) \,|\, pa_G(c_0)} \otimes_p \gamma_{c_0,pa_G(c_0)} \\
= \; & \gamma_{nd_G(c_0)-pa_G(c_0) \,|\, pa_G(c_0)} \otimes_p \gamma_{pa_G(c_0)} \otimes_p \varphi_{c_0,pa_G(c_0)} \\
= \; & \gamma_{nd_G(c_0)} \otimes_p \varphi_{c_0,pa_G(c_0)} \\
= \; & \gamma_{S-\{c_0\}} \otimes_p \varphi_{c_0,pa_G(c_0)} \text{ (because } c_0 \text{ is a leaf in } G\text{)} \\
= \; & (\otimes_{p_{c \in \mathcal{C}(G)-\{c_0\}}} \varphi_{c,pa_G(c)}) \otimes_p \varphi_{c_0,pa_G(c_0)} \\
= \; & \otimes_{p_{c \in \mathcal{C}(G)}} \varphi_{c,pa_G(c)} \\
= \; & \gamma_S.
\end{aligned}$$

   Using axiom (e) of Definition 18, this entails that $\gamma_{nd_G(c_0)-pa_G(c_0) \,|\, pa_G(c_0)} \otimes_p \gamma_{c_0 \,|\, pa_G(c_0)} = \gamma_{S-pa_G(c_0) \,|\, pa_G(c_0)}$, i.e., as $S = c_0 \cup nd_G(c_0)$, that $I(c_0, nd_G(c_0) - pa_G(c_0) \,|\, pa_G(c_0))$.

2. Otherwise, $c \neq c_0$. Let $G'$ be the DAG obtained from $G$ by deleting $c_0$. $G'$ contains $n$ components: by the induction hypothesis, $I(c, nd_{G'}(c) - pa_{G'}(c) \,|\, pa_{G'}(c))$. As $c_0$ is a leaf in $G$, we have $c_0 \notin pa_G(c)$, which implies that $pa_{G'}(c) = pa_G(c)$. Thus, $I(c, nd_{G'}(c) - pa_G(c) \,|\, pa_G(c))$.

   (i) If $nd_{G'}(c) = nd_G(c)$, then it is immediate that $I(c, nd_G(c) - pa_G(c) \,|\, pa_G(c))$.

   (ii) Otherwise, $nd_{G'}(c) \neq nd_G(c)$. As $c_0$ is a leaf in $G$, this is equivalent to say that $nd_G(c) = nd_{G'}(c) \cup c_0$. This means that $c$ is not an ancestor of $c_0$, and a fortiori $c \notin pa_G(c_0)$. In the following, the four semigraphoid axioms are used to prove the required result. From the decomposition axiom, from $I(c_0, nd_G(c_0) - pa_G(c_0) \,|\, pa_G(c_0))$, and from $(c \cup nd_{G'}(c)) \subseteq nd_G(c_0)$ (because $nd_G(c_0) = S - c_0$), it follows that $I(c_0, (c \cup nd_{G'}(c)) - pa_G(c_0) \,|\, pa_G(c_0))$, or, in other words, as $c \cap pa_G(c_0) = \emptyset$, that $I(c_0, c \cup (nd_{G'}(c) - pa_G(c_0)) \,|\, pa_G(c_0))$. Using the weak union axiom leads to $I(c_0, c \,|\, (nd_{G'}(c) - pa_G(c_0)) \cup pa_G(c_0))$ and, using the symmetry axiom, to $I(c, c_0 \,|\, (nd_{G'}(c) - pa_G(c_0)) \cup pa_G(c_0))$. As shown previously, $I(c, nd_{G'}(c) - pa_G(c) \,|\, pa_G(c))$. Together with $I(c, c_0 \,|\, (nd_{G'}(c) - pa_G(c_0)) \cup pa_G(c_0))$, the contraction axiom implies that $I(c, (nd_{G'}(c) - pa_G(c)) \cup c_0 \,|\, pa_G(c))$. As $c_0 \notin pa_G(c)$ and $nd_G(c) = nd_{G'}(c) \cup c_0$, this means that $I(c, nd_G(c) - pa_G(c)) \,|\, pa_G(c))$.

We have proved that $G$ is compatible with $\gamma_S$. Consequently, the proposition holds if there are $n+1$ components in $G$, which ends the proof by induction.

$\square$

**Proposition 5** Let $(E_p, \oplus_p, \otimes_p)$ be a conditionable plausibility structure. Then, for all $n \in \mathbb{N}^*$, there exists a unique $p_0$ such that $\oplus_{p\, i\in[1,n]}\, p_0 = 1_p$.

*Proof.* Let $n \in \mathbb{N}^*$. If $\oplus_{p\, i\in[1,n]}\, 1_p = 1_p$, then $p_0 = 1_p$ satisfies the required property. Moreover, in this case, the distributivity of $\otimes_p$ over $\oplus_p$ implies that for all $p \in E_p$, $\oplus_{p\, i\in[1,n]}\, p = p$. Therefore, if $\oplus_{p\, i\in[1,n]}\, p = 1_p$, then $p = 1_p$, which shows that $p_0$ is unique.

Otherwise, $\oplus_{p\, i\in[1,n]}\, 1_p \neq 1_p$. In this case, as $1_p \preceq_p \oplus_{p\, i\in[1,n]}\, 1_p$ by monotonicity of $\oplus_p$, we can write $1_p \prec_p \oplus_{p\, i\in[1,n]}\, 1_p$. The second item of Definition 19 then implies that there exists a unique $p_0 \in E_p$ such that $1_p = p_0 \otimes_p (\oplus_{p\, i\in[1,n]}\, 1_p)$, i.e. such that $1_p = \oplus_{p\, i\in[1,n]}\, p_0$. $\square$

**Proposition 6** Let $\mathcal{P}_{V_E,V_D}$ be the completion of a controlled plausibility distribution $\mathcal{P}_{V_E \,||\, V_D}$. Then, $\mathcal{P}_{V_E,V_D}$ is a plausibility distribution over $V_E \cup V_D$ and $\mathcal{P}_{V_E \,|\, V_D} = \mathcal{P}_{V_E \,||\, V_D}$.

*Proof.* $\mathcal{P}_{V_E,V_D} = \mathcal{P}_{V_E \,||\, V_D} \otimes_p p_0$, where $p_0$ is the element of $E_p$ such that $\oplus_{p\, i\in[1,|dom(V_D)|]}\, p_0 = 1_p$. Then $\oplus_{p\, V_E \cup V_D}\, \mathcal{P}_{V_E,V_D} = \oplus_{p\, V_E \cup V_D}(\mathcal{P}_{V_E \,||\, V_D} \otimes_p p_0) = \oplus_{p\, V_D}((\oplus_{p\, V_E}\, \mathcal{P}_{V_E \,||\, V_D}) \otimes_p p_0) = \oplus_{p\, V_D}\, p_0 = \oplus_{p\, i\in[1,|dom(V_D)|]}\, p_0 = 1_p$. This proves that $\mathcal{P}_{V_E,V_D}$ is a plausibility distribution over $V_E \cup V_D$.

As $\mathcal{P}_{V_E,V_D} = \mathcal{P}_{V_E \,||\, V_D} \otimes_p p_0$ and $\mathcal{P}_{V_E,V_D} = \mathcal{P}_{V_E \,|\, V_D} \otimes_p \mathcal{P}_{V_D}$, we can write $\mathcal{P}_{V_E \,||\, V_D} \otimes_p p_0 = \mathcal{P}_{V_E \,|\, V_D} \otimes_p \mathcal{P}_{V_D}$. Moreover, $\mathcal{P}_{V_D} = \oplus_{p\, V_E}\, \mathcal{P}_{V_E,V_D} = \oplus_{p\, V_E}(\mathcal{P}_{V_E \,||\, V_D} \otimes_p p_0) = p_0$. Thus, $\mathcal{P}_{V_E \,||\, V_D} \otimes_p p_0 = \mathcal{P}_{V_E \,|\, V_D} \otimes_p p_0$. Summing this equation $|dom(V_D)|$ times with $\oplus_p$ gives $\mathcal{P}_{V_E \,||\, V_D} = \mathcal{P}_{V_E \,|\, V_D}$. $\square$

**Proposition 7** Let $G$ be a typed DAG of components over $V_E \cup V_D$. Let $G_p$ be the partial graph of $G$ induced by the arcs of $G$ incident to environment components. Let $G_f$ be the partial graph of $G$ induced by the arcs of $G$ incident to decision components. If $G_p$ is compatible with the completion of $\mathcal{P}_{V_E \,||\, V_D}$ (cf. Definition 22) and $G_f$ is compatible with the completion of $\mathcal{F}_{V_D \,||\, V_E}$, then

$$\mathcal{P}_{V_E \,|\, V_D} = \underset{c \in \mathcal{C}_E(G)}{\otimes_p} \mathcal{P}_{c \,|\, pa_G(c)} \text{ and } \mathcal{F}_{V_D \,|\, V_E} = \bigwedge_{c \in \mathcal{C}_D(G)} \mathcal{F}_{c \,|\, pa_G(c)}.$$

*Proof.* The result is proved only for $\mathcal{P}_{V_E \,|\, V_D}$ (the proof for $\mathcal{F}_{V_D \,|\, V_E}$ is similar). The completion of $\mathcal{P}_{V_E \,||\, V_D}$ looks like $\mathcal{P}_{V_E,V_D} = \mathcal{P}_{V_E \,||\, V_D} \otimes_p p_0$. $G_p$ being compatible with this completion, Theorem 1a entails that $\mathcal{P}_{V_E,V_D} = \otimes_{p\, c\in\mathcal{C}(G_p)}\, \mathcal{P}_{c \,|\, pa_{G_p}(c)}$. As the decision components are roots in $G_p$, we can infer, by successively eliminating the environment components, that $\mathcal{P}_{V_D} = \oplus_{p\, V_E}\, \mathcal{P}_{V_E,V_D} = \otimes_{p\, c\in\mathcal{C}_D(G_p)}\, \mathcal{P}_c$.

On the other hand, $\mathcal{P}_{V_D} = \oplus_{p\, V_E}\left(\mathcal{P}_{V_E \,||\, V_D} \otimes_p p_0\right) = p_0$. This proves that $\otimes_{p\, c\in\mathcal{C}_D(G_p)}\, \mathcal{P}_c = p_0$. Therefore, $\mathcal{P}_{V_E,V_D} = \mathcal{P}_{V_E \,|\, V_D} \otimes_p p_0 = (\otimes_{p\, c\in\mathcal{C}_E(G_p)}\, \mathcal{P}_{c \,|\, pa_{G_p}(c)}) \otimes_p p_0$. Summing this equation $|dom(V_D)|$ times with $\oplus_p$ gives $\mathcal{P}_{V_E \,|\, V_D} = \otimes_{p\, c\in\mathcal{C}_E(G_p)}\, \mathcal{P}_{c \,|\, pa_{G_p}(c)}$. As $\mathcal{C}_E(G_p) = \mathcal{C}_E(G)$ and $pa_{G_p}(c) = pa_G(c)$ for every $c \in \mathcal{C}_E(G)$, this entails that $\mathcal{P}_{V_E \,|\, V_D} = \otimes_{p\, c\in\mathcal{C}_E(G)}\, \mathcal{P}_{c \,|\, pa_G(c)}$. $\square$

**Proposition 8** Assume that the plausibility structure used is conditionable. Let $Q = (\mathcal{N}, Sov)$ be a query where $Sov = (op_1, S_1) \cdot (op_2, S_2) \cdots (op_k, S_k)$. Let $V_{fr}$ denote the set of free variables of $Q$.

(1) If $S_i \subseteq V_E$ and $\mathcal{P}_{S_i \,|\, l(S_i)}(A)$ is well-defined, then there exists at least one $A' \in dom(S_i)$ satisfying $\mathcal{P}_{S_i \,|\, l(S_i)}(A.A') \neq 0_p$.

(2) If $S_i \subseteq V_D$ and $\mathcal{F}_{S_i \,|\, l(S_i)}(A)$ is well-defined, then there exists at least one $A' \in dom(S_i)$ satisfying $\mathcal{F}_{S_i \,|\, l(S_i)}(A.A') = t$.

(3) If $V_E \neq \emptyset$ and $S_i$ is the leftmost set of environment variables appearing in $Sov$, then, for all $A \in dom(l(S_i))$, $\mathcal{P}_{S_i \,|\, l(S_i)}(A)$ is well-defined.

(4) If $i, j \in [1,k]$, $i < j$, $S_i \subseteq V_E$, $S_j \subseteq V_E$, $r(S_i) \cap l(S_j) \subseteq V_D$ ($S_j$ is the first set of environment variables appearing to the right of $S_i$ in $Sov$), $(A, A') \in dom(l(S_i)) \times dom(S_i)$, $\mathcal{P}_{S_i \,|\, l(S_i)}(A)$ is well-defined, and $\mathcal{P}_{S_i \,|\, l(S_i)}(A.A') \neq 0_p$, then, for all $A''$ extending $A.A'$ over $l(S_j)$, $\mathcal{P}_{S_j \,|\, l(S_j)}(A'')$ is well-defined.

(5) If $i, j \in [1,k]$, $i < j$, $S_i \subseteq V_D$, $S_j \subseteq V_D$, $r(S_i) \cap l(S_j) \subseteq V_E$ ($S_j$ is the first set of decision variables appearing to the right of $S_i$ in $Sov$), $(A, A') \in dom(l(S_i)) \times dom(S_i)$, $\mathcal{F}_{S_i \,|\, l(S_i)}(A)$ is well-defined, and $\mathcal{F}_{S_i \,|\, l(S_i)}(A.A') = t$, then, for all $A''$ extending $A.A'$ over $l(S_j)$, $\mathcal{F}_{S_j \,|\, l(S_j)}(A'')$ is well-defined.

(6) For all $i \in [1,k]$ such that $S_i \subseteq V_E$, $\mathcal{P}_{S_i \,|\, l(S_i)} = \mathcal{P}_{S_i \,|\, l(S_i) \cap V_E \,||\, V_D}$.

(7) For all $i \in [1,k]$ such that $S_i \subseteq V_D$, $\mathcal{F}_{S_i \,|\, l(S_i)} = \mathcal{F}_{S_i \,|\, l(S_i) \cap V_D \,||\, V_E}$.

*Proof.* We denote by $p_0$ the element in $E_p$ such that the completion of $\mathcal{P}_{V_E \,||\, V_D}$ equals $\mathcal{P}_{V_E \,||\, V_D} \otimes p_0$. Note that $p_0 \neq 0_p$, since it must satisfy $\oplus_{p_{i \in [1, |dom(V_D)|]}} p_0 = 1_p$.

**Lemma 1.** *Let $(E_p, \oplus_p, \otimes_p)$ be a conditionable plausibility structure. Then, $(p_1 \otimes_p p_2 = 0_p) \leftrightarrow ((p_1 = 0_p) \vee (p2 = 0_p))$.*

*Proof.* First, if $p_1 = 0_p$ or $p_2 = 0_p$, then $p_1 \otimes_p p_2 = 0_p$. Conversely, assume that $p_1 \otimes_p p_2 = 0_p$. Then, if $p_1 \succ_p 0_p$, the conditionability of the plausibility structure together with $p_1 \otimes_p 0_p = 0_p$ entails that $p_2 = 0_p$. Similarly, if $p_2 \succ_p 0_p$, then $p_1 = 0_p$. Therefore $(p_1 \otimes_p p_2 = 0_p) \rightarrow ((p_1 = 0_p) \vee (p_2 = 0_p))$. □

**Lemma 2.** *Assume that the plausibility structure is conditionable. Let $S_1$, $S_2$ be disjoint subsets of $V_E$. Then, $\mathcal{P}_{S_1 \,|\, S_2 \,||\, V_D} = \mathcal{P}_{S_1 \,|\, S_2, V_D}$.*

*Proof.* Note that $\mathcal{P}_{S_1, S_2 \,|\, V_D} = \mathcal{P}_{S_1 \,|\, S_2, V_D} \otimes_p \mathcal{P}_{S_2 \,|\, V_D}$. Moreover, we can also write $\mathcal{P}_{S_1, S_2 \,|\, V_D} = \mathcal{P}_{S_1, S_2 \,||\, V_D} = \mathcal{P}_{S_1 \,|\, S_2 \,||\, V_D} \otimes_p \mathcal{P}_{S_2 \,||\, V_D} = \mathcal{P}_{S_1 \,|\, S_2 \,||\, V_D} \otimes_p \mathcal{P}_{S_2 \,|\, V_D}$. Let $A$ be an assignment of $V$. If $\mathcal{P}_{S_1, S_2 \,|\, V_D}(A) \prec_p \mathcal{P}_{S_2 \,|\, V_D}(A)$, then the conditionability of the plausibility structure entails that $\mathcal{P}_{S_1 \,|\, S_2, V_D}(A) = \mathcal{P}_{S_1 \,|\, S_2 \,||\, V_D}(A)$. Otherwise, $\mathcal{P}_{S_1, S_2 \,|\, V_D}(A) = \mathcal{P}_{S_2 \,|\, V_D}(A)$, which also entails that $\mathcal{P}_{S_1, S_2 \,||\, V_D}(A) = \mathcal{P}_{S_2 \,||\, V_D}(A)$. In this case, $\mathcal{P}_{S_1 \,|\, S_2, V_D}(A) = \mathcal{P}_{S_1 \,|\, S_2 \,||\, V_D}(A) = 1_p$. Therefore, $\mathcal{P}_{S_1 \,|\, S_2, V_D} = \mathcal{P}_{S_1 \,|\, S_2 \,||\, V_D}$. □

(1) Assume that $S_i \subseteq V_E$ and $\mathcal{P}_{S_i \,|\, l(S_i)}(A)$ is well-defined. Then, $\mathcal{P}_{S_i \,|\, l(S_i)}(A)$ is a plausibility distribution over $S_i$. Hence, $\oplus_{p_{A' \in dom(S_i)}} \mathcal{P}_{S_i \,|\, l(S_i)}(A.A') = 1_p$, which implies that there exists at least one $A' \in dom(S_i)$ such that $\mathcal{P}_{S_i \,|\, l(S_i)}(A.A') \neq 0_p$.

(2) Proof similar to point (2).

(3) Assume that $V_E \neq \emptyset$. Let $S_i$ be the leftmost set of environment variables appearing in $Sov$ and let $A \in dom(l(S_i))$. Since $l(S_i) \cap V_E = \emptyset$, we can write $\mathcal{P}_{l(S_i)}(A) = \oplus_{p_{V - l(S_i)}} \mathcal{P}_{V_E, V_D}(A) = \oplus_{p_{V_D - l(S_i)}} (\oplus_{p_{V_E}} \mathcal{P}_{V_E, V_D}(A)) = \oplus_{p_{V_D - l(S_i)}} p_0 \neq 0_p$. Therefore, $\mathcal{P}_{S_i \,|\, l(S_i)}(A)$ is well-defined.

(6) Let $l_E(S_i) = l(S_i) \cap V_E$ and $l_D(S_i) = l(S_i) \cap V_D$. For a set of variables $S$, we denote by $d_G(S)$ the set of variables in $V$ that are descendant in the DAG $G$ of at least one variable in $S$.

First, $\mathcal{P}_{S_i, l_E(S_i) \,||\, V_D} = \oplus_{p\,V_E - (S_i \cup l_E(S_i))} \mathcal{P}_{V_E \,||\, V_D} = \oplus_{p\,V_E - (S_i \cup l_E(S_i))} (\otimes_{p\,P_j \in P} P_j)$. By definition of a query, variables in $V_E \cap d_G(V_D - l_D(S_i))$ do not belong to $S_i \cup l_E(S_i)$ (the environment variables that are descendants of as-yet-unassigned decision variables are not assigned yet). Thus, $\mathcal{P}_{S_i, l_E(S_i) \,||\, V_D} = \oplus_{p\,V_E - (S_i \cup l_E(S_i) \cup d_G(V_D - l_D(S_i)))} (\otimes_{p\,P_j \in Fact(c), c \not\subseteq V_E \cap d_G(V_D - l_D(S_i))} P_j)$. The last equality is obtained by successively eliminating the environment components included in $d_G(V_D - l_D(S_i))$ (using the normalization conditions). As the scope of a plausibility function $P_j \in Fact(c)$ is included in $c \cup pa_G(c)$, this equality entails that $\mathcal{P}_{S_i, l_E(S_i) \,||\, V_D}$ does not depend on the assignment of $V_D - l_D(S_i)$. Morever, $\mathcal{P}_{l_E(S_i) \,||\, V_D} = \oplus_{S_i} \mathcal{P}_{S_i, l_E(S_i) \,||\, V_D}$ does not depend on the assignment of $V_D - l_D(S_i)$ too. As $\mathcal{P}_{S_i \,|\, l_E(S_i) \,||\, V_D} = max\{p \in E_p \,|\, \mathcal{P}_{S_i, l_E(S_i) \,||\, V_D} = p \otimes_p \mathcal{P}_{l_E(S_i) \,||\, V_D}\}$, this also entails that $\mathcal{P}_{S_i \,|\, l_E(S_i) \,||\, V_D}$ does not depend on the assignment of $V_D - l_D(S_i)$. It can be denoted $\mathcal{P}_{S_i \,|\, l_E(S_i) \,||\, l_D(S_i)}$.

We now show that $\mathcal{P}_{S_i \,|\, l(S_i)} = \mathcal{P}_{S_i \,|\, l_E(S_i) \,||\, l_D(S_i)}$. First, note that

$$\begin{aligned}
\mathcal{P}_{S_i, l(S_i)} &= \oplus_{p\,V_D - l_D(S_i)} \mathcal{P}_{S_i, l_E(S_i), V_D} = \oplus_{p\,V_D - l_D(S_i)} (\mathcal{P}_{S_i \,|\, l_E(S_i), V_D} \otimes_p \mathcal{P}_{l_E(S_i), V_D}) \\
&= \oplus_{p\,V_D - l_D(S_i)} (\mathcal{P}_{S_i \,|\, l_E(S_i) \,||\, V_D} \otimes_p \mathcal{P}_{l_E(S_i), V_D}) \text{ (using Lemma 2)} \\
&= \mathcal{P}_{S_i \,|\, l_E(S_i) \,||\, V_D} \otimes_p (\oplus_{p\,V_D - l_D(S_i)} \mathcal{P}_{l_E(S_i), V_D}) \\
&\quad \text{(since } \mathcal{P}_{S_i \,|\, l_E(S_i) \,||\, V_D} \text{ does not depend on the assignment of } V_D - l_D(S_i)) \\
&= \mathcal{P}_{S_i \,|\, l_E(S_i) \,||\, V_D} \otimes_p \mathcal{P}_{l(S_i)}.
\end{aligned}$$

Let $A$ be an assignment of $V$.

- If $\mathcal{P}_{S_i, l(S_i)}(A) \prec_p \mathcal{P}_{l(S_i)}(A)$, then the conditionability of the plausibility structure directly entails that $\mathcal{P}_{S_i \,|\, l(S_i)}(A) = \mathcal{P}_{S_i \,|\, l_E(S_i) \,||\, V_D}(A)$.
- Otherwise, $\mathcal{P}_{S_i, l(S_i)}(A) = \mathcal{P}_{l(S_i)}(A)$. In this case, $\mathcal{P}_{S_i \,|\, l(S_i)}(A) = 1_p$. Next, as $V - l(S_i) = (V_D - l_D(S_i)) \cup (V_E - l_E(S_i))$, observe that $\mathcal{P}_{l(S_i)} = \oplus_{p\,V - l(S_i)} (\mathcal{P}_{V_E \,||\, V_D} \otimes_p p_0) = \oplus_{p\,V_D - l_D(S_i)} (\mathcal{P}_{l_E(S_i) \,||\, V_D} \otimes_p p_0)$. Similarly, we have $\mathcal{P}_{S_i, l(S_i)} = \oplus_{p\,V - (S_i \cup l(S_i))} (\mathcal{P}_{V_E \,||\, V_D} \otimes_p p_0) = \oplus_{p\,V_D - l_D(S_i)} (\mathcal{P}_{S_i, l_E(S_i) \,||\, V_D} \otimes_p p_0)$. As $\mathcal{P}_{S_i, l(S_i)}(A) = \mathcal{P}_{l(S_i)}(A)$, we can infer that $\oplus_{p\,V_D - l_D(S_i)} (\mathcal{P}_{l_E(S_i) \,||\, V_D}(A) \otimes_p p_0) = \oplus_{p\,V_D - l_D(S_i)} (\mathcal{P}_{S_i, l_E(S_i) \,||\, V_D}(A) \otimes_p p_0)$. As neither $\mathcal{P}_{l_E(S_i) \,||\, V_D}$ nor $\mathcal{P}_{S_i, l_E(S_i) \,||\, V_D}$ depends on the assignment of $V_D - l_D(S_i)$, this entails that $\mathcal{P}_{l_E(S_i) \,||\, V_D}(A) \otimes_p (\oplus_{p\,V_D - l_D(S_i)} p_0) = \mathcal{P}_{S_i, l_E(S_i) \,||\, V_D}(A) \otimes_p (\oplus_{p\,V_D - l_D(S_i)} p_0)$. Summing this equation $|dom(l_D(S_i))|$ times gives $\mathcal{P}_{S_i, l_E(S_i) \,||\, V_D}(A) = \mathcal{P}_{l_E(S_i) \,||\, V_D}(A)$, and thus $\mathcal{P}_{S_i \,|\, l_E(S_i) \,||\, V_D}(A) = 1_p = \mathcal{P}_{S_i \,|\, l(S_i)}(A)$.

(7) Proof similar to point (6).

(4) Let $i, j \in [1, k]$ such that $i < j$, $S_i \subseteq V_E$, $S_j \subseteq V_E$, and $r(S_i) \cap l(S_j) \subseteq V_D$ ($S_j$ is the first set of environment variables appearing to the right of $S_i$ in $Sov$). Let $(A, A') \in dom(l(S_i)) \times dom(S_i)$ such that $\mathcal{P}_{S_i \,|\, l(S_i)}(A)$ is well-defined (i.e. $\mathcal{P}_{l(S_i)}(A) \neq 0_p$) and $\mathcal{P}_{S_i \,|\, l(S_i)}(A.A') \neq 0_p$. Let $A''$ be an extension of $A.A'$ over $l(S_j)$. We must show that $\mathcal{P}_{S_j \,|\, l(S_j)}(A'')$ is well-defined, i.e. that $\mathcal{P}_{l(S_j)}(A'') \neq 0_p$. As $\mathcal{P}_{S_i \,|\, l(S_i)}(A.A') \neq 0_p$ and $\mathcal{P}_{l(S_i)}(A) \neq 0_p$, Lemma 1 implies that $\mathcal{P}_{S_i, l(S_i)}(A.A') \neq 0_p$. Similarly to the proof of point (6), it is possible to show that $\mathcal{P}_{l(S_j)}$ does not depend on the assignment of $l(S_j) - (S_i \cup l(S_i))$. Therefore, for every $A''$ extending $A.A'$ over $l(S_j)$, $\oplus_{p\,l(S_j) - (S_i \cup l(S_i))} \mathcal{P}_{l(S_j)}(A'') \neq 0_p$, which implies that $\mathcal{P}_{l(S_j)}(A'') \neq 0_p$.

(5) Proof similar to point (4), except that plausibilities are replaced by feasibilities and decision variables are replaced by environment variables.

□

**Theorem 2** If the plausibility structure is conditionable, then, for all queries $Q$ on a PFU network, $Sem\text{-}Ans(Q) = Op\text{-}Ans(Q)$ and the optimal policies for the decisions are the same with $Sem\text{-}Ans(Q)$ and $Op\text{-}Ans(Q)$.

*Proof.* Let $A_{fr}$ be an assignment of the set of free variables $V_{fr}$ such that $\mathcal{F}_{V_{fr}}(A_{fr}) = f$. The semantic definition gives $(Sem\text{-}Ans(Q))(A_{fr}) = \lozenge$. Given that $\mathcal{F}_{V_{fr}}(A_{fr}) = \vee_{V-V_{fr}} \mathcal{F}_{V_E,V_D}(A_{fr}) = \vee_{V-V_{fr}} \mathcal{F}_{V_D \,||\, V_E}(A_{fr}) = \vee_{V-V_{fr}}(\wedge_{F_i \in F} F_i(A_{fr}))$ (since the completion of $\mathcal{F}_{V_D \,||\, V_E}$ gives $\mathcal{F}_{V_D \,||\, V_E} = \mathcal{F}_{V_D,V_E}$), we can infer that for every complete assignment $A''$ extending $A_{fr}$, $\wedge_{F_i \in F} F_i(A'') = f$ and $(\wedge_{F_i \in F} F_i(A'')) \star (\otimes_{p\,P_i \in P} P_i(A'')) \otimes_{pu} (\otimes_{u\,U_i \in U} U_i(A'')) = \lozenge$. As $\min(\lozenge, \lozenge) = \max(\lozenge, \lozenge) = \lozenge \oplus_u \lozenge = \lozenge$, this entails that $(Op\text{-}Ans(Q))(A_{fr}) = \lozenge$ too.

We now analyze the case $\mathcal{F}_{V_{fr}}(A_{fr}) = t$. We use $A''$ to denote a complete assignment which must be considered with the semantic definition. Using the properties:

- $p \otimes_{pu} \min(u_1, u_2) = \min(p \otimes_{pu} u_1, p \otimes_{pu} u_2)$ (right monotonicity of $\otimes_{pu}$),
- $p \otimes_{pu} \max(u_1, u_2) = \max(p \otimes_{pu} u_1, p \otimes_{pu} u_2)$ (right monotonicity of $\otimes_{pu}$),
- $p \otimes_{pu} (u_1 \oplus_u u_2) = (p \otimes_{pu} u_1) \oplus_u (p \otimes_{pu} u_2)$ (distributivity of $\otimes_{pu}$ over $\oplus_u$),
- $p_1 \otimes_{pu} (p_2 \otimes_{pu} u) = (p_1 \otimes_p p_2) \otimes_{pu} u$,

we can "move" all the $\mathcal{P}_{S_i \,|\, l(S_i)}(A.A')$ to get, starting from the semantic definition,

$$(\otimes_{p\,i \in [1,k], S_i \subseteq V_E} \mathcal{P}_{S_i \,|\, l(S_i)})(A'') \otimes_{pu} \mathcal{U}_V(A'')$$

on the right of the elimination operators.

We now prove that this quantity equals $\mathcal{P}_{V_E \,|\, V_D}(A'') \otimes_{pu} \mathcal{U}_V(A'')$. Let $S$ be the rightmost set of quantified environment variables. The chain rule enables us to write $\mathcal{P}_{V_E \,|\, V_D} = \mathcal{P}_{S \,|\, l_E(S), V_D} \otimes_p \mathcal{P}_{l_E(S) \,|\, V_D}$, where $l_E(S) = l(S) \cap V_E$. Moreover, using Lemma 2 and Proposition 8(6), we can write $\mathcal{P}_{S \,|\, l_E(S), V_D} = \mathcal{P}_{S \,|\, l_E(S) \,||\, V_D} = \mathcal{P}_{S \,|\, l(S)}$. Therefore, $\mathcal{P}_{V_E \,|\, V_D} = \mathcal{P}_{S \,|\, l(S)} \otimes_p \mathcal{P}_{l_E(S) \,|\, V_D}$. Recursively applying this mechanism leads to: $\mathcal{P}_{V_E \,|\, V_D} = \otimes_{p\,i \in [1,k], S_i \subseteq V_E} \mathcal{P}_{S_i \,|\, l(S_i)}$. Therefore, we obtain $\mathcal{P}_{V_E \,|\, V_D}(A'') \otimes_{pu} \mathcal{U}_V(A'')$ on the right of the elimination operators.

The semantic definition of the query meaning can be simplified a bit, thanks to Lemma 1. This lemma implies that conditions like $\mathcal{P}_{S \,|\, l(S)}(A.A') \neq 0_p$, which are used only when $\mathcal{P}_{l(S)}(A) \neq 0_p$, are equivalent to $\mathcal{P}_{S,l(S)}(A.A') \neq 0_p$, since $\mathcal{P}_{S,l(S)}(A.A') = \mathcal{P}_{S \,|\, l(S)}(A.A') \otimes_p \mathcal{P}_{l(S)}(A)$. As a result, the operators $\oplus_{u\,A' \in dom(S), \mathcal{P}_{S \,|\, l(S)}(A.A') \neq 0_p}$ can be replaced by $\oplus_{u\,A' \in dom(S), \mathcal{P}_{S,l(S)}(A.A') \neq 0_p}$. Similarly, in the eliminations $\min_{A' \in dom(S), \mathcal{F}_{S \,|\, l(S)}(A.A') = t}$, the conditions $\mathcal{F}_{S \,|\, l(S)}(A.A') = t$ can be replaced by $\mathcal{F}_{S,l(S)}(A.A') = t$. The same holds for the eliminations with $\max_{a \in dom(x_i), \mathcal{F}_{S \,|\, l(S)}(A.A') = t}$.

We now start from the operational definition and show that it can be reformulated as above. The operational definition applies a sequence of variable eliminations on the global function $(\wedge_{F_i \in F} F_i) \star (\otimes_{p\,P_i \in P} P_i) \otimes_{pu} (\wedge_{U_i \in U} U_i)$, which also equals $\mathcal{F}_{V_D \,|\, V_E} \star \mathcal{P}_{V_E \,|\, V_D} \otimes_{pu} \mathcal{U}_V$. Let $S$ be the leftmost set of quantified decision variables. Let $A$ be an assignment of $l(S)$. Assume that $S$ is quantified by min. Let $A_0 \in dom(S)$ such that $\mathcal{F}_{S,l(S)}(A.A_0) = f$. It can be inferred that for all complete assignment $A''$ extending $A.A_0$, $\mathcal{F}_{V_E,V_D}(A'') = f$, and consequently $\mathcal{F}_{V_D \,|\, V_E}(A'') = f$. This implies that $\mathcal{F}_{V_D \,|\, V_E}(A'') \star \mathcal{P}_{V_E \,|\, V_D}(A'') \otimes_{pu} \mathcal{U}_V(A'') = \lozenge$. Given that $\min(\lozenge, \lozenge) = \max(\lozenge, \lozenge) = \lozenge \oplus_u \lozenge = \lozenge$, we obtain $Qo(\mathcal{N}, Sov, A.A_0) = \lozenge$. As $\min(d, \lozenge) = d$, this entails that $\min_{A' \in dom(S)} Qo(\mathcal{N}, Sov, A.A') = \min_{A' \in dom(S) - \{A_0\}} Qo(\mathcal{N}, Sov, A.A')$. Thus, $\min_{A' \in dom(S)}$ can be replaced by $\min_{A' \in dom(S), \mathcal{F}_{S,l(S)}(A.A') = t}$ (as $\mathcal{F}_{V_{fr}}(A) = t$, there exists at least one assignment $A' \in dom(S)$ such that $\mathcal{F}_{S,l(S)}(A.A') = t$). The same result holds if $S$ is quantified by max. Applying this mechanism to each set of quantified decision variables from the left to the right of $Sov$, we obtain that $\min_{A' \in dom(S)}$ and $\max_{A' \in dom(S)}$ can be replaced by $\min_{A' \in dom(S), \mathcal{F}_{S,l(S)}(A.A') = t}$ and $\max_{A' \in dom(S), \mathcal{F}_{S,l(S)}(A.A') = t}$ respectively. Moreover, it can be shown that for every complete assignment $A''$ considered in the corresponding transformed operational definition, $\mathcal{F}_{V_D \,|\, V_E}(A'') = t$. It is thus possible to replace $\mathcal{F}_{V_D \,|\, V_E}(A'') \star \mathcal{P}_{V_E \,|\, V_D}(A'') \otimes_{pu} \mathcal{U}_V(A'')$ by $\mathcal{P}_{V_E \,|\, V_D}(A'') \otimes_{pu} \mathcal{U}_V(A'')$.

We now transform each $\oplus_{u A' \in dom(S)} Qo(\mathcal{N}, Sov, A.A')$ so that it looks like the expression in the semantic definition. Let $S$ be the leftmost set of quantified environment variables. Let $A$ be an assignment of $l(S)$. Let $A_0 \in dom(S)$ be such that $\mathcal{P}_{S,l(S)}(A.A_0) = 0_p$. Then, for all complete assignments $A''$ extending $A.A_0$, $\mathcal{P}_{V_E \mid V_D}(A'') = 0_p$, and thus $\mathcal{P}_{V_E \mid V_D}(A'') \otimes_{pu} \mathcal{U}_V(A'') = 0_u$. As $\min(0_u, 0_u) = \max(0_u, 0_u) = 0_u \oplus_u 0_u = 0_u$, we obtain $Qo(\mathcal{N}, Sov, A.A_0) = 0_u$. As $d \oplus_u 0_u = d$, computing $\oplus_{u A' \in dom(S)} Qo(\mathcal{N}, Sov, A.A')$ is equivalent to computing $\oplus_{u A' \in dom(S)-\{A_0\}} Qo(\mathcal{N}, Sov, A.A')$. Thus, $\oplus_{u A' \in dom(S)}$ can be replaced by $\oplus_{u A' \in dom(S), \mathcal{P}_{S,l(S)}(A.A') \neq 0_p}$ (as $\mathcal{P}_{l(S)}(A) \neq 0_p$, there exists at least one assignment $A' \in dom(S)$ satisfying $\mathcal{P}_{S,l(S)}(A.A') \neq 0_p$). Applying this mechanism, considering each set of quantified environment variables from the left to the right of $Sov$, we get $\oplus_{u A' \in dom(S), \mathcal{P}_{S,l(S)}(A,A') \neq 0_p}$ instead of $\oplus_{u A' \in dom(S)}$.

Consequently, we have found a function $\Phi$ such that $Sem\text{-}Ans(Q) = \Phi$ and $Op\text{-}Ans(Q) = \Phi$. Moreover, the optimal policies for the decisions for $Sem\text{-}Ans(Q)$ are optimal policies for decisions for $\Phi$. Indeed, the transformation rules used preserve the set of optimal policies. The same holds for $Op\text{-}Ans(Q)$ and $\Phi$. It entails that $Sem\text{-}Ans(Q) = Op\text{-}Ans(Q)$, and that the optimal policies for $Sem\text{-}Ans(Q)$ are the same as those for $Op\text{-}Ans(Q)$. □

**Theorem 3** Queries and bounded queries can be used to express and solve the following nonexhaustive list of problems:

1. SAT framework: SAT, MAJSAT, E-MAJSAT, quantified boolean formula, stochastic SAT (SSAT) and extended-SSAT (Littman et al., 2001).

2. CSP (or CN) framework:

   - Check consistency for a CSP (Mackworth, 1977); find a solution to a CSP; count the number of solutions of a CSP.
   - Find a solution of a valued CSP (Bistarelli et al., 1999).
   - Solve a quantified CSP (Bordeaux & Monfroy, 2002).
   - Find a conditional decision or an unconditional decision for a mixed CSP or a probabilistic mixed CSP (Fargier et al., 1996).
   - Find an optimal policy for a stochastic CSP or a policy with a value greater than a threshold; solve a stochastic COP (Constraint Optimization Problem) (Walsh, 2002).

3. Integer Linear Programming (Schrijver, 1998) with finite domain variables.

4. Search for a solution plan with a length $\leq$ k in a classical planning problem (STRIPS planning, Fikes & Nilsson, 1971; Ghallab et al., 2004).

5. Answer classical queries on Bayesian networks (Pearl, 1988), Markov random fields (Chellappa & Jain, 1993), and chain graphs (Frydenberg, 1990), with plausibilities expressed as probabilities, possibilities, or $\kappa$-rankings:

   - Compute plausibility distributions.
   - MAP (Maximum A Posteriori hypothesis) and MPE (Most Probable Explanation).

- Compute the plausibility of an evidence.
- CPE task for hybrid networks (Dechter & Larkin, 2001) (CPE means CNF Probability Evaluation, a CNF being a formula in Conjunctive Normal Form).

6. Solve an influence diagram (Howard & Matheson, 1984).

7. With a finite horizon, solve a probabilistic MDP, a possibilistic MDP, a MDP based on $\kappa$-rankings, completely or partially observable (POMDP), factored or not (Puterman, 1994; Monahan, 1982; Sabbadin, 1999; Boutilier et al., 1999, 2000).

*Proof.*

**Lemma 3.** *Let $(E_p, E_u, \oplus_u, \otimes_{pu})$ be an expected utility structure such that $E_u$ is totally ordered by $\preceq_u$. Let $\gamma_{S_1,S_2}$ be a local function on $E_u$ whose scope is $S_1 \cup S_2$. Then*

$$\max_{\phi: dom(S_2)\rightarrow dom(S_1)} \bigoplus_{u\, A\in dom(S_2)} \gamma_{S_1,S_2}(\phi(A).A) = \bigoplus_{u\, S_2} \max_{S_1} \gamma_{S_1,S_2}.$$

*Moreover, $\psi : dom(S_2) \rightarrow dom(S_1)$ satisfies $(\max_{S_1} \gamma_{S_1,S_2})(A) = \gamma_{S_1,S_2}(\psi(A).A)$ for all $A \in dom(S_2)$ iff $\max_{\phi:dom(S_2)\rightarrow dom(S_1)} \oplus_{u A\in dom(S_2)} \gamma_{S_1,S_2}(\phi(A).A) = \oplus_{u A\in dom(S_2)} \gamma_{S_1,S_2}(\psi(A).A)$. In other words, the two sides of the equality have the same set of optimal policies for $S_1$.*

*Proof.* Let $\phi_0 : dom(S_2) \rightarrow dom(S_1)$ be a function such that
$\max_{\phi:dom(S_2)\rightarrow dom(S_1)} \oplus_{u A\in dom(S_2)} \gamma_{S_1,S_2}(\phi(A).A) = \oplus_{u A\in dom(S_2)} \gamma_{S_1,S_2}(\phi_0(A).A)$.
Given that, for all $A \in dom(S_2)$, $\gamma_{S_1,S_2}(\phi_0(A).A) \preceq_u \max_{A'\in dom(S_1)} \gamma_{S_1,S_2}(A'.A)$, the monotonicity of $\oplus_u$ entails that $\oplus_{u A\in dom(S_2)} \gamma_{S_1,S_2}(\phi_0(A).A) \preceq_u \oplus_{u A\in dom(S_2)} \max_{A'\in dom(S_1)} \gamma_{S_1,S_2}(A'.A)$. Thus,
$\max_{\phi:dom(S_2)\rightarrow dom(S_1)} \oplus_{u A\in dom(S_2)} \gamma_{S_1,S_2}(\phi(A).A) \preceq_u \oplus_{u S_2} \max_{S_1} \gamma_{S_1,S_2}$.

On the other hand, let $\psi_0 : dom(S_2) \rightarrow dom(S_1)$ be a function such that $\forall A \in dom(S_2)$, $(\max_{S_1} \gamma_{S_1,S_2})(A) = \gamma_{S_1,S_2}(\psi_0(A).A)$. Then,
$$\oplus_{u S_2} \max_{S_1} \gamma_{S_1,S_2} = \bigoplus_{u\, A\in dom(S_2)} \gamma_{S_1,S_2}(\psi_0(A).A) \preceq_u \max_{\phi:dom(S_2)\rightarrow dom(S_1)} \bigoplus_{u\, A\in dom(S_2)} \gamma_{S_1,S_2}(\phi(A).A).$$

The antisymmetry of $\preceq_u$ implies the required equality. The equality of the set of optimal policies over $S_1$ is directly implied by the equality. □

We now give the proof of the theorem, which uses for some cases the previous lemma.

1. (*CSP based problems, Mackworth, 1977*)
   Let us consider a CSP over a set of variables $V$ and with a set of constraints $\{C_1, \ldots, C_m\}$.

   (a) (*Consistency and solution finding*) Consistency can be checked by using the query $Q = (\mathcal{N}, (\max, V))$, where $\mathcal{N} = (V, G, \emptyset, \emptyset, U)$ (all variables in $V$ are decision variables, $G$ is reduced to a unique decision component containing all variables, and $U = \{C_1, \ldots, C_m\}$), and where the expected utility structure is boolean optimistic expected conjunctive utility (row 6 in Table 1). Computing $Ans(Q) = \max_V (C_1 \wedge \ldots \wedge C_m)$ is equivalent to checking consistency, because $Ans(Q) = t$ iff there exists an assignment of $V$ satisfying $C_1 \wedge \ldots \wedge C_m$, i.e. iff the CSP is consistent. In order to get a solution when $Ans(Q) = t$, it suffices to record an optimal decision rule for $V$. Integer Linear Programming (Schrijver, 1998) with finite domain variables can be formulated as a CSP.

   (b) (*Counting the number of solutions*) The expected utility structure considered for this task is probabilistic expected satisfaction (row 2 in Table 1). The PFU network is $\mathcal{N} = (V, G, P, \emptyset, U)$, where all variables in $V$ are environment variables, $G$ is a DAG with a unique component $c_0 = V$, $P = \{1/\varphi_0\}$, $\varphi_0$ being a constant factor equal to $|dom(V)|$ such that $Fact(c_0) = \{\varphi_0\}$, and $U = \{C_1, \ldots, C_m\}$. Implicitly, $1/\varphi_0$ specifies that the complete assignments are equiprobable. It enables the normalization condition "for all $c \in \mathcal{C}_E(G)$, $\oplus_{p_c} \otimes_{p_{P_i \in Fact(c)}} P_i = 1_p$" to be satisfied, since $\sum_V (1/|dom(V)|) = 1$. The query to consider is then $Q = (\mathcal{N}, (+, V))$. It is not hard to check that this satisfies the conditions imposed on queries and $Ans(Q) = \sum_V (1/\varphi_0 \times (C_1 \times \ldots \times C_m))$ gives the percentage of solutions of the CSP. $\varphi_0 \times Ans(Q)$ gives the number of solutions.

2. (*Solving a Valued CSP (VCSP), Bistarelli et al., 1999*)
In order to model this problem, the only difficulty lies in the definition of an expected utility structure. In a VCSP, a triple $(E, \circledast, \succ)$ called a valuation structure is introduced. It satisfies properties such as $(E, \circledast)$ is a commutative semigroup, $\succ$ is a total order on $E$, and $E$ has a minimum element denoted $\top$. The expected utility structure to consider is the following one: $(E_p, \oplus_p, \otimes_p) = (\{t, f\}, \vee, \wedge)$, $(E_u, \otimes_u) = (E, \circledast)$, and the expected utility structure is $(E_p, E_u, \oplus_u, \otimes_{pu})$, with $\oplus_u = \min$ and $\otimes_{pu}$ defined by "$false \otimes_{pu} u = \top$ and $true \otimes_{pu} u = u$" (it is not hard to check that this structure is an expected utility structure). Next, the PFU network is $\mathcal{N} = (V, G, \emptyset, \emptyset, U)$, where $V$ is the set of variables of the VCSP, $G$ is a DAG with only one decision component containing all the variables, and $U$ contains the soft constraints. The query $Q = (\min, V)$ enables us to find the minimum violation degree of the soft constraints. A solution for the VCSP is an optimal (argmin) decision rule for $V$.

3. (*Problems from the SAT framework, Littman et al., 2001*)
In the SAT framework, queries on a conjunctive normal form boolean formula $\phi$ over a set of variables $V = \{x_1, \ldots, x_n\}$ are asked. Let us first prove that an *extended SSAT* formula can be evaluated with a PFU query. An extended SSAT formula is defined by a triple $(\phi, \theta, q)$ where $\phi$ is a boolean formula in conjunctive normal form, $\theta$ is a threshold in $[0, 1]$, and $q = (q_1 x_1) \ldots (q_n x_n)$ is a sequence of quantifier/variable pairs (the quantifiers are $\exists$, $\forall$, or Я; the meaning of Я appears below). If we take $f \prec t$, the value of $\phi$ under the quantification sequence $q$, $val(\phi, q)$, is defined recursively by: (i) $val(\phi, \emptyset) = 1$ if $\phi$ is $t$, 0 otherwise; (ii) $val(\phi, (\exists x)\, q') = \max_x val(\phi, q')$; (iii) $val(\phi, (\forall x)\, q') = \min_x val(\phi, q')$; (iv) $val(\phi, (\text{Я} x)\, q') = \sum_x 0.5 \cdot val(\phi, q')$. Intuitively, the last case means that Я quantifies boolean variables taking equiprobable values. An extended SSAT formula $(\phi, \theta, q)$ is $t$ iff $val(\phi, q) \geq \theta$. If $S$ denotes the set of variables quantified by Я, an equivalent definition of $val(\phi, q)$ is: (i') $val(\phi, \emptyset) = 0.5^{|S|}$ if $\phi$ is $t$, 0 otherwise; (ii') $val(\phi, (\exists x)\, q') = \max_x val(\phi, q')$; (iii') $val(\phi, (\forall x)\, q') = \min_x val(\phi, q')$; (iv') $val(\phi, (\text{Я} x)\, q') = \sum_x val(\phi, q')$. This second definition proves that $val(\phi, q)$ can be computed with the PFU query defined by: (a) expected utility structure: probabilistic expected satisfaction (row 2 in Table 1); (b) PFU network: $\mathcal{N} = (V, G, P, \emptyset, U)$, with $V$ the set of variables of the formula $\phi$ (the decision variables are the variables quantified by $\exists$ or $\forall$), $G$ a DAG without arcs, with one decision component per decision variable and a unique environment component containing all variables quantified by Я, $P = \{\varphi_0\}$, $\varphi_0$ being a constant factor equal to $0.5^{|V_E|}$, and $U$ the set of clauses of $\phi$; (c) query: $Q = (\mathcal{N}, Sov)$, $Sov$ being obtained from $q$ by replacing $\exists$, $\forall$, and Я by max, min, and $+$ respectively. Then, $Ans(Q) = val(\phi, q)$, which implies that the value of an extended SSAT formula $(\phi, \theta, q)$ is the value of the bounded query $(\mathcal{N}, Sov, \theta)$.

*SSAT* is a particular case of extended-SSAT and is therefore covered. *SAT, MAJSAT, E-MAJSAT, QBF* are also particular cases of extended SSAT. As a result, they are instances of PFU bounded queries. More precisely, SAT corresponds to a bounded query of the form $Q = (\mathcal{N}, (\max, V), 1)$; MAJSAT ("given a boolean formula over a set of variables $V$, is it satisfied for at least half of the assignments of $V$") corresponds to a bounded query of the form $(\mathcal{N}, (+, V), 0.5)$; E-MAJSAT ("given a boolean formula over $V = V_E \cup V_D$, does there exist an assignment of $V_D$ such that the formula is satisfied for at least half of the assignments of $V_E$?") corresponds to a bounded query of the form $(\mathcal{N}, (\max, V_D).(+, V_E), 0.5)$; QBF corresponds to a bounded query in which max over existentially quantified variables and min over universally quantified variables alternate.

4. (*Solving a Quantified CSP (QCSP), Bordeaux & Monfroy, 2002*)
A QCSP represents a formula of the form $Q_1x_1 \ldots Q_nx_n\ (C_1 \wedge \ldots \wedge C_m)$, where each $Q_i$ is a quantifier ($\forall$ or $\exists$) and each $C_i$ is a constraint. The value of a QCSP is defined recursively as follows: the value of a QCSP without variables (i.e. containing only $t$, $f$, and connectives) is given by the definition of the connectives. A QCSP $\exists x\ qcsp$ is $t$ iff either $qcsp((x,t)) = t$ or $qcsp((x,f)) = t$. Assuming $f \prec t$, it gives that $\exists x\ qcsp$ is $t$ iff $\max_x qcsp = t$. A QCSP $\forall x\ qcsp$ is $t$ iff $qcsp((x,t)) = t$ and $qcsp((x,f)) = t$. Equivalently, $\forall x\ qcsp$ is $t$ iff $\min_x qcsp = t$. It implies that the value of a QCSP is actually given by the formula $op(Q_1)_{x_1} \ldots op(Q_n)_{x_n}\ (C_1 \wedge \ldots \wedge C_m)$, with $op(\exists) = \max$ and $op(\forall) = \min$. It corresponds to the answer to the query $(\mathcal{N}, (op(Q_1), x_1). \ldots .(op(Q_n), x_n))$, where $\mathcal{N} = (V, G, \emptyset, \emptyset, U)$ ($V$ is the set of variables of the QCSP, $G$ is a DAG with only one decision component containing all variables, and $U$ is the set of constraints), and where the expected utility structure is boolean optimistic expected conjunctive utility (row 6 in Table 1).

5. (*Solving a mixed CSP or a probabilistic mixed CSP, Fargier et al., 1996*)

A *probabilistic mixed CSP* is defined by (i) a set of variables partitioned into a set $W$ of *contingent* variables and a set $X$ of *decision* variables; an assignment $A_W$ of $W$ is called a world and an assignment $A_X$ of $X$ is called a decision; (ii) a set $C = \{C_1, \ldots, C_m\}$ of constraints involving at least one decision variable; (iii) a probability distribution $P_W$ over the worlds; a possible world $A_W$ (i.e. such that $P_W(A_W) > 0$) is covered by a decision $A_X$ iff the assignment $A_W.A_X$ satisfies all the constraints in $C$.

On one hand, if a decision must be made without knowing the world, the task is to find an optimal *non-conditional decision*, i.e. to find an assignment $A_X$ of the decision variables that maximizes the probability that the world is covered by $A_X$. This probability is equal to $\sum_{A_W \,|\, (C_1 \times \ldots \times C_m)(A_X, A_W)=1} P_W(A_W) = \sum_W (P_W \times C_1 \times \ldots \times C_m)$. As a result, an optimal non-conditional decision can be found by recording an optimal decision rule for $X$ for the formula $\max_X \sum_W (P_W \times C_1 \times \ldots \times C_m)$. The previous formula actually specifies how to solve such a problem with PFUs. The algebraic structure is probabilistic expected additive utility (row 2 in Table 1), the PFU network is $\mathcal{N} = (V, G, P, \emptyset, U)$, with $V_D = X$, $V_E = W$, $G$ a DAG without arc, with one decision component $X$ and a set of environment components that depends on how $P_W$ is specified, $P$ is the set of multiplicative factors that define $P_W$, and finally $U = \{C_1, \ldots, C_m\}$. The query is then $Q = (\mathcal{N}, (\max, X).(+, W))$.

On the other hand, if the world is known when the decision is made, the task is to look for an optimal *conditional decision*, i.e. to look for a decision rule $\phi_0 : dom(W) \rightarrow dom(X)$ which maximizes the probability that the world is covered. In other words, the goal is to compute $\max_{\phi: dom(W) \rightarrow dom(X)} \sum_{A_W \in dom(W) \,|\, (C_1 \times \ldots \times C_m)(A_W.\phi(A_W))=1} P_W(A_W) =$
$\max_{\phi: dom(W) \rightarrow dom(X)} \sum_{A_W \in dom(W)} (P_W \times C_1 \times \ldots \times C_m)\,(A_W.\phi(A_W))$. Due to Lemma 3, it also equals $\sum_W \max_X (P_W \times C_1 \times \ldots \times C_m)$, and $\phi_0$ can be found by recording an optimal decision rule for $X$. It proves that the query $Q = (\mathcal{N}, (+, W).(\max, X))$ enables us to compute an optimal conditional decision.

With *Mixed CSPs*, $P_W$ is replaced by a set $K$ of constraints defining the possible worlds. The goal is then to look for a decision, either conditional or non-conditional, that maximizes the number of covered worlds. This task is equivalent, ignoring a normalizing constant, to find a decision that maximizes the percentage of covered worlds. This can be solved using the set of plausibility functions $P = K \cup \{N_0\}$, with $N_0$ a normalizing constant ensuring that the normalization condition on plausibilities holds. $N_0$ is the number of possible worlds, but it does actually not need to be computed, since it is a constant factor and we are only interested in optimal decisions.

6. (*Stochastic CSP (SCSP) and stochastic COP (SCOP), Walsh, 2002*)
Formally, a SCSP is a tuple $(V, S, P, C, \theta)$, where $V$ is a list of variables (each variable $x$ having a finite domain $dom(x)$), $S$ is the set of stochastic variables in $V$, $P = \{P_s \mid s \in S\}$ is a set of probability distributions (in a more advanced version of SCSPs, probabilities over $S$ may be defined by a Bayesian network; the subsumption result is still valid for this case), $C = \{C_1, \ldots, C_m\}$ is a set of constraints, and $\theta$ is a threshold in $[0, 1]$.

A SCSP-policy is a tree with internal nodes labeled with variables. The root is labeled with the first variable in $V$, and the parents of the leaves are labeled with the last variable in $V$. Nodes labeled with a decision variable have only one child, whereas nodes labeled with a stochastic variable $s$ have $|dom(s)|$ children. Leaf nodes are labeled with 1 if the complete assignment they define satisfies all the constraints in $C$, and with 0 otherwise. With each leaf node can be associated a probability $\prod_{s \in S} P_s(A_S)$, where $A_S$ stands for the assignment of $S$ implicitly defined by the path from the root to the leaf. The satisfaction of a SCSP-policy is the sum of the values of the leaves weighted by their probabilities. A SCSP is satisfiable iff there exists a SCSP-policy with a satisfaction of at least $\theta$. The optimal satisfaction of a SCSP is the maximum satisfaction over all SCSP-policies.

For the subsumption proof, we first consider the problem of looking for the optimal satisfaction of a SCSP. In a SCSP-policy, each decision variable $x$ can take one value per assignment of the set $pred_s(x)$ of stochastic variables which precede $x$ in the list of variables $V$. Instead of being described as a tree, a SCSP-policy can be viewed as a set of functions $\Delta = \{\phi^x : dom(pred_s(x)) \rightarrow dom(x), x \in V - S\}$, and its value is $val(\Delta) = \sum_{A_S \in dom(S)} (\prod_{s \in S} P_s \times \prod_{C_i \in C} C_i)(A_S.(\underset{x \in V-S}{\cdot} \phi^x(A_S)))$. The goal is to maximize the previous quantity among the sets $\Delta$. Let $y$ be the last decision variable in $V$, and let $\Phi^y$ be the set of local functions $\phi^y : dom(pred_s(y)) \rightarrow dom(y)$ defining a decision rule for $y$. Then

$$\max_{\phi^y \in \Phi^y} val(\Delta) = \max_{\phi^y \in \Phi^y} \sum_{A_S \in dom(pred_s(y))} (\sum_{S - pred_s(y)} \prod_{s \in S} P_s \times \prod_{C_i \in C} C_i)(A_S.(\underset{x \in V-S}{\cdot} \phi_x(A_S))).$$

By Lemma 3, the previous quantity also equals:
$\sum_{pred_s(y))} \max_y \sum_{S - pred_s(y)} (\prod_{s \in S} P_s \times \prod_{C_i \in C} C_i)$. A recursive application of this mechanism shows that the answer $Ans(Q)$ to the query $Q$ described below is equal to the optimal satisfaction of a SCSP:

- expected utility structure: row 2 in Table 1 (probabilistic expected satisfaction);
- PFU network: $\mathcal{N} = (V', G, P, \emptyset, U)$, with $V'$ the set of variables of the SCSP; $V_E = S$ and $V_D = V' - S$; $G$ is a DAG without arcs, with one component per variable; $P = \{P_s \mid s \in S\}$; $Fact(\{s\}) = \{P_s\}$; $U$ is the set of constraints of the SCSP;
- query: $Q = (\mathcal{N}, Sov)$, with $Sov = t(V)$ ($V$ is the list of variables of the SCSP), $t(V)$ being recursively defined by $t(\emptyset) = \emptyset$ and $t(x.V'') = \begin{cases} (+, \{x\}).t(V'') \text{ if } x \in S \\ (\max, \{x\}).t(V'') \text{ otherwise} \end{cases}$.

An optimal SCSP-policy can be recorded during the evaluation of $Ans(Q)$. The satisfiability of a SCSP can be answered with the bounded query $(\mathcal{N}, Sov, \theta)$. Again, a corresponding SCSP-policy can be obtained by recording optimal decision rules.

With Stochastic Constraint Optimization Problem (SCOP), the constraints in $C$ are additive soft constraints. The subsumption proof is similar.

7. (*Classical planning problems (STRIPS planning), Fikes & Nilsson, 1971; Ghallab et al., 2004*)
In order to search for a plan of length lesser than $k$, we can simply model a classical planning problem as a CSP. Such a transformation is already available in the literature (Ghallab et al., 2004). However, we can also model a classical planning problem more directly in the PFU framework. More precisely, the state at one step can be described by a set of boolean environment variables. For each step, there is a unique decision variable whose set of values corresponds to the actions available. Plausibility functions are deterministic and link variables in step $t$ to variables in step $t+1$ (these functions simply specify the positive and negative effects of actions). The initial state is also represented by a plausibility function which links variables in the first step. Feasibility functions define preconditions for an action to be feasible. They link variables in a step $t$ to the decision variable of that step. Utility functions are boolean functions which describe the goal states. They hold over variables in step $k$. In order to search for a plan of length lesser than $k$, the sequence of elimination is a max-elimination on all variables. The expected utility structure used is the boolean optimistic expected disjunctive utility.

8. (*Influence diagrams, Howard & Matheson, 1984*)
We start from the definition of influence diagrams of Section 3. With each decision variable $d$, we can associate a decision rule $\delta^d : dom(pa_G(d)) \to dom(d)$. An influence diagram policy (ID-policy) is a set $\Delta = \{\delta^d \mid d \in D\}$ of decision rules (one for each decision variable). The value $val(\Delta)$ of an ID-policy $\Delta$ is given by the probabilistic expectation of the utility:
$$val(\Delta) = \sum_{A_S \in dom(S)} ((\prod_{s \in S} P_{s \mid pa_G(s)}) \times (\sum_{U_i \in U} U_i))(A_S.(\underset{d \in D}{\cdot} \delta^d(A_S))).$$
To solve an influence diagram, we must compute the maximum value of the previous quantity and find an associated optimal ID-policy. Using Lemma 3 and the DAG structure, it is possible to show, using the same ideas as in the SCSP subsumption proof, that the optimal expected utility is given by the answer to the query $Q$ below (associated optimal decision rules can be recorded during the evaluation of $Ans(Q)$):

 - expected utility structure: row 1 in Table 1 (probabilistic expected additive utility);
 - PFU network: $\mathcal{N} = (V, G', P, \emptyset, U)$; $V$ is the set of variables of the influence diagram, $G'$ is the DAG obtained from the DAG of the influence diagram by removing utility nodes and arcs into decision nodes; in $G'$, there is one component per variable; $P = \{P_{s \mid pa_G(s)}, s \in V_E\}$ and $Fact(\{s\}) = \{P_{s \mid pa_G(s)}\}$; $U$ is the set of utility functions associated with utility nodes.
 - PFU query: $Q = (\mathcal{N}, Sov)$, with $Sov$ obtained from the DAG of the influence diagram as follows. Initially, $Sov = \emptyset$. In the DAG of an influence diagram, the decisions are totally ordered. Let $d$ be the first decision variable in the DAG $G$ of the influence diagram (i.e. the decision variable with no parent decision variable). Then, repeatedly update $Sov$ by $Sov \leftarrow Sov.(+, pa_G(d)).(\max, \{d\})$ and delete $d$ and the variables in $pa_G(d)$ from $G$ until no decision variable remains. Then, perform $Sov \leftarrow Sov.(+, S)$, where $S$ is the set of chance variables that have not been deleted from $G$.

9. (*Finite horizon MDPs, Puterman, 1994; Monahan, 1982; Sabbadin, 1999; Boutilier et al., 1999, 2000*) In order to prove that the encoding in the PFU framework given in Sections 5.6 and 6.6 actually enables us to solve a $T$ time-steps probabilistic MDP, we start by reminding the algorithm used to compute an optimal MDP-policy. Usually, a decision rule for $d_T$ is chosen by computing $V^*_{s_T} = \max_{d_T} R_{s_T,d_T}$. $V^*_{s_T}$ is the optimal reward which can be obtained in state $s_T$. At a time-step $i \in [1, T[$, a decision rule for $d_i$ is chosen by computing $V^*_{s_i} = \max_{d_i} (R_{s_i,d_i} + \sum_{s_{i+1}} P_{s_{i+1} \mid s_i,d_i} \times V^*_{s_{i+1}})$. Last, the optimal expected value of the reward, which depends on the initial state $s_1$, is $V^*_{s_1}$.
Let us prove by induction that for all $i \in [1, T-1]$,
$$V^*_{s_1} = \max_{d_1} \sum_{s_2} \ldots \max_{d_i} \sum_{s_{i+1}} ((\prod_{k \in [1,i]} P_{s_{k+1} \mid s_k,d_k}) \times ((\sum_{k \in [1,i]} R_{s_k,d_k}) + V^*_{s_{i+1}})).$$
This proposition holds for $i = 1$, since
$$\begin{aligned} V^*_{s_1} &= \max_{d_1} (R_{s_1,d_1} + \textstyle\sum_{s_2} P_{s_2 \mid s_1,d_1} \times V^*_{s_2}) \\ &= \max_{d_1} \textstyle\sum_{s_2} (P_{s_2 \mid s_1,d_1} \times (R_{s_1,d_1} + V^*_{s_2})) \text{ (since } \textstyle\sum_{s_2} P_{s_2 \mid s_1,d_1} = 1). \end{aligned}$$

Moreover, if the proposition holds at step $i-1$ (with $i > 1$), then
$V^*_{s_1} = \max_{d_1} \sum_{s_2} \ldots \max_{d_{i-1}} \sum_{s_i} ((\prod_{k\in[1,i-1]} P_{s_{k+1} \,|\, s_k,d_k}) \times ((\sum_{k\in[1,i-1]} R_{s_k,d_k}) + V^*_{s_i}))$.
Given that

$$\begin{aligned}(\textstyle\sum_{k\in[1,i-1]} R_{s_k,d_k}) + V^*_{s_i} &= (\textstyle\sum_{k\in[1,i-1]} R_{s_k,d_k}) + \max_{d_i} (R_{s_i,d_i} + \sum_{s_{i+1}} P_{s_{i+1} \,|\, s_i,d_i} \times V^*_{s_{i+1}}) \\ &= \max_{d_i} ((\textstyle\sum_{k\in[1,i]} R_{s_k,d_k}) + \sum_{s_{i+1}} P_{s_{i+1} \,|\, s_i,d_i} \times V^*_{s_{i+1}}) \\ &= \max_{d_i} \textstyle\sum_{s_{i+1}} P_{s_{i+1} \,|\, s_i,d_i} \times ((\sum_{k\in[1,i]} R_{s_k,d_k}) + V^*_{s_{i+1}})\end{aligned}$$

(the last equality holds since $\sum_{s_{i+1}} P_{s_{i+1} \,|\, s_i,d_i} = 1$), it can be inferred that

$$\begin{aligned}&(\textstyle\prod_{k\in[1,i-1]} P_{s_{k+1} \,|\, s_k,d_k}) \times ((\sum_{k\in[1,i-1]} R_{s_k,d_k}) + V^*_{s_i}) \\ &= \max_{d_i} \textstyle\sum_{s_{i+1}} ((\prod_{k\in[1,i]} P_{s_{k+1} \,|\, s_k,d_k}) \times ((\sum_{k\in[1,i]} R_{s_k,d_k}) + V^*_{s_{i+1}})),\end{aligned}$$

which proves that the proposition holds at step $i$. This proves that it also holds at step $T$, and therefore $V^*_{s_1} = Ans(Q)$. Furthermore, as each step in the proof preserves the set of optimal decision rules, an optimal MDP-policy can be recorded during the evaluation of $Ans(Q)$.

We now study the case of partially observable finite horizon MDPs (finite horizon POMDPs). In a POMDP, we add for each time step $t > 1$ a conditional probability distribution $P_{o_t \,|\, s_t}$ of making observation $o_t$ at time step $t$ given the state $s_t$. The value of $s_t$ remains unobserved. We also assume that a probability distribution $P_{s_1}$ over the initial state is available. The subsumption proof for this case is more difficult. We consider the approach of POMDPs which consists in finding an optimal *policy tree*. This approach is equivalent to compute, for each decision variable $d_t$, a decision rule for $d_t$ depending on the observations made so far, i.e. a function $\phi^{d_t} : dom(\{o_2, \ldots, o_t\}) \to dom(d_t)$. The set of such functions is denoted $\Phi^{d_t}$. A set $\Delta = \{\phi^{d_1}, \ldots, \phi^{d_T}\}$ is called a POMDP-policy. The value of a POMDP-policy is recursively defined as follows. First, the value of the reward at the last decision step, which depends on the assignment $A_{s_T}$ of $s_T$ and on the observations $O_{2\to T}$ made from the beginning, is $V(\Delta)_{s_T,o_2,\ldots,o_t}(A_{s_T}.O_{2\to T}) = R_{s_T,d_T}(A_{s_T}, \phi^{d_T}(O_{2\to T}))$. At a time step $i$, the obtained reward depends on the actual state $A_{s_i}$ and on the observations $O_{2\to i}$ made so far. Its expression is:

$$\begin{aligned}&V(\Delta)_{s_i,o_2,\ldots,o_i}(A_{s_i}.O_{2\to i}) \\ &= (R_{s_i,d_i} + \textstyle\sum_{s_{i+1}} P_{s_{i+1} \,|\, s_i,d_i} \times \sum_{o_{i+1}} P_{o_{i+1} \,|\, s_{i+1}} \times V(\Delta)_{s_{i+1},o_1,\ldots,o_{i+1}})(A)\end{aligned}$$

where $A = A_{s_i}.\phi^{d_i}(O_{2\to i}).O_{2\to i}$ This equation and the recursive formula used to define the value of a policy tree for a POMDP (Kaelbling, Littman, & Cassandra, 1998) are equivalent. Finally, the expected reward of the POMDP-policy $\Delta$ is $V(\Delta) = \sum_{s_1} P_{s_1} \times V(\Delta)_{s_1}$. Solving a finite horizon POMDP consists in computing the optimal expected reward among all POMDP-policies (i.e. in computing $V^* = \max_{\phi^{d_1},\ldots,\phi^{d_T}} V(\{\phi^{d_1}, \ldots, \phi^{d_T}\})$), as well as associated optimal decision rules.

Using a proof by induction as in the observable MDP case, it is first possible to prove that for a problem with $T$ steps,

$$V^* = \max_{\phi^{d_1},\ldots,\phi^{d_T}} \textstyle\sum_{o_2,\ldots,o_T} \sum_{s_1,\ldots,s_T} \beta_V$$

with $\beta_V = (P_{s_1} \times \prod_{i\in[1,T[} P_{s_{i+1} \,|\, s_i,d_i} \times \prod_{i\in[1,T[} P_{o_{i+1} \,|\, s_{i+1}}) \times (\sum_{i\in[1,T]} R_{s_i,d_i})$.

From this, a recursive use of Lemma 3 enables us to infer that

$$V^* = \max_{d_1} \textstyle\sum_{o_2} \max_{d_2} \sum_{o_3} \max_{d_3} \ldots \sum_{o_T} \max_{d_T} \sum_{s_1,\ldots,s_T} \beta_V.$$

It proves that the query defined below enables us to compute $V^*$ as well as an optimal policy:

- algebraic structure: probabilistic expected additive utility (row 1 in Table 1);
- PFU network: $\mathcal{N} = (V, G, P, \emptyset, U)$; $V$ equals $\{s_i \,|\, i \in [1,T]\} \cup \{o_i \,|\, i \in [2,T]\} \cup \{d_i \,|\, i \in [1,T]\}$, with $V_D = \{d_i \,|\, i \in [1,T]\}$; $G$ is a DAG with one variable per component; a decision component does not have any parents, an environment component $\{o_i\}$ has $\{s_i\}$ as parent, and a component $\{s_{i+1}\}$ has $\{s_i\}$ and $\{d_i\}$ as parents; $P = \{P_{s_1}\} \cup \{P_{s_{i+1} \,|\, s_i,d_i}, i \in [1,T-1]\} \cup \{P_{o_i \,|\, s_i} \,|\, i \in [2,T]\}$; $Fact(\{s_1\}) = \{P_{s_1}\}$, $Fact(\{s_{i+1}\}) = \{P_{s_{i+1} \,|\, s_i,d_i}\}$, and $Fact(\{o_i\}) = \{P_{o_i \,|\, s_i}\}$; last, $U = \{R_{s_i,d_i} \,|\, i \in [1,T]\}$;

- PFU query: based on the DAG, a necessary condition for a query to be defined is that each decision $d_i$ must appear to the left of the variables in $\{s_k \mid k \in [i+1,T]\} \cup \{o_k \mid k \in [i+1,T]\}$; the query considered is $Q = (\mathcal{N}, Sov)$, with
$$Sov = (\max, d_1).(+, o_2).(\max, d_2).\ldots.(+, o_T).(\max, d_T).(+, \{s_1, \ldots, s_T\}).$$

The proofs for finite horizon (PO)MDPs based on possibilities or on $\kappa$-rankings are similar. As for the subsumption of factored MDPs, we can first argue that every factored MDP can be represented as a usual MDP, and therefore as a PFU query on a PFU network. Even if this is a sufficient argument, we can define a better representation of factored MDPs in the PFU framework: it corresponds to a representation where the variables describing states are directly used together with the local plausibility functions and rewards, which can be modeled by scoped functions (defined as decision trees, binary decision diagrams...).

10. (*Queries on Bayesian networks, Pearl, 1988, Markov random fields, Chellappa & Jain, 1993, and chain graphs, Frydenberg, 1990*)
It suffices to consider chain graphs, since Bayesian networks and Markov random fields are particular cases of chain graphs. The subsumption proofs are provided for the general case of plausibility distributions defined on a totally ordered conditionable plausibility structure.

  (a) (*MAP, MPE, and probability of an evidence*) As MPE (Most Probable Explanation) and the computation of the probability of an evidence are particular cases of MAP (Maximum A Posteriori hypothesis), it suffices to prove that MAP is subsumed. The probabilistic MAP problem consists in finding, given a probability distribution $\mathcal{P}_V$, a Maximum A Posteriori explanation to an assignment of a subset $O$ of $V$ which has been observed (also called evidence). More formally, let $D$ denote the set of variables on which an explanation is sought and let $e$ denote the observed assignment of $O$. The MAP problem consists in finding an assignment $A^*$ of $D$ such that $\max_{A \in dom(D)} P_{D \mid O}(A.e) = P_{D \mid O}(A^*.e)$. As $P_{D \mid O} = P_{D,O}/P_O$, we can write:
$$\begin{aligned}\max_{A \in dom(D)} P_{D \mid O}(A.e) &= (\max_{A \in dom(D)} P_{D,O}(A.e))/P_O(e)\\ &= (\max_{A \in dom(D)} \textstyle\sum_{A' \in dom(V-(D \cup O))} P_V(A.e.A'))/P_O(e)\end{aligned}$$
Thus, computing $\max_D \sum_{V-(D\cup O)} P_V(e)$ is sufficient (the difference lies only in a normalizing constant). This result can be generalized to all totally ordered conditionable plausibility structures.
Indeed, as $\otimes_p$ is monotonic, $\max_{A \in dom(D)} \mathcal{P}_{D,O}(A.e) = (\max_{A \in dom(D)} \mathcal{P}_{D \mid O}(A.e)) \otimes_p \mathcal{P}_O(e)$. If $\max_{A \in dom(D)} \mathcal{P}_{D,O}(A.e) \prec_p \mathcal{P}_O(e)$, then there exists a unique $p \in E_p$ such that $\max_{A \in dom(D)} \mathcal{P}_{D,O}(A.e) = p \otimes_p \mathcal{P}_O(e)$. This gives us $p = \max_{A \in dom(D)} \mathcal{P}_{D \mid O}(A.e)$. Otherwise, if $\max_{A \in dom(D)} \mathcal{P}_{D,O}(A.e) = \mathcal{P}_O(e)$, then we can infer that there exists $A^* \in dom(D)$ such that $\mathcal{P}_{D,O}(A^*.e) = \mathcal{P}_O(e)$, and therefore $\mathcal{P}_{D \mid O}(A^*.e) = 1_p$. Thus, $\max_{A \in dom(D)} \mathcal{P}_{D \mid O}(A.e) = 1_p$ too. This shows that determining $\max_{A \in dom(D)} \mathcal{P}_{D,O}(A.e)$ gives $\max_{A \in dom(D)} \mathcal{P}_{D \mid O}(A.e)$.
Moreover, if $A^* \in \operatorname{argmax}\{\mathcal{P}_{D,O}(A'.e), A' \in dom(D)\}$, then $\max\{p \in E_p \mid \mathcal{P}_{D,O}(A^*.e) = p \otimes_p \mathcal{P}_O(e)\} \succeq_p \max\{p \in E_p \mid \mathcal{P}_{D,O}(A.e) = p \otimes_p \mathcal{P}_O(e)\}$ for all $A \in dom(D)$. Therefore, an optimal assignment of $D$ for $\max_D \mathcal{P}_{D,O}(e)$ is also an optimal assignment of $D$ for $\max_D \mathcal{P}_{D \mid O}(e)$. As a result, the MAP problem can be reduced to the computation of
$$\max_D \mathcal{P}_{D,O}(e) = \max_D \oplus_{pV-(D\cup O)} \mathcal{P}_V(e) = \max_D \oplus_{pV-D} (\mathcal{P}_V \otimes_p \delta_O)$$
where $\delta_O$ is the scoped function with scope $O$ such that $\delta_O(e') = 1_p$ if $e' = e$, $0_p$ otherwise. We define a PFU query whose answer is $Ans(Q) = \max_D \oplus_{pV-D}(\mathcal{P}_V \otimes_p \delta_O)$:

  - the plausibility structure is $(E_p, \oplus_p, \otimes_p)$, the utility structure is $(E_u, \otimes_u) = (E_p, \otimes_p)$, and the expected utility structure is $(E_p, E_u, \oplus_u, \otimes_{pu}) = (E_p, E_p, \oplus_p, \otimes_p)$;
  - PFU network: the difficulty in the definition of the PFU network lies in the fact that normalization conditions on components must be satisfied. The idea is that only the components in which a variable in $D \cup O$ is involved have to be modified. The PFU network is $\mathcal{N} = (V, G, P, \emptyset, U)$; $V$ the set of variables of the chain graph; $V_D = D$ and $V_E = V - D$; $G$ is a DAG of components obtained from the DAG $G'$ of the chain graph by splitting every component $c$ in which a variable in $D \cup O$ is involved: such a component $c$ is transformed into $|c|$ components containing only one variable; all these $|c|$ components become parents of the child components of $c$; for a component $\{x_0\}$ included in one of these $|c|$ components, if $x_0 \in D$, then $\{x_0\}$ is a decision component; otherwise, $\{x_0\}$ is an environment component, and we create a plausibility function $P_i$, equal to a constant $p_0(x_0)$ such that $\oplus_{p_{i \in [1, |dom(x_0)|]}} p_0(x_0) = 1_p$, and such that $Fact(\{x_0\}) = \{p_0(x_0)\}$; $P$ contains first the constants defined above, and second the factors expressing $P_{c \,|\, pa_{G'}(c)}$ in the chain graph for the components $c$ satisfying $c \cap (D \cup O) = \emptyset$; last, $U$ contains the factors expressing $P_{c \,|\, pa_{G'}(c)}$ in the chain graph for the components $c$ such that $c \cap (D \cup O) \neq \emptyset$, and a constant factor $p_1(x_0)$ satisfying $p_1(x_0) \otimes_p p_0(x_0) = 1_p$ for each component $\{x_0\}$ created in the splitting process described above, and hard constraints representing $\delta_O$; with this PFU network, the local normalization conditions are satisfied, and the combination of the local functions equals $\mathcal{P}_V \otimes_p \delta_O$;

- PFU query: the query is simply $Q = (\mathcal{N}, (\max, D).(\oplus_u, V - D))$.

An optimal decision rule for $D$ can be recorded during the computation of $Ans(Q)$.

(b) (*Plausibility distribution computation task*) Given a plausibility distribution $\mathcal{P}_V$ expressed as a combination of plausibility functions as in chain graphs, the goal is to compute the plausibility distribution $\mathcal{P}_S$ over a set $S \subseteq V$. The basic formula $\mathcal{P}_S = \oplus_{p_{V-S}} \mathcal{P}_V$ proves that the query defined below actually computes $\mathcal{P}_S$. This query shows the usefulness of free variables.

- the plausibility structure is $(E_p, \oplus_p, \otimes_p)$, the utility structure is $(E_u, \otimes_u) = (E_p, \otimes_p)$, and the expected utility structure is $(E_p, E_u, \oplus_u, \otimes_{pu}) = (E_p, E_p, \oplus_p, \otimes_p)$;
- PFU network: $\mathcal{N} = (V, G, P, \emptyset, U)$, with $V_E = V - S$, $V_D = S$, and with the DAG $G$ and the sets $P$, $U$ obtained similarly as for the MAP case;
- PFU query: $Q = (\mathcal{N}, (\oplus_u, V - S))$

11. (*Hybrid networks, Dechter & Larkin, 2001*)
A hybrid network is a triple $(G, P, F)$, where $G$ is a DAG on a set of variables $V$ partitioned into $R$ and $D$, $P$ is a set of probability distributions expressing $P_{r \,|\, pa_G(r)}$ for all $r \in R$, and $F$ is a set of functions $f_{pa_G(d)}$ for all $d \in D$ (variables in $D$ are deterministic, in the sense that their value is completely determined by the assignment of their parents). The most general task on hybrid networks is the task of belief assessment conditioned on a formula $\phi$ in conjunctive normal form. It consists of computing the probability distribution of a variable $x$ given a complex evidence $\phi$ (complex because it may involve several variables). Ignoring a normalizing constant, it requires to compute, for all assignments $(x, a)$ of $x$, $\sum_{A \in dom(V - \{x\}) \,|\, \phi(A.(x,a)) = t} P_V(A.(x, a))$. If $C = \{C_1, \ldots, C_m\}$ denotes the set of clauses of $\phi$, it also equals $(\sum_{V - \{x\}} (\prod_{r \in R} P_{r \,|\, pa_G(r)}) \times (\prod_{d \in D} f_{pa_G(d)}) \times (\prod_{C_i \in C} C_i))((x, a))$.
The query corresponding to this computation uses the probabilistic expected satisfaction structure (row 2 in Table 1), and the PFU network $\mathcal{N} = (V, G, P, \emptyset, U)$, with $V_E = V$, $V_D = \{x\}$, $P = \{P_{r \,|\, pa_G(r)} \,|\, r \in R - \{x\}\} \cup \{f_{pa_G(d)} \,|\, d \in D - \{x\}\}$, and either $U = C \cup \{P_{x \,|\, pa_G(x)}\}$ if $x \in R$ or $U = C \cup \{f_{pa_G(x)}\}$ if $x \in D$. The query is $Q = (\mathcal{N}, (+, V - \{x\}))$.

□

## References

Bacchus, F., & Grove, A. (1995). Graphical Models for Preference and Utility. In *Proc. of the 11th International Conference on Uncertainty in Artificial Intelligence (UAI-95)*, pp. 3–10, Montréal, Canada.

Bahar, R., Frohm, E., Gaona, C., Hachtel, G., Macii, E., Pardo, A., & Somenzi, F. (1993). Algebraic Decision Diagrams and Their Applications. In *IEEE /ACM International Conference on CAD*, pp. 188–191, Santa Clara, California, USA. IEEE Computer Society Press.

Bertelé, U., & Brioschi, F. (1972). *Nonserial Dynamic Programming*. Academic Press.

Bistarelli, S., Montanari, U., Rossi, F., Schiex, T., Verfaillie, G., & Fargier, H. (1999). Semiring-Based CSPs and Valued CSPs: Frameworks, Properties and Comparison. *Constraints*, *4*(3), 199–240.

Bodlaender, H. (1997). Treewidth: Algorithmic techniques and results. In *Proc. of the 22nd International Symposium on Mathematical Foundations of Computer Science (MFCS-97)*.

Bordeaux, L., & Monfroy, E. (2002). Beyond NP: Arc-consistency for Quantified Constraints. In *Proc. of the 8th International Conference on Principles and Practice of Constraint Programming (CP-02)*, Ithaca, New York, USA.

Boutilier, C., Brafman, R., Domshlak, C., Hoos, H., & Poole, D. (2004). CP-nets: A Tool for Representing and Reasoning with Conditional Ceteris Paribus Preference Statements. *Journal of Artificial Intelligence Research*, *21*, 135–191.

Boutilier, C., Dean, T., & Hanks, S. (1999). Decision-Theoretic Planning: Structural Assumptions and Computational Leverage. *Journal of Artificial Intelligence Research*, *11*, 1–94.

Boutilier, C., Dearden, R., & Goldszmidt, M. (2000). Stochastic Dynamic Programming with Factored Representations. *Artificial Intelligence*, *121*(1-2), 49–107.

Boutilier, C., Friedman, N., Goldszmidt, M., & Koller, D. (1996). Context-Specific Independence in Bayesian Networks. In *Proc. of the 12th International Conference on Uncertainty in Artificial Intelligence (UAI-96)*, pp. 115–123, Portland, Oregon, USA.

Chellappa, R., & Jain, A. (1993). Markov Random Fields: Theory and Applications. Academic Press.

Chu, F., & Halpern, J. (2003a). Great Expectations. Part I: On the Customizability of Generalized Expected Utility. In *Proc. of the 18th International Joint Conference on Artificial Intelligence (IJCAI-03)*, Acapulco, Mexico.

Chu, F., & Halpern, J. (2003b). Great Expectations. Part II: Generalized Expected Utility as a Universal Decision Rule. In *Proc. of the 18th International Joint Conference on Artificial Intelligence (IJCAI-03)*, pp. 291–296, Acapulco, Mexico.

Cooper, M., & Schiex, T. (2004). Arc Consistency for Soft Constraints. *Artificial Intelligence*, *154*(1-2), 199–227.

Darwiche, A. (2001). Recursive Conditioning. *Artificial Intelligence*, *126*(1-2), 5–41.

Darwiche, A., & Ginsberg, M. (1992). A Symbolic Generalization of Probability Theory. In *Proc. of the 10th National Conference on Artificial Intelligence (AAAI-92)*, pp. 622–627, San Jose, CA, USA.

Dechter, R. (1999). Bucket Elimination: a Unifying Framework for Reasoning. *Artificial Intelligence*, *113*(1-2), 41–85.

Dechter, R., & Fattah, Y. E. (2001). Topological Parameters for Time-Space Tradeoff. *Artificial Intelligence*, *125*(1-2), 93–118.

Dechter, R., & Larkin, D. (2001). Hybrid Processing of Beliefs and Constraints. In *Proc. of the 17th International Conference on Uncertainty in Artificial Intelligence (UAI-01)*, pp. 112–119, Seattle, WA, USA.

Dechter, R., & Mateescu, R. (2004). Mixtures of Deterministic-Probabilistic Networks and their AND/OR Search Space. In *Proc. of the 20th International Conference on Uncertainty in Artificial Intelligence (UAI-04)*, Banff, Canada.

Demirer, R., & Shenoy, P. (2001). Sequential Valuation Networks: A New Graphical Technique for Asymmetric Decision Problems. In *Proc. of the 6th European Conference on Symbolic and Quantitavive Approaches to Reasoning with Uncertainty (ECSQARU-01)*, pp. 252–265, London, UK.

Dubois, D., & Prade, H. (1995). Possibility Theory as a Basis for Qualitative Decision Theory. In *Proc. of the 14th International Joint Conference on Artificial Intelligence (IJCAI-95)*, pp. 1925–1930, Montréal, Canada.

Fargier, H., Lang, J., & Schiex, T. (1996). Mixed Constraint Satisfaction: a Framework for Decision Problems under Incomplete Knowledge. In *Proc. of the 13th National Conference on Artificial Intelligence (AAAI-96)*, pp. 175–180, Portland, OR, USA.

Fargier, H., & Perny, P. (1999). Qualitative Models for Decision Under Uncertainty without the Commensurability Assumption. In *Proc. of the 15th International Conference on Uncertainty in Artificial Intelligence (UAI-99)*, pp. 188–195, Stockholm, Sweden.

Fikes, R., & Nilsson, N. (1971). STRIPS: a New Approach to the Application of Theorem Proving. *Artificial Intelligence*, *2*(3-4), 189–208.

Fishburn, P. (1982). *The Foundations of Expected Utility*. D. Reidel Publishing Company, Dordrecht.

Friedman, N., & Halpern, J. (1995). Plausibility Measures: A User's Guide. In *Proc. of the 11th International Conference on Uncertainty in Artificial Intelligence (UAI-95)*, pp. 175–184, Montréal, Canada.

Frydenberg, M. (1990). The Chain Graph Markov Property. *Scandinavian Journal of Statistics*, *17*, 333–353.

Ghallab, M., Nau, D., & Traverso, P. (2004). *Automated Planning: Theory and Practice*. Morgan Kaufmann.

Giang, P., & Shenoy, P. (2000). A Qualitative Linear Utility Theory for Spohn's Theory of Epistemic Beliefs. In *Proc. of the 16th International Conference on Uncertainty in Artificial Intelligence (UAI-00)*, pp. 220–229, Stanford, California, USA.

Giang, P., & Shenoy, P. (2005). Two Axiomatic Approaches to Decision Making Using Possibility Theory. *European Journal of Operational Research*, *162*(2), 450–467.

Goldman, R., & Boddy, M. (1996). Expressive Planning and Explicit Knowledge. In *Proc. of the 3rd International Conference on Artificial Intelligence Planning Systems (AIPS-96)*, pp. 110–117, Edinburgh, Scotland.

Halpern, J. (2001). Conditional Plausibility Measures and Bayesian Networks. *Journal of Artificial Intelligence Research*, *14*, 359–389.

Halpern, J. (2003). *Reasoning about Uncertainty*. The MIT Press.

Howard, R., & Matheson, J. (1984). Influence Diagrams. In *Readings on the Principles and Applications of Decision Analysis*, pp. 721–762. Strategic Decisions Group, Menlo Park, CA, USA.

Jégou, P., & Terrioux, C. (2003). Hybrid Backtracking bounded by Tree-decomposition of Constraint Networks. *Artificial Intelligence*, *146*(1), 43–75.

Jensen, F., Nielsen, T., & Shenoy, P. (2004). Sequential Influence Diagrams: A Unified Asymmetry Framework. In *Proceedings of the Second European Workshop on Probabilistic Graphical Models (PGM-04)*, pp. 121–128, Leiden, Netherlands.

Jensen, F., & Vomlelova, M. (2002). Unconstrained Influence Diagrams. In *Proc. of the 18th International Conference on Uncertainty in Artificial Intelligence (UAI-02)*, pp. 234–241, Seattle, WA, USA.

Kaelbling, L., Littman, M., & Cassandra, A. (1998). Planning and Acting in Partially Observable Stochastic Domains. *Artificial Intelligence*, *101*(1-2), 99–134.

Kolhas, J. (2003). *Information Algebras: Generic Structures for Inference*. Springer.

Kushmerick, N., Hanks, S., & Weld, D. (1995). An Algorithm for Probabilistic Planning. *Artificial Intelligence*, *76*(1-2), 239–286.

Larrosa, J., & Schiex., T. (2003). In the quest of the best form of local consistency for weighted csp. In *Proc. of the 18th International Joint Conference on Artificial Intelligence (IJCAI-03)*, Acapulco, Mexico.

Lauritzen, S., & Nilsson, D. (2001). Representing and Solving Decision Problems with Limited Information. *Management Science*, *47*(9), 1235–1251.

Littman, M., Majercik, S., & Pitassi, T. (2001). Stochastic Boolean Satisfiability. *Journal of Automated Reasoning*, *27*(3), 251–296.

Mackworth, A. (1977). Consistency in Networks of Relations. *Artificial Intelligence*, *8*(1), 99–118.

Monahan, G. (1982). A Survey of Partially Observable Markov Decision Processes: Theory, Models, and Algorithms. *Management Science*, *28*(1), 1–16.

Ndilikilikesha, P. (1994). Potential Influence Diagrams. *International Journal of Approximated Reasoning*, *10*, 251–285.

Nielsen, T., & Jensen, F. (2003). Representing and solving asymmetric decision problems. *International Journal of Information Technology and Decision Making*, *2*, 217–263.

Pearl, J. (1988). *Probabilistic Reasoning in Intelligent Systems: Networks of Plausible Inference*. Morgan Kaufmann.

Perny, P., Spanjaard, O., & Weng, P. (2005). Algebraic Markov Decision Processes. In *Proc. of the 19th International Joint Conference on Artificial Intelligence (IJCAI-05)*, Edinburgh, Scotland.

Pralet, C. (2006). *A Generic Algebraic Framework for Representing and Solving Sequential Decision Making Problems with Uncertainties, Feasibilities, and Utilities*. Ph.D. thesis, Ecole Nationale Supérieure de l'Aéronautique et de l'Espace, Toulouse, France.

Pralet, C., Schiex, T., & Verfaillie, G. (2006a). Decomposition of Multi-Operator Queries on Semiring-based Graphical Models. In *Proc. of the 12th International Conference on Principles and Practice of Constraint Programming (CP-06)*, pp. 437–452, Nantes, France.

Pralet, C., Schiex, T., & Verfaillie, G. (2006b). From Influence Diagrams to Multioperator Cluster DAGs. In *Proc. of the 22nd International Conference on Uncertainty in Artificial Intelligence (UAI-06)*, Cambridge, MA, USA.

Pralet, C., Verfaillie, G., & Schiex, T. (2006c). Decision with Uncertainties, Feasibilities, and Utilities: Towards a Unified Algebraic Framework. In *Proc. of the 17th European Conference on Artificial Intelligence (ECAI-06)*, pp. 427–431, Riva del Garda, Italy.

Puterman, M. (1994). *Markov Decision Processes, Discrete Stochastic Dynamic Programming*. John Wiley & Sons.

Sabbadin, R. (1999). A Possibilistic Model for Qualitative Sequential Decision Problems under Uncertainty in Partially Observable Environments. In *Proc. of the 15th International Conference on Uncertainty in Artificial Intelligence (UAI-99)*, pp. 567–574, Stockholm, Sweden.

Sang, T., Beame, P., & Kautz, H. (2005). Solving Bayesian Networks by Weighted Model Counting. In *Proc. of the 20th National Conference on Artificial Intelligence (AAAI-05)*, pp. 475–482, Pittsburgh, PA, USA.

Schmeidler, D. (1989). Subjective Probability and Expected Utility without Additivity. *Econometrica*, *57*(3), 571–587.

Schrijver, A. (1998). *Theory of Linear and Integer Programming*. John Wiley and Sons.

Shafer, G. (1976). *A Mathematical Theory of Evidence*. Princeton University Press.

Shenoy, P. (1991). Valuation-based Systems for Discrete Optimization. *Uncertainty in Artificial Intelligence*, *6*, 385–400.

Shenoy, P. (1992). Valuation-based Systems for Bayesian Decision Analysis. *Operations Research*, *40*(3), 463–484.

Shenoy, P. (1994). Conditional Independence in Valuation-Based Systems. *International Journal of Approximated Reasoning*, *10*(3), 203–234.

Shenoy, P. (2000). Valuation Network Representation and Solution of Asymmetric Decision Problems. *European Journal of Operational Research*, *121*, 579–608.

Smith, J., Holtzman, S., & Matheson, J. (1993). Structuring Conditional Relationships in Influence Diagrams. *Operations Research*, *41*, 280–297.

Spohn, W. (1990). A General Non-Probabilistic Theory of Inductive Reasoning. In *Proc. of the 6th International Conference on Uncertainty in Artificial Intelligence (UAI-90)*, pp. 149–158, Cambridge, MA, USA.

von Neumann, J., & Morgenstern, O. (1944). *Theory of Games and Economic Behaviour*. Princeton University Press.

Walsh, T. (2002). Stochastic Constraint Programming. In *Proc. of the 15th European Conference on Artificial Intelligence (ECAI-02)*, pp. 111–115, Lyon, France.

Weydert, E. (1994). General Belief Measures. In *Proc. of the 10th International Conference on Uncertainty in Artificial Intelligence (UAI-94)*, pp. 575–582.

Wilson, N. (1995). An Order of Magnitude Calculus. In *Proc. of the 11th International Conference on Uncertainty in Artificial Intelligence (UAI-95)*, pp. 548–555, Montréal, Canada.